\documentclass[letterpaper]{article}
\usepackage[letterpaper]{geometry}

\usepackage[utf8]{inputenc}
\usepackage[T1]{fontenc} 
\usepackage[cjk]{kotex}
\usepackage{amsmath}
\usepackage{graphicx}
\usepackage{url}
\usepackage[table]{xcolor}

\usepackage{latexsym}
\usepackage{amssymb}   
\usepackage{amsfonts}

\usepackage{latexsym}

\usepackage{booktabs}
\usepackage{longtable}
\usepackage{pdflscape}

\usepackage{arydshln} 
\usepackage{bbding} 

\usepackage{hyperref}
\definecolor{darkblue}{rgb}{0, 0, 0.5}
\hypersetup{colorlinks=true,citecolor=darkblue, linkcolor=darkblue, urlcolor=darkblue}

\usepackage{tabularx}

\usepackage{setspace}
\usepackage{lineno}

\usepackage[authoryear,round]{natbib}
\usepackage{algorithmic,algorithm}
\renewcommand{\algorithmiccomment}[1]{\bgroup\hfill$\triangleright$~#1\egroup}

\usepackage[linguistics]{forest} 
\usepackage{synttree}
\usepackage{subcaption}
\usepackage{tikz-dependency}

\newcommand{\zh}[1]{\begin{CJK}{UTF8}{gbsn}#1\end{CJK}}
\newcommand{\np}[1]{\begin{CJK}{UTF8}{min}#1\end{CJK}}

\newcommand*{\jp}[1]{{\color{black}#1}}
\newcommand*{\mq}[1]{{\color{black}#1}}

\usepackage{langsci-gb4e}

\title{\textbf{Chinese Grammatical Error Correction: A Survey}}

\author{ 
\textbf{Mengyang Qiu}$^{\dagger\ddagger}$~~~ \textbf{Qingyu Gao$^{\ddagger}$}~~~ \textbf{Linxuan Yang}$^{\S}$~~~ \textbf{Yang Gu}$^{\ddagger}$ \\
\textbf{Tran Minh Nguyen}$^{\ddagger}$~~~ \textbf{Zihao Huang}$^{\ddagger}$~~~ \textbf{Jungyeul Park}$^{\P\ddagger}$\thanks{Corresponding author: Jungyeul Park, Email:\url{jungyeul@kaist.ac.kr}.} \\
\url{https://open-writing-evaluation.github.io}
}
\date{ 
June 2026
}

\begin{document}

\maketitle

\begin{abstract}
Chinese Grammatical Error Correction (CGEC) is a critical task in natural language processing, addressing the growing demand for automated writing assistance in both second-language (L2) and native-speaker (L1) Chinese writing. While L2 learners face challenges in mastering complex grammatical structures, L1 users also benefit from CGEC in academic, professional, and formal contexts where writing precision is essential. This survey provides a comprehensive review of CGEC research, covering datasets and their annotation schemes, evaluation methodologies, and system developments. We examine widely used CGEC datasets, highlighting their characteristics, limitations, and the need for improved standardization. We also discuss dataset-specific annotation schemes and error representation formats, with particular attention to word segmentation ambiguity and the classification of Chinese-specific error types. Furthermore, we review evaluation metrics, focusing on their adaptation from English GEC to Chinese, including character-level scoring and the use of multiple references. In terms of system development, we trace the evolution from rule-based and statistical approaches to neural architectures, including Transformer-based models and large pre-trained language models. By consolidating existing research and identifying key challenges, this survey characterizes the current state of CGEC and outlines future directions, including the refinement of annotation standards for segmentation-sensitive error representation and the use of multilingual approaches to improve CGEC.
\end{abstract}

\tableofcontents

\doublespacing

\section{Introduction} \label{sec:intro}

Grammatical Error Correction (GEC), which aims to automatically detect and correct errors in written text, has established itself as one of the key tasks in Natural Language Processing (NLP). It has broad applications, serving both native speakers (L1) and language learners (L2). For L1 users, GEC enhances clarity and fluency, improving overall writing quality, especially in professional communication. For L2 learners, it functions as an instructional tool, providing structured feedback that reinforces correct grammatical patterns, builds writing confidence, and ultimately aids language acquisition. Given its practical significance, GEC has been an active area of research, with notable progress in system development, error annotation, and evaluation. These two roles, computational correction and pedagogically meaningful feedback, are closely related rather than separate. Choices about how errors are segmented, represented, annotated, and evaluated shape not only what systems learn to correct, but also what kinds of learner difficulties become visible and what forms of feedback can be meaningfully supported.

The evolution of GEC systems has followed a clear technological progression over the past two decades. Early approaches relied primarily on rule-based methods and statistical classifiers that targeted specific error types, such as article or preposition errors. A significant shift occurred when researchers began framing GEC as a translation problem, applying Statistical Machine Translation (SMT) techniques to ``translate'' from erroneous to correct text. This approach demonstrated its effectiveness at the \textit{CoNLL-2014 Shared Task} \citep{ng-etal-2014-conll}, where SMT-based systems outperformed other methods \citep[e.g.,][]{junczys-dowmunt-grundkiewicz-2014-amu}. Neural Machine Translation (NMT) subsequently emerged as the dominant paradigm, with Transformer-based architectures establishing state-of-the-art performance through their ability to capture long-range dependencies and complex error patterns \citep{junczys-dowmunt-etal-2018-approaching,zhao-etal-2019-improving}. Recent advances include sequence tagging approaches that directly generate edit operations \citep{omelianchuk-etal-2020-gector}, and the application of large pre-trained language models that can be fine-tuned or prompted to perform correction \citep{rothe-etal-2021-simple,zeng-etal-2024-evaluating-prompting}. For a comprehensive review of these approaches, see \citet{wang-etal-2021-comprehensive,bryant-etal-2023-grammatical}.

Parallel to system development, significant progress has been made in error annotation and evaluation methodologies. For error annotation, \texttt{errant} (ERRor ANnotation Toolkit) has provided a systematic framework for identifying and categorizing linguistic errors, offering a detailed annotation scheme with 55 possible error types for English \citep{bryant-etal-2017-automatic}. On the evaluation front, metrics, such as \texttt{M$^2$} \citep{dahlmeier-ng-2012-better}, \texttt{GLEU} \citep{napoles-etal-2015-ground}, and \texttt{PT M$^2$} \citep{gong-etal-2022-revisiting}, have enabled standardized assessment of GEC systems. \texttt{errant} also serves as an evaluation tool, combining annotation capabilities with performance measurement. It not only enables consistent comparison across different systems but also provides insights into specific error categories that systems handle well or poorly. The development of these annotation and evaluation methods has been instrumental in benchmarking progress, particularly through shared tasks such as \textit{CoNLL-2013/2014} \citep{ng-EtAl:2013:CoNLL,ng-etal-2014-conll} and \textit{BEA-2019} \citep{bryant-etal-2019-bea}, which have significantly contributed to the field's advancement.

While these advancements in GEC have primarily focused on English, recent years have seen growing interest in developing similar capabilities for other languages, such as Czech \citep{naplava-etal-2022-czech}, German \citep{boyd-2018-using}, Korean \citep{yoon-etal-2023-towards}, and Ukrainian \citep{syvokon-etal-2023-ua}. 
Chinese, as one of the world's most widely spoken languages with over a billion native speakers, has naturally attracted significant attention from the research community. The increasing global importance of Chinese in business, education, and international relations has further accelerated demand for effective Chinese GEC (CGEC) systems. However, developing such systems presents unique challenges due to the fundamental differences between Chinese and Indo-European languages. These challenges can be broadly attributed to two main factors: the distinctive characteristics of the Chinese writing system and the language's particular syntactic features. In CGEC, these properties matter not only as general linguistic background, but because they directly affect the central computational units of the task: what counts as an error span, how edits are extracted, how corrections are aligned, and how system outputs can be compared across datasets and evaluation settings.

A primary challenge in CGEC stems from the lack of explicit word boundaries in written text. Unlike alphabetic languages where spaces clearly mark word boundaries, Chinese appears as a continuous sequence of characters with no visual separation between words. This creates difficulties for tokenization and word segmentation, critical first steps in most NLP pipelines, as words can consist of one, two, or more characters. The complexity is increased by characters that can serve multiple grammatical functions depending on context. For instance, the character \zh{把} \textit{bǎ} (`structural particle') can function as a structural particle in the \textit{ba}-construction, as a classifier as in \zh{一把葱} \textit{yī bǎ cōng} (`a bunch of green onions'), or as part of a word like \zh{把手} \textit{bǎ shǒu} (`handle') and \zh{把握} \textit{bǎ wò} (`to seize'). This functional ambiguity creates challenges during word segmentation, particularly in erroneous text where standard patterns may be disrupted. For CGEC, this is not merely a preprocessing inconvenience. Segmentation choices determine the units over which edits are localized, the granularity at which annotation schemes operate, and the comparability of evaluation results across resources and systems.

While word segmentation has become less problematic in current NMT approaches, which can effectively learn contextual representations without explicit segmentation, it remains a critical challenge in sequence tagging approaches, where models must make token-level predictions about grammatical operations \citep{liang-etal-2020-bert}. For error annotation systems, word boundary ambiguity poses even greater difficulties. The consequences for annotation accuracy are notable: incorrect segmentation can lead to misidentification of errors or failure to detect genuine errors. However, if we bypass word segmentation and rely solely on character-level annotation \citep{hinson-etal-2020-heterogeneous}, we lose critical grammatical information. Consider two sentences where \zh{把} \textit{bǎ} (`structural particle') is missing: \zh{他（把）书拿走了} \textit{tā (bǎ) shū ná zǒu le} (`He took the book away'), and \zh{菜市场买了一（把）葱} \textit{cài shì chǎng mǎi le yī (bǎ) cōng} (`Bought a bunch of green onions at the market'). Character-level annotation would label both cases as a generic ``missing'' error; distinguishing the underlying grammatical function then requires extra structure (e.g., word segmentation and syntactic attachment), which is precisely where Chinese CGEC pipelines can become unstable on noisy learner text.\label{r1-p3} This creates a direct link between computational design and pedagogical interpretation: a representation that collapses distinct grammatical functions into the same edit type may still support coarse scoring, but it is less informative about the specific grammatical difficulty the learner actually faced.

Apart from segmentation difficulties, the logographic nature of Chinese introduces another layer of complexity, namely errors stemming from phonetic and visual similarities. Unlike alphabetic systems where words are composed of phonetic letters, Chinese employs thousands of distinct characters, many of which represent morphemes and sometimes words, creating complex relationships between sound, meaning, and written form. Chinese has numerous homophones, characters with identical pronunciation but different meanings and written forms. For example, \zh{目} \textit{mù} (`eye') and \zh{木} \textit{mù} (`wood') sound identical but have distinct meanings and appearances. These homophonic relationships frequently lead to substitution errors. Additionally, many characters have similar visual structures with only minor differences, such as \zh{日} \textit{rì} (`sun') and \zh{目} \textit{mù} (`eye'), which differ by just one stroke. These properties help explain why some previous CGEC work has adopted character-level operations in automatic annotation and evaluation, since many plausible errors are naturally realized as substitutions between visually similar or similarly pronounced characters.

More challenging still are cases where characters share both phonetic and orthographic similarities. Characters like \zh{进} \textit{jìn} (`enter') and \zh{近} \textit{jìn} (`near') are not only pronounced identically but also share similar visual components, making them especially prone to confusion. These spelling-like errors are not limited to L2 learners, native speakers also frequently make such mistakes, particularly in casual handwriting or when using phonetic input methods (e.g., \textit{pinyin}-based typing, where homophonic characters may be mistakenly selected by autocomplete or input ambiguity). Given the prevalence of these errors among both L1 and L2 populations, a more comprehensive annotation system is needed to differentiate between various error sources, whether phonological, visual, or semantic in nature. However, standard error annotation frameworks developed for alphabetic languages typically lack the granularity needed to capture these Chinese-specific error patterns. For this reason, Chinese-specific annotation and evaluation cannot be treated as a straightforward extension of English GEC practice. The structure of the writing system itself affects which error types are likely to be observed, how they are grouped into categories, and how correction quality should be interpreted.

Beyond the challenges posed by the writing system, structural aspects of Chinese present additional difficulties. Unlike Indo-European languages, many of which use inflectional morphology to mark tense, number, gender, and case, Chinese has little inflectional morphology of this kind. Instead, the language features several structural particles. The versatile particle 了 \textit{le} serves multiple grammatical functions, including marking completed actions when placed after verbs, indicating change of state when positioned at sentence end, and functioning as a modal particle to express tone or attitude. Similarly, the three homophonic particles \zh{的} \textit{de}, \zh{地} \textit{de}, and \zh{得} \textit{de} serve as markers for attribution, adverbial modification, and complementation respectively. Their identical pronunciation yet distinct grammatical roles make them frequent sources of error for both L1 and L2 writers. Chinese also employs unique syntactic constructions that reorganize elements within sentences in distinctive ways. The \textit{ba}-construction (\zh{把} \textit{bǎ}) shifts the object before the verb to emphasize the action's result. The \textit{bei}-construction (\zh{被} \textit{bèi}) creates passive voice expressions following the pattern ``Object + \zh{被} \textit{bèi} + Agent + Verb'', where the semantic object becomes the syntactic subject. \mq{Beyond these specific constructions, Chinese is often described as relatively topic-prominent: topic-comment organization, in which a topic, such as the object of a verb, appears at the beginning of the sentence followed by information about that topic, is pervasive and not signaled by overt case or agreement marking. Although topicalization is not unique to Chinese, its prominence, together with the availability of null arguments and flexible word order, means that topic-comment variation interacts with annotation and evaluation in CGEC in ways that have few direct parallels in English GEC.}\label{r2-topicalization} For a comprehensive review of Chinese grammar, see \citet{ross-2024-modern}. From the perspective of CGEC, these grammatical properties are relevant because they shape how learner errors arise, how they are localized in annotation, and how fine-grained error categories may or may not be recoverable from corrected sentence pairs. They also affect how useful a correction system can be for language learning, since feedback that ignores such distinctions may improve surface acceptability without clearly reflecting the underlying grammatical problem.

Recognizing the unique challenges of CGEC, researchers have fueled progress through a series of shared tasks that continue to push the field forward. Some of these tasks have concentrated on Chinese spelling error detection and correction, such as the \textit{SIGHAN 2013--2015 Bake-off for Chinese Spelling Check} \citep{wu-etal-2013-chinese,yu-etal-2014-overview,tseng-etal-2015-introduction} and the \textit{NLP-TEA 2017} and \textit{NLPCC 2023} Shared Tasks for Chinese Spelling Check \citep{fung-etal-2017-nlptea,yin-etal-2023-overview}. Others have focused on identifying grammatical error types and their spans within sentences, including the \textit{NLP-TEA 2014--2020} Shared Tasks for Chinese Grammatical Error Diagnosis \citep{yu-lee-chang-2014-overview,lee-etal-2015-overview,lee-etal-2016-overview,rao-etal-2017-ijcnlp,rao-etal-2018-overview,rao-etal-2020-overview}. Meanwhile, broader Chinese GEC tasks have been introduced, such as the \textit{NLPCC 2018} \citep{zhao-etal-2018-nlpcc} and \textit{NLPCC 2023}\footnote{\url{http://tcci.ccf.org.cn/conference/2023/taskdata.php}} shared tasks, which target the entire GEC pipeline, including error detection, classification, and correction.

Alongside shared tasks, ongoing efforts to collect and refine CGEC datasets have continued to expand resources for system development and linguistic analysis. Many datasets are based on L2 learner texts, including corpora derived from the \textit{Hanyu Shuiping Kaoshi} (HSK, literally Chinese Proficiency Test), a standardized Chinese proficiency exam for non-native speakers. Because native speakers and L2 learners tend to produce different types of errors, some datasets, including FCGEC \citep{xu-etal-2022-fcgec} and CCTC \citep{wang-etal-2022-cctc}, specifically target L1 writing. Given the flexibility of Chinese structure and its reliance on context rather than explicit markers, datasets like YACLC \citep{wang-etal-2021-yaclc} and MuCGEC \citep{zhang-etal-2022-mucgec} incorporate multiple reference corrections to reflect the variety of acceptable edits. Collectively, these datasets provide diverse sources of error data, supporting research on both L1 and L2 errors while accommodating the unique characteristics of the language. At the same time, differences in learner population, segmentation convention, correction format, and annotation granularity make direct comparison across datasets less straightforward than benchmark results alone may suggest. One goal of the present survey is therefore to bring these design choices into a common analytical discussion, rather than treating datasets, annotation practices, and evaluation settings as isolated components.

With the increasing availability of annotated datasets, shared tasks, and advanced computational methods, Chinese GEC research has reached a stage where a comprehensive review is warranted. The field has witnessed notable achievements, from the development of large-scale learner corpora to the adoption of transformer models that improve correction accuracy. However, challenges remain, such as refining annotation standards, improving evaluation metrics tailored to Chinese linguistic properties, and adapting CGEC models for diverse real-world applications. In the structure of this survey, the main text addresses three core components of the field: datasets together with their annotation schemes (Section~\ref{sec:datasets}), evaluation in CGEC (Section~\ref{sec:evaluation}), and CGEC systems (Section~\ref{sec:methods}). These components define the main empirical and methodological dimensions along which CGEC research has developed. 
The survey concludes in Section~\ref{sec:conclusion} with a summary of the main themes of the field and a discussion of directions for future research.\footnote{Automatic grammatical error annotation for edit-based evaluation is presented in Appendix~\ref{sec:annotation}, since it constitutes an important practical component of CGEC evaluation. Because evaluation in CGEC has been shaped primarily by edit-based approaches, the main text focuses on that line of work. Appendix~\ref{sec:evaluation-gec} broadens the discussion to general GEC evaluation methods, providing the wider methodological context needed for a more complete survey. This survey aims to synthesize existing work, highlight major developments, and identify directions for future research in CGEC.}

\section{Chinese GEC datasets and error annotation}\label{sec:datasets}

Chinese grammatical error correction relies not only on the availability of parallel or annotated data, but also on how learner deviations are defined, localized, and represented in each resource. This section therefore surveys major Chinese GEC datasets together with their error annotation schemes, since the two are closely intertwined in practice. The resources reviewed here cover texts produced by both L2 learners of Chinese and L1 native speakers. In principle, this distinction matters because the distribution, linguistic nature, and pedagogical interpretation of errors may differ across these two populations. In GEC research, however, such a distinction is often not maintained consistently at the level of task formulation or evaluation, where datasets from different learner populations are frequently treated within a single correction framework.\footnote{In English GEC, datasets often distinguish between L1 and L2 writing, but GEC systems themselves have generally tended not to treat them as fundamentally separate correction tasks. In Chinese GEC, most datasets are built for L2 learners, while L1 native-speaker datasets remain relatively rare. Even so, Chinese GEC research likewise often treats these resources within a unified correction framework, although some studies have begun to examine the distinction more directly.} For this reason, a comparative review of Chinese GEC datasets must consider not only data source, scale, and correction format, but also the annotation assumptions that shape what counts as an error and how it is operationalized for training and evaluation.

\subsection{CGED}

The Chinese Grammatical Error Diagnosis (CGED) shared tasks constitute one of the longest running benchmark series for Chinese learner error processing, with major releases in 2014, 2015, 2016, 2017, 2018, and 2020 \citep{yu-lee-chang-2014-overview,lee-etal-2015-overview,lee-etal-2016-overview,rao-etal-2017-ijcnlp,rao-etal-2018-overview,rao-etal-2020-overview}. The earlier releases, especially CGED 2014 and 2015, were distributed by National Taiwan Normal University and consist of traditional Chinese learner texts drawn from the Test of Chinese as a Foreign Language (TOCFL). CGED 2016 broadened this setting by including not only traditional Chinese data from TOCFL but also simplified Chinese data from the \textit{Hanyu Shuiping Kaoshi} (HSK), thus extending the benchmark across script and testing contexts \citep{lee-etal-2016-overview}. From CGED 2017 onward, the shared task shifted primarily toward simplified Chinese data, much of it associated with Beijing Language and Culture University \citep{rao-etal-2017-ijcnlp,rao-etal-2018-overview}. Later releases, especially CGED 2018 and CGED 2020, further expanded the data sources through contributions from the Beijing Language and Culture University Advanced Innovation Center for Language Resources, the BLCU Language Monitoring and Intelligent Learning Research Group, and the Global Chinese Interlanguage Corpus (QQK) \citep{rao-etal-2018-overview,rao-etal-2020-overview}.\footnote{\url{http://yuyanziyuan.blcu.edu.cn/en/info/1050/1297.htm}}

This historical progression is important because CGED is not a single uniform resource, but a sequence of related benchmarks whose data sources, scripts, and annotation formats evolved over time. The following subsections therefore review each release in turn, with attention to both dataset composition and error annotation scheme.

\paragraph{CGED 2014, Traditional Chinese}
Figure~\ref{example-of-cged2014} illustrates the annotation format used in CGED 2014. Each learner sentence is recorded in a \texttt{<SENTENCE>} element and assigned a unique \texttt{id}, such as \texttt{A2-0018-1}. This identifier is then reused in a corresponding \texttt{<MISTAKE>} element, which links the error annotation to the original sentence. In this way, grammatical errors are annotated at the sentence level rather than through explicit character or word offsets inside the sentence. The prefix in the identifier, such as \texttt{A2}, indicates the learner proficiency level in the TOCFL framework.

The grammatical error itself is encoded by the \texttt{<TYPE>} element, which assigns one of a small set of diagnosis labels to the sentence. In CGED 2014, these labels include \texttt{Selection} for inappropriate word choice, \texttt{Missing} for omitted elements, \texttt{Disorder} for word-order errors, and \texttt{Redundant} for unnecessary material. The annotation therefore specifies the error category, but not the exact span of the error in the source sentence.

The \texttt{<CORRECTION>} element provides a fully corrected version of the sentence associated with the diagnosed error. As shown in Figure~\ref{example-of-cged2014}, the correction is given as a complete sentence rather than as a localized edit operation. CGED 2014 thus combines sentence-level error classification with sentence-level correction, without yet introducing explicit position based annotation.

\begin{figure}[ht!]
    \centering
\begin{subfigure}[b]{\textwidth}    
\centering
\scriptsize{
\begin{tabular}{l}
\verb+<ESSAY title="+{不能參加朋友\zh{找}到工作的慶祝會}\verb+">+\\
\verb+<TEXT>+\\
\verb+<SENTENCE id="A2-0018-1">+{不過我想\zh{跟你}說我\zh{真的很}對不起我不能參加\zh{你的}慶祝會}\verb+</SENTENCE>+\\
\verb+<SENTENCE id="A2-0018-2">+{我希望我不在的時候\zh{你}會快樂地慶祝慶祝}\verb+</SENTENCE>+\\
\verb+</TEXT>+\\
\verb+<MISTAKE id="A2-0018-1">+\\
\verb+<TYPE>Selection</TYPE>+\\
\verb+<CORRECTION>+{不過我想\zh{跟你}說我\zh{真的很抱歉}我不能參加\zh{你}的慶祝會}\verb+</CORRECTION>+\\
\verb+</MISTAKE>+\\
\verb+<MISTAKE id="A2-0018-2">+\\
\verb+<TYPE>Redundant</TYPE>+\\
\verb+<CORRECTION>+{我希望我不在的時候\zh{你}會快樂地慶祝}\verb+</CORRECTION>+\\
\verb+</MISTAKE>+\\
\verb+</ESSAY>+\\
\end{tabular}
}
\caption{File format of CGED 2014} \label{CGED-2014}
\end{subfigure}

\hfill

\begin{subfigure}[b]{\textwidth}    
\centering
\scriptsize{
\begin{tabular}{@{}r p{0.8\textwidth}@{}}
\footnotesize{Original sentence}     & {不過我想\zh{跟你}說我\zh{真的很}對不起我不能參加\zh{你}的慶祝會} \\
& \textit{Bùguò wǒ xiǎng gēn nǐ shuō wǒ zhēn de hěn duìbùqǐ wǒ bùnéng cānjiā nǐ de qìngzhù huì}\\
& `However, I want to tell you that I am really sorry I cannot attend your celebration.'\\
\footnotesize{Reference correction} & {不過我想\zh{跟你}說我\zh{真的很抱歉}我不能參加\zh{你}的慶祝會} \\ 
& \textit{Bùguò wǒ xiǎng gēn nǐ shuō wǒ zhēn de hěn bàoqiàn wǒ bùnéng cānjiā nǐ de qìngzhù huì}\\
& `However, I want to tell you that I am really sorry I cannot attend your celebration.'\\
\hdashline
\footnotesize{Original sentence}     & {我希望我不在的時候\zh{你}會快樂地慶祝慶祝} \\
& \textit{Wǒ xīwàng wǒ bù zài de shíhòu nǐ huì kuàilè de qìngzhù qìngzhù} \\
& `I hope that while I am not there, you will happily celebrate.'\\
\footnotesize{Reference correction} & {我希望我不在的時候\zh{你}會快樂地慶祝} \\
& \textit{Wǒ xīwàng wǒ bù zài de shíhòu nǐ huì kuàilè de qìngzhù} \\
& `I hope that while I am not there, you will happily celebrate.'\\
\end{tabular}
}    
\caption{Sentence examples from CGED 2014} \label{CGED2014-translation}
\end{subfigure}

\caption{File format and sentence examples: CGED 2014}\label{example-of-cged2014}
\end{figure}

\paragraph{CGED 2015, Traditional Chinese}

Figure~\ref{example-of-cged2015} shows that CGED 2015 moves beyond the sentence level annotation used in CGED 2014 by explicitly marking the location of each grammatical error within the sentence. The learner sentence is still recorded in a \texttt{<SENTENCE>} element with a unique \texttt{id}, but the error itself is annotated in a \texttt{<MISTAKE>} element that includes the attributes \texttt{start\_off} and \texttt{end\_off}. These two attributes specify the position of the error span using character offsets, allowing the annotation to localize the grammatical problem directly in the source sentence.

The error category is encoded in the \texttt{<TYPE>} element. In CGED 2015, the annotation labels include \texttt{Missing} for omitted material, \texttt{Permutation} for word order errors, \texttt{Addition} for redundant material, \texttt{Deletion} for omission related edits, and \texttt{Substitution} for inappropriate word choice. In the example in Figure~\ref{example-of-cged2015}, the error is labeled as \texttt{Missing}, indicating that an element is absent at the annotated position.

As in the previous release, the \texttt{<CORRECTION>} element provides the corrected sentence in full rather than encoding the correction as an atomic edit. CGED 2015 therefore combines localized error detection through character based offsets with sentence level correction, representing a clear step toward more explicit and operationalized error annotation.

\begin{figure}[ht!]
    \centering
\begin{subfigure}[b]{\textwidth}    
\centering
\scriptsize{
\begin{tabular}{l}
\verb+<DOC>+\\
\verb+<SENTENCE id="B1-0108">+{還有是一個\zh{很}好的機會家人在一起}\verb+</SENTENCE>+\\
\verb+<MISTAKE start_off="11" end_off="11">+\\
\verb+<TYPE>Missing</TYPE>+\\
\verb+<CORRECTION>+{還有是一個\zh{很}好的機會讓家人在一起}\verb+</CORRECTION>+\\
\verb+</MISTAKE>+\\
\verb+</DOC>+\\
\end{tabular}
}
\caption{File format of CGED 2015} \label{CGED-2015}
\end{subfigure}

\hfill

\begin{subfigure}[b]{\textwidth}    
\centering
\scriptsize{
\begin{tabular}{@{}r p{0.8\textwidth}@{}}\footnotesize{Original sentence}     & {還有是一個\zh{很}好的機會家人在一起} \\
& \textit{Hái yǒu shì yī gè hěn hǎo de jī huì jiā rén zài yī qǐ}\\
& `It is also a very good opportunity for the family to be together.'\\
\footnotesize{Reference correction} & {還有是一個\zh{很}好的機會讓家人在一起} \\ 
& \textit{Hái yǒu shì yī gè hěn hǎo de jī huì ràng jiā rén zài yī qǐ}\\
& `It is also a very good opportunity to bring the family together.'\\
\end{tabular}
}    \caption{Sentence examples from CGED 2015} \label{CGEC2015-translation}
\end{subfigure}

\caption{File format and sentence examples: CGED 2015} \label{example-of-cged2015}
\end{figure}

\paragraph{CGED 2016, Simplified Chinese and Traditional Chinese}

Figure~\ref{example-of-cged2016} illustrates the annotation format used in CGED 2016 for both simplified and traditional Chinese data. Each instance contains an original learner sentence in the \texttt{<TEXT>} element, identified by a unique \texttt{id}, together with a fully corrected sentence in the \texttt{<CORRECTION>} element. Grammatical errors in the source sentence are annotated through one or more \texttt{<ERROR>} elements, each of which specifies the location of an error by means of the attributes \texttt{start\_off} and \texttt{end\_off}, along with an error category encoded in the \texttt{type} attribute.

In CGED 2016, the error types are defined as \texttt{R} for redundant words, \texttt{M} for missing words, \texttt{S} for word selection errors, and \texttt{W} for word ordering errors. Compared with earlier releases, this format makes the annotation more compact and more explicitly operationalized, since each error is represented as a localized span together with a categorical diagnosis. In the example in Figure~\ref{CGED-2016}, \verb+<ERROR start_off="9" end_off="11" type="R">+ marks a span associated with a redundancy error, while other \texttt{<ERROR>} elements identify additional missing or selection related problems elsewhere in the sentence.

As in CGED 2015, the correction itself is still given as a complete corrected sentence rather than as a sequence of independently specified edit operations. CGED 2016 thus combines span based grammatical error diagnosis with full sentence correction, while adopting a more standardized symbolic label set that was also retained in later CGED releases.

\begin{figure}[ht!]
    \centering
\begin{subfigure}[b]{\textwidth}    
\centering
\scriptsize{
\begin{tabular}{l}
\verb+<DOC>+\\
\verb+<TEXT id="200405212522200051_2_2x2">+\\
\zh{例如：青少年吸烟的害处不仅是他本身还会对社会的未来发展。 因为青少年是未来社会的主要发展因原之一。}\\
\verb+</TEXT>+\\
\verb+<CORRECTION>+\\
\zh{例如：青少年吸烟不仅是对他本身还会对社会的未来发展有害。 因为青少年是未来社会的主要发展原因之一。}\\
\verb+</CORRECTION>+\\
\verb+<ERROR start_off="9" end_off="11" type="R"></ERROR>+\\
\verb+<ERROR start_off="15" end_off="15" type="M"></ERROR>+\\
\verb+<ERROR start_off="28" end_off="28" type="M"></ERROR>+\\
\verb+<ERROR start_off="44" end_off="45" type="S"></ERROR>+\\
\verb+</DOC>+\\
\end{tabular}
}
\caption{File format of CGED 2016} \label{CGED-2016}
\end{subfigure}

\hfill

\begin{subfigure}[b]{\textwidth}    
\centering
\scriptsize{
\begin{tabular}{@{}r p{0.8\textwidth}@{}}
\footnotesize{Original sentence}     & \zh{青少年吸烟的害处不仅是他本身还会对社会的未来发展。 因为青少年是未来社会的主要发展因原之一。} \\
& \textit{Qīngshàonián xīyān de hàichù bùjǐn shì tā běnshēn hái huì duì shèhuì de wèilái fāzhǎn. Yīnwèi qīngshàonián shì wèilái shèhuì de zhǔyào fāzhǎn yīnyuán zhī yī.} \\
& `The harm of teenagers smoking is not only to themselves but also to the future development of society. Because teenagers are one of the main driving forces for the future development of society.' \\
\footnotesize{Reference correction} & \zh{青少年吸烟不仅是对他本身还会对社会的未来发展有害。 因为青少年是未来社会的主要发展原因之一。} \\ 
& \textit{Qīngshàonián xīyān bùjǐn shì duì tā běnshēn hái huì duì shèhuì de wèilái fāzhǎn yǒuhài. Yīnwèi qīngshàonián shì wèilái shèhuì de zhǔyào fāzhǎn yuányīn zhī yī.} \\
& `Teenagers smoking is not only harmful to themselves but also to the future development of society. Because teenagers are one of the main reasons for the future development of society.' \\ 
\end{tabular}
}    
\caption{Sentence examples from CGEC2016} \label{CGEC2016-translation}
\end{subfigure}
\caption{File format and sentence examples: CGEC2016}\label{example-of-cged2016}
\end{figure}

\paragraph{CGED 2017 and 2018, Simplified Chinese}
CGED 2017 and 2018 continue the annotation design introduced in CGED 2016 for simplified Chinese learner data. In both releases, each instance contains an original sentence, a fully corrected sentence, and a set of localized error annotations. Errors are encoded through span based \texttt{<ERROR>} elements with \texttt{start\_off} and \texttt{end\_off} attributes, while the \texttt{type} attribute assigns one of the same diagnosis labels used in CGED 2016, namely \texttt{R} for redundant words, \texttt{M} for missing words, \texttt{S} for word selection errors, and \texttt{W} for word ordering errors. These releases therefore preserve the same basic representation of grammatical errors as localized spans paired with categorical labels, while continuing to provide full sentence corrections rather than decomposed edit operations.

\paragraph{CGED 2020--2021, Simplified Chinese}

Figure~\ref{example-of-cged2020-2021} illustrates the annotation format used in the later CGED releases for simplified Chinese learner data. As in CGED 2016, 2017, and 2018, each instance contains an original sentence in the \texttt{<TEXT>} element, identified by a unique \texttt{id}, together with a fully corrected sentence in the \texttt{<CORRECTION>} element. Grammatical errors in the source sentence are annotated through \texttt{<ERROR>} elements that specify the span of the error with \texttt{start\_off} and \texttt{end\_off}, and assign an error category through the \texttt{type} attribute.

The label inventory remains the same as in the earlier releases: \texttt{R} denotes redundant words, \texttt{M} missing words, \texttt{S} word selection errors, and \texttt{W} word ordering errors. A notable extension in CGED 2020 and 2021 is the optional \texttt{answer} attribute, which provides one or more suggested corrections for certain errors. In Figure~\ref{example-of-cged2020-2021}, for instance, \verb+<ERROR start_off="6" end_off="8" type="S" answer="+\zh{邻居们}\verb+">+ marks a word selection error spanning characters 6 to 8, and the \texttt{answer} attribute explicitly suggests replacing \zh{亲属们} \textit{qīnshǔmen} (`relatives') with \zh{邻居们} \textit{línjūmen} (`neighbors').

This format therefore preserves the span based diagnostic structure introduced in CGED 2016 while making the annotation slightly more informative through optional candidate corrections. Even so, the dataset still provides a full corrected sentence in parallel, rather than decomposing the correction into a complete sequence of edit operations.

\begin{figure}[ht!]
    \centering
\begin{subfigure}[b]{\textwidth}    
\centering
\scriptsize{
\begin{tabular}{l}
\verb+<DOC>+\\
\verb+<TEXT id="6">+\\
\zh{1973年亲属们一知道我母亲生了一个女孩时，邻居们份份的来看父亲，表示自己对他的同情。}\\
\verb+</TEXT>+\\
\verb+<CORRECTION>+\\
\zh{1973年邻居们一知道我母亲生了一个女孩时，纷纷来看父亲，表示对他的同情。}\\
\verb+</CORRECTION>+\\
\verb+<ERROR start_off="6" end_off="8" type="S" answer="+\zh{邻居们}\verb+"></ERROR>+\\
\verb+<ERROR start_off="23" end_off="25" type="R"></ERROR>+\\
\verb+<ERROR start_off="26" end_off="27" type="S" answer="+\zh{纷纷，一起，都}\verb+"></ERROR>+\\
\verb+<ERROR start_off="28" end_off="28" type="R"></ERROR>+\\
\verb+<ERROR start_off="36" end_off="37" type="R"></ERROR>+\\
\verb+</DOC>+\\
\end{tabular}
}
\caption{File format of CGED 2020-2021} \label{CGED-2020-2021}
\end{subfigure}

\hfill

\begin{subfigure}[b]{\textwidth}    
\centering
\scriptsize{
\begin{tabular}{@{}r p{0.8\textwidth}@{}}
\footnotesize{Original sentence}     & \zh{1973年亲属们一知道我母亲生了一个女孩时，邻居们份份的来看父亲，表示自己对他的同情。} \\
& \textit{1973 nián qīnshǔmen yī zhīdào wǒ mǔqīn shēng le yīgè nǚhái shí, línjūmen fènfèn de lái kàn fùqīn, biǎoshì zìjǐ duì tā de tóngqíng}. \\
& `In 1973, when my relatives found out that my mother gave birth to a girl, the neighbors came to visit my father, expressing their sympathy for him.' \\
\footnotesize{Reference correction} & \zh{1973年邻居们一知道我母亲生了一个女孩时，纷纷来看父亲，表示对他的同情。} \\ 
& \textit{1973 nián línjūmen yī zhīdào wǒ mǔqīn shēng le yīgè nǚhái shí, fēnfēn lái kàn fùqīn, biǎoshì duì tā de tóngqíng.} \\
& `In 1973, when the neighbors found out that my mother gave birth to a girl, they came to visit my father, expressing their sympathy for him.' \\ 
\end{tabular}
}    
\caption{Sentence examples from CGEC2020-2021} \label{CGEC2020-2021-translation}
\end{subfigure}

\caption{File format and sentence examples: CGEC2020-2021}\label{example-of-cged2020-2021}
\end{figure}

\paragraph{Summary of CGED}

Table~\ref{cged-since-2014} summarizes the major CGED shared task releases since 2014. These datasets constitute a long-running benchmark series for grammatical error diagnosis in Chinese learner writing. Their primary objective is not full sentence correction, but the identification and categorization of learner errors, especially with respect to error location and type. For this reason, CGED occupies an important position in the history of Chinese GEC, even though its design is more diagnosis-oriented than correction-oriented.

Viewed diachronically, CGED shows a clear evolution over time. First, the underlying learner data shifted in both script and source. The earlier releases were based mainly on traditional Chinese learner texts from TOCFL, while later releases increasingly incorporated simplified Chinese data associated with HSK and other learner resources. This change broadened the benchmark beyond a single testing context and made CGED more representative of the wider landscape of Chinese learner writing.

Second, the annotation scheme became progressively more explicit and operationalized. CGED 2014 mainly linked sentence level error categories to full sentence corrections, without directly marking the position of the error in the source sentence. CGED 2015 introduced span localization through offset based annotation, and CGED 2016 onward adopted a more compact symbolic label inventory with explicit error spans. Later releases also added optional answer fields for some errors. In this respect, CGED evolved from relatively coarse sentence linked diagnosis toward more structured and machine usable error annotation.

Third, the series maintained a persistent tension between diagnosis and correction. Although corrected sentences are often provided, the annotation design is centered on detecting and classifying errors rather than representing corrections as fully decomposed edit operations. This makes CGED highly valuable for grammatical error diagnosis, but less directly compatible with correction oriented pipelines that require stable edit extraction or multi reference target sentences.

The availability of training and test data also changed over time. CGED 2014 to 2018 generally provide both training and test sets, whereas the later releases place greater emphasis on evaluation data, with CGED 2021 providing only a test set. Overall, CGED is best understood not as a single uniform dataset, but as a sequence of related benchmarks whose data sources, scripts, and annotation assumptions changed across years. This historical development is important for CGEC research because it shaped how Chinese learner errors came to be represented, localized, and evaluated in subsequent work.

\begin{table}[ht!]
\caption{CGED since 2014}\label{cged-since-2014}
\footnotesize{
\begin{tabular}{r r | c c} \toprule 
 & & Train & Test\\ \midrule 
\textsc{cged2014} (nlptea14cfl) & A2\_CFL & \Checkmark & \Checkmark \\ 
& B1\_CFL & \Checkmark & \Checkmark \\ 
& B2\_CFL & \Checkmark & \Checkmark \\ 
& C1\_CFL & \Checkmark & \Checkmark \\  
\hdashline
\textsc{cged2015} (nlptea15cged) & NLPTEA15\_CGED & \Checkmark & \Checkmark \\ 
\hdashline
\textsc{cged2016} (nlptea16cged) & CGED16\_HSK  & \Checkmark & \Checkmark \\
& CGED16\_TOCFL  & \Checkmark & \Checkmark \\ 
\hdashline
\textsc{cged2017} & HSK & \Checkmark & \Checkmark \\ 
\hdashline
\textsc{cged2018} & CGED2018 & \Checkmark & \Checkmark \\ 
\hdashline
\textsc{cged2020} & nlptea2020 & \Checkmark & \Checkmark \\ 
\hdashline
\textsc{cged2021} & - & - & \Checkmark \\ 
\bottomrule 
\end{tabular}
} 
\end{table}

\subsection{NLPCC2018}

The NLPCC 2018 shared task on Chinese grammatical error correction introduced one of the earliest large scale benchmarks explicitly designed for CGEC rather than grammatical error diagnosis \citep{zhao-etal-2018-nlpcc}. The training data were collected from Lang-8, a language learning platform, and therefore consist of learner sentences paired with one or more corrected versions. The dataset is heterogeneous in the number of available references: 123,501 sentences (17.21\%) have no correction, 300,004 sentences (41.82\%) have only one corrected version, and some sentences have as many as 21 references. 
The test set, by contrast, contains 2,000 sentences written by foreign learners and corrected by language instructors. Unlike the training set, it was distributed only with source sentences in word segmented form, without gold target references,\label{r1-p8} because evaluation was conducted against hidden references using the MaxMatch (\texttt{M\textsuperscript{2}}) metric \citep{dahlmeier-ng-2012-better}.

Figure~\ref{example-of-NLPCC2018} illustrates the basic representation used in NLPCC2018. Unlike CGED, which explicitly marks error spans and assigns diagnosis labels, NLPCC2018 does not provide typed error annotations in the released training data. Instead, the dataset is organized as sentence aligned source sentences with zero, one, or more reference corrections. According to the task description, each training instance is represented as a tab separated record containing a sentence identifier, the number of available corrections, the original sentence, and, when available, one or more corrected sentences provided by different annotators. The guideline further notes that some input sentences may admit multiple corrections.

For example, the record \verb+1 1+ ~\zh{我在家里一个人学习中文。~ 我在家里自学中文。} indicates that the source sentence \zh{我在家里一个人学习中文。} \textit{wǒ zài jiālǐ yí gè rén xuéxí zhōngwén.} (`I study Chinese alone at home.') is associated with one reference correction, namely \zh{我在家里自学中文。} \textit{wǒ zài jiālǐ zìxué zhōngwén.} (`I study Chinese on my own at home.'). More generally, NLPCC2018 represents correction as sentence level source reference alignment rather than as an explicit inventory of labeled error spans. This design makes the resource directly usable for correction-oriented training and M$^2$-based evaluation, but it does not by itself specify an interpretable error taxonomy comparable to those used in diagnosis-oriented corpora such as CGED.

\begin{figure}[ht!]
    \centering
\begin{subfigure}[b]{\textwidth}    
\centering
\scriptsize{
\begin{tabular}{l}
\verb+1       1       +\zh{我在家里一个人学习中文。}\verb+        +\zh{我在家里自学中文。}\\
\end{tabular}
}
\caption{File format of NLPCC2018} \label{NLPCC2018}
\end{subfigure}

\hfill

\begin{subfigure}[b]{\textwidth}    
\centering
\scriptsize{
\begin{tabular}{@{}r p{0.8\textwidth}@{}}
\footnotesize{Original sentence} & \zh{我在家里一个人学习中文。} \\
& \textit{wǒ zài jiālǐ yí gè rén xuéxí zhōngwén.} \\
& `I study Chinese alone at home.' \\
\footnotesize{Reference correction} & \zh{我在家里自学中文。} \\
& \textit{wǒ zài jiālǐ zìxué zhōngwén.} \\
& `I study Chinese on my own at home.' \\
\end{tabular}
}    
\caption{Sentence examples from NLPCC2018} \label{NLPCC2018-translation}
\end{subfigure}

\caption{File format and sentence examples: NLPCC2018}\label{example-of-NLPCC2018}
\end{figure}

From the perspective of error annotation, NLPCC2018 is therefore largely implicit. As suggested in Figure~\ref{example-of-NLPCC2018}, errors are not directly represented as labeled spans or atomic edit operations in the dataset itself, but must instead be inferred from the difference between the source sentence and its correction or corrections. The annotation scheme is thus defined by sentence level parallelism rather than by an explicit inventory of error categories. In this respect, NLPCC2018 is fundamentally correction oriented rather than diagnosis oriented.

A further important characteristic of NLPCC2018 is that its test setting imposes a word segmented view of correction. The test data consist of two aligned files, one containing the original sentences and the other containing the same sentences after word segmentation with \texttt{PKUNLP}.\footnote{Since the official \texttt{PKUNLP} tool is no longer publicly available, recent studies often use the Language Technology Platform (LTP), a Chinese NLP toolkit \citep{che-etal-2010-ltp}: \url{https://github.com/HIT-SCIR/ltp}.} For example, the sentence \zh{另外，冬阴功对外国人的喜爱不断地增加。} \textit{lìngwài, dōng yīngōng duì wàiguó rén de xǐ'ài bùduàn de zēngjiā} (`Moreover, foreigners' love for Tom Yum soup continues to increase.') appears in the source file, while the segmented file provides \zh{另外$\sqcup$，$\sqcup$冬$\sqcup$阴功$\sqcup$对$\sqcup$外国人$\sqcup$的$\sqcup$喜爱$\sqcup$不断$\sqcup$地$\sqcup$增加$\sqcup$。}, where $\sqcup$ represents a space. The evaluation is then performed against word based \texttt{M$^{2}$} references constructed on the basis of these predefined segmentation boundaries.

This design has two important consequences. First, NLPCC2018 helped establish sentence parallel correction as a central format for Chinese GEC, in contrast to the diagnosis centered span annotation of CGED. Second, because evaluation depends on predefined word segmentation, the notion of an edit is no longer purely character based but mediated by a tokenization scheme. NLPCC2018 therefore marks an important transition in CGEC resource design: it shifts the task toward correction oriented benchmarking, while at the same time making segmentation dependence a central methodological issue.

\subsection{MuCGEC}

MuCGEC, short for the Multi-Reference Multi-Source Evaluation Dataset for Chinese Grammatical Error Correction, is an evaluation oriented CGEC resource built from L2 learner texts \citep{zhang-etal-2022-mucgec}.\footnote{\url{https://github.com/HillZhang1999/MuCGEC}} The term \textit{multi-reference} refers to the fact that each erroneous sentence is paired with multiple corrected versions, with the dataset providing on average more than two references per sentence. The term \textit{multi-source} indicates that these sentences are drawn from several preexisting Chinese learner datasets, including 1,996 sentences from NLPCC18 \citep{zhao-etal-2018-nlpcc}, 3,125 sentences from CGED18 and CGED20 \citep{rao-etal-2018-overview,rao-etal-2020-overview}, and 1,942 sentences from Lang-8 \citep{zhang-etal-2022-mucgec}. For example, the NLPCC18 portion in MuCGEC contains an average of 2.5 references per sentence, compared with only 1.1 references in the original NLPCC18 test set. This enriched multi-reference design is intended to make evaluation more reliable by better capturing the range of acceptable corrections.

In terms of representation, MuCGEC itself is not manually annotated with explicit error spans or hand assigned error categories in the way that some diagnosis oriented resources are. Its core format is sentence parallel: each line in the released plain text files contains a source sentence followed by one or more reference corrections, separated by tab characters. The dataset is therefore designed primarily for correction oriented evaluation rather than for the direct study of manually annotated error structure.

At the same time, MuCGEC is closely connected to automatic error annotation. To support evaluation and edit extraction, the authors introduce \texttt{ChERRANT}, a character based span level annotation framework, together with a heuristic span level treatment of word order errors \citep{hinson-etal-2020-heterogeneous}. These annotations are not part of the original manual dataset construction, but are generated automatically in order to derive edit based comparisons from sentence level source target pairs. For this reason, MuCGEC occupies an important intermediate position in Chinese GEC resource development: it is not an error annotated corpus in the strict manual sense, yet it strongly influenced how automatic annotation for Chinese corrections came to be operationalized. We return to this issue in \ref{sec:annotation}, where we discuss automatic annotation frameworks in more detail.

\subsection{YACLC}
Yet Another Chinese Learner Corpus (YACLC) is a large scale Chinese learner corpus that provides richly layered correction annotations for L2 Chinese writing \citep{wang-etal-2021-yaclc}.\footnote{\url{https://github.com/blcuicall/YACLC}} It is collaboratively developed by a consortium including Beijing Language and Culture University, Tsinghua University, Beijing Normal University, Yunnan Normal University, Northeastern University, and Shanghai University of Finance and Economics. The annotation process involves 183 annotators with backgrounds in Chinese International Education, Linguistics, and Applied Linguistics, and each sentence is reviewed by 10 annotators through a crowd-sourcing based procedure. The resulting resource is designed not only as a correction dataset, but as a multidimensional learner corpus that records sentence acceptability, grammatical correction, and fluency oriented rewriting.

Figure~\ref{example-of-yaclc} illustrates the annotation format used in YACLC. Each instance contains metadata such as \texttt{sentence\_id}, \texttt{article\_id}, \texttt{article\_name}, and \texttt{total\_annotators}, together with the original learner sentence in the \texttt{sentence\_text} field. The core annotation is stored in the \texttt{sentence\_annos} field, which contains a list of corrected versions associated with the sentence. Each correction entry records whether it is grammatical or fluency oriented through the binary field \texttt{is\_grammatical}, the corrected sentence itself in \texttt{correction}, the number of edits in \texttt{edits\_count}, and the number of annotators who produced that correction in \texttt{annotator\_count}.

The annotation scheme therefore distinguishes two kinds of revision. When \texttt{is\_grammatical} is set to 1, the correction is intended to satisfy grammatical well formedness under a principle of minimal modification. These annotations aim to repair the learner sentence while preserving as much of its original form as possible. When \texttt{is\_grammatical} is set to 0, the correction is instead fluency oriented, meaning that it seeks a more natural or native like expression even when this requires more substantial rewriting. As shown in Figure~\ref{example-of-yaclc}, a single source sentence may thus receive several grammatical corrections and one or more fluency based rewritings, each supported by a different number of annotators.

From the perspective of error annotation, YACLC differs from datasets that explicitly represent localized edits or categorical error labels. Its annotation is sentence based and multi-reference, with the correction type encoded at the sentence level rather than through token or span level diagnosis. Errors are therefore not directly annotated as typed spans. Instead, they are implicitly represented through the contrast between the source sentence and multiple corrected versions, with the additional distinction between grammatical acceptability and fluency. This makes YACLC particularly valuable for studying correction variation, annotator agreement, and the boundary between grammatical correction and stylistic improvement.

At present, the corpus provides 1,000 corrected sentences for validation and 1,000 source sentences for testing. Evaluation is reported using Levenshtein distance and \texttt{m2scorer}. More broadly, YACLC is important because it expands the notion of CGEC annotation beyond single reference grammatical repair. By recording multiple human corrections, annotator distributions, and both grammatical and fluency oriented revisions, it offers a richer view of learner sentence revision than datasets built only around one gold correction.

\begin{figure}[ht!]
\centering
\scriptsize
\begin{subfigure}[b]{0.95\textwidth} 
\centering
\begin{tabular}{l}
\verb+"sentence_id": 4308, // sentence id +\\
\verb+"sentence_text": "+\zh{我只可以是坐飞机去的，因为巴西离英国到远极了。}\verb+", // original sentence +\\
\verb+"article_id": 7267, // article id where the original sentence is from+\\
\verb+"article_name": "+\zh{我放假的打算}\verb+", // title of the article+\\
\verb+"total_annotators": 10, // number of annotators who contributed+ \\
\verb+"sentence_annos": [ // multidimentional annotation info+ \\
\verb+{+ \\
\verb+  "is_grammatical": 1, // 1 refers to grammatical correction and 0 refers to fluency correction+  \\
\verb+  "correction": "+\zh{我只能坐飞机去，因为巴西离英国远极了。}\verb+", // corrected sentence+ \\
\verb+  "edits_count": 3, // number of corrections+ \\
\verb+  "annotator_count": 6 // number of annotators corrected the original sentence into this version+ \\
\verb+}, +\\
\verb+{ +\\
\verb+"is_grammatical": 1, +\\
\verb+"correction": "+\zh{我只能是坐飞机去的，因为巴西离英国远极了。}\verb+", +\\
\verb+"edits_count": 2,+ \\
\verb+"annotator_count": 1+ \\
\verb+}, +\\
\verb+{ +\\
\verb+"is_grammatical": 1, +\\
\verb+"correction": "+\zh{我只可以坐飞机去，因为巴西离英国远极了。}\verb+", +\\
\verb+"edits_count": 3, +\\
\verb+"annotator_count": 2 +\\
\verb+},+\\
\verb+{ +\\
\verb+"is_grammatical": 0, +\\
\verb+"correction": "+\zh{我只能坐飞机去，因为巴西离英国太远了。}\verb+", +\\
\verb+"edits_count": 6, +\\
\verb+"annotator_count": 2 +\\
\verb+} +\\
\verb+] +\\ 
\end{tabular}
\caption{File format of YACLC}
\label{YACLC-2014}
\end{subfigure}

\hfill

\begin{subfigure}[b]{\textwidth}    
\centering
\scriptsize{
\begin{tabular}{@{}r p{0.8\textwidth}@{}}
\footnotesize{Original sentence}     & \zh{我只可以是坐飞机去的，因为巴西离英国到远极了。} \\
& \textit{Wǒ zhǐ kěyǐ shì zuò fēijī qù de, yīnwèi Bāxī lí Yīngguó dào yuǎn jí le}\\
& `I could only go by plane because Brazil is extremely far from the UK.'\\
\footnotesize{Reference correction 1} & \zh{我只能坐飞机去，因为巴西离英国远极了。} \\ 
& \textit{Wǒ zhǐ néng zuò fēijī qù, yīnwèi Bāxī lí Yīngguó yuǎn jí le}\\
& `I can only take a plane because Brazil is extremely far from the UK.'\\
\footnotesize{Reference correction 2} & \zh{我只能是坐飞机去的，因为巴西离英国远极了。} \\ 
& \textit{Wǒ zhǐ néng shì zuò fēijī qù de, yīnwèi Bāxī lí Yīngguó yuǎn jí le.}\\
& `I can only go by plane because Brazil is extremely far from the UK.'\\
\footnotesize{Reference correction 3} & \zh{我只可以坐飞机去，因为巴西离英国远极了。} \\ 
& \textit{Wǒ zhǐ kěyǐ zuò fēijī qù, yīnwèi Bāxī lí Yīngguó yuǎn jí le}\\
& `I can only take a plane because Brazil is extremely far from the UK.'\\
\footnotesize{Reference correction 4} & \zh{我只能坐飞机去，因为巴西离英国太远了。} \\ 
& \textit{Wǒ zhǐ néng zuò fēijī qù, yīnwèi Bāxī lí Yīngguó tài yuǎn le}\\
& `I can only take a plane because Brazil is too far from the UK.'\\
\end{tabular}
}
\caption{Sentence examples from YACLC} \label{YACLC-translation}
\end{subfigure}
\caption{File format and sentence examples: YACLC}\label{example-of-yaclc}
\end{figure}

\subsection{FlaCGEC}

The Chinese Grammatical Error Correction Dataset with Fine-grained Linguistic Annotation (FlaCGEC) provides 10,804 training sentences, 1,334 development sentences, and 1,325 test sentences, with corrections represented in an \texttt{M2}-style format \citep{du-etal-2023-flacgec}.\footnote{\url{https://github.com/hyDududu/FlaCGEC}} Its target sentences are collected mainly from HSK reading materials and grammar-standard examples, and erroneous source sentences are then generated through rule-based edits and manually filtered. FlaCGEC is therefore best understood as a synthetic, grammar-schema-driven benchmark with a L2-oriented linguistic inventory, rather than as a learner-authored L2 corpus. System performance on the test set is reported using $F_{0.5}$ over \texttt{M2}-based precision and recall.\label{r1-p10a}

Figure~\ref{example-of-FlaCGEC} illustrates the annotation format used in FlaCGEC. Each instance contains a \texttt{source} sentence, a corrected \texttt{target} sentence, an \texttt{operation} field that stores a structured internal representation of the edits, and an \texttt{annotation} field that linearizes these edits into a semicolon separated string. In the \texttt{annotation} field, each edit is encoded by a start and end character offset, an edit operation, a fine-grained error type, and a correction string. Thus, unlike sentence pair datasets in which errors must be inferred indirectly from source target comparison, FlaCGEC makes the error annotation explicit at the level of localized edits.

In Figure~\ref{example-of-FlaCGEC}, for example, the edit \texttt{7 8|||S-\zh{频率、重复副词}|||\zh{再}} indicates that characters 7 and 8 in the source sentence, namely \zh{再三} \textit{zài sān} (`repeatedly'), are replaced by \zh{再} \textit{zài} (`again'). The operation is \texttt{S}, that is, substitution, and the error is further classified under the fine-grained category \zh{频率、重复副词} (`frequency/repetitive adverb'). The second edit in the same example applies the same annotation logic to \zh{往往} \textit{wǎngwǎng} (`often'), which is corrected to \zh{总是} \textit{zǒngshì} (`always').\label{r1-p10b}

This representation is noteworthy because it combines two layers of annotation at once. On the one hand, it preserves the explicit source target correction relation familiar from GEC datasets. On the other hand, it enriches that relation with localized edit spans and linguistically more specific error labels. FlaCGEC therefore goes beyond minimal edit extraction by making the grammatical interpretation of each edit part of the released annotation itself. In this respect, its annotation scheme is more informative than plain \texttt{m2} style correction alone, since it not only identifies where a correction occurs, but also records what kind of grammatical phenomenon the edit is intended to capture.

FlaCGEC creates erroneous source sentences from grammatical target sentences through three rule-based operations: \textit{removing words} (\texttt{M}), which removes words associated with particular grammar points and thereby produces missing-component errors; \textit{substituting words} (\texttt{S}), which replaces a word with another word of the same grammatical type and may create inappropriate collocations; and \textit{reordering words} (\texttt{W}), which permutes words and produces word-order errors. In correction annotation, these operations are represented as localized edits from the erroneous source sentence back to the target sentence. Because the synthetic generation process does not insert extra material into otherwise correct target sentences, redundant or unnecessary-token errors are not a primary generated error class in FlaCGEC.\label{r1-p11a}

From the perspective of error annotation, this design has both strengths and limitations. Its main strength lies in its explicit and fine-grained representation of localized edits, which makes the dataset attractive for analysis as well as evaluation. At the same time, because the errors are synthetically generated through a restricted set of operations, the annotation scheme reflects the logic of the generation process itself. In other words, FlaCGEC offers a highly structured and linguistically interpretable error representation, but one whose empirical coverage is shaped by the controlled inventory of edits used to create the data.

\begin{figure}[ht!]
    \centering
\begin{subfigure}[b]{\textwidth}    
\centering
\footnotesize{
\begin{tabular}{l}
\verb+"3": {+\\
\verb+  "source": +\zh{"可是后来，他们再三来镇上，西瓜往往第一个卖完。"}\verb+,+ \\
\verb+  "target": +\zh{"可是后来，他们再来镇上，西瓜总是第一个卖完。"}\verb+,+\\
\verb+  "operation": "[+\\
\verb+    [(7, 8, '+\zh{再三}\verb+'), '+\zh{频率、重复副词}\verb+', 'S', ('null', 'null', '+\zh{再}\verb+')], +\\
\verb+    [(15, 16, '+\zh{往往}\verb+'), '+\zh{频率、重复副词}\verb+', 'S', ('null', 'null', '+\zh{总是}\verb+')]+\\
\verb+   ]",+\\
\verb+  "annotation": "7 8|||S-+\zh{频率、重复副词}\verb+|||+\zh{再}\verb+;15 16|||S-+\zh{频率、重复副词}\verb+|||+\zh{总是}\verb+"+\\
\verb+},+\\
\end{tabular}
}
\caption{File format of FlaCGEC} \label{FlaCGEC-json}
\end{subfigure}

\hfill

\begin{subfigure}[b]{\textwidth}    
\centering
\scriptsize{
\begin{tabular}{@{}r p{0.8\textwidth}@{}}
\footnotesize{Original sentence}     & \zh{可是后来，他们再三来镇上，西瓜往往第一个卖完。} \\
& \textit{kě shì hòu lái, tā men zài sān lái zhèn shàng, xī guā wǎng wǎng dì yī gè mài wán.} \\
& `However, later on, when they came to the town again and again, the watermelons were often the first to sell out.' \\
\footnotesize{Reference correction} & \zh{可是后来，他们再来镇上，西瓜总是第一个卖完。} \\ 
& \textit{kě shì hòu lái, tā men zài lái zhèn shàng, xī guā zǒng shì dì yī gè mài wán.} \\
& `However, later on, when they came to the town again, the watermelons were always the first to sell out.' \\ 
\end{tabular}
}    
\caption{Sentence examples from FlaCGEC} \label{FlaCGEC-translation}
\end{subfigure}
\caption{File format and sentence examples: FlaCGEC}\label{example-of-FlaCGEC}
\end{figure}

\subsection{FCGEC}

Fine-Grained Corpus for Chinese Grammatical Error Correction (FCGEC) is a Chinese GEC dataset designed specifically for native speakers, and therefore occupies an important place among L1 Chinese correction resources \citep{xu-etal-2022-fcgec}.\footnote{\url{https://github.com/xlxwalex/FCGEC}} Whereas much earlier CGEC work focused on L2 learner writing, FCGEC addresses grammatical errors that arise in native Chinese usage, including school examination writing and edited web text. The dataset is derived from examinations written by native students from elementary school to high school, together with content collected from news aggregation websites. After removing duplicate and incomplete entries, the dataset contains 35,354 training sentences, of which 16,224 (45.89\%) are error-free, as well as 2,000 validation sentences and 3,000 test sentences.

Figure~\ref{example-of-fcgec} illustrates the annotation format of FCGEC. Each instance is stored as a JSON style record indexed by a unique identifier. The field \texttt{sentence} contains the original sentence, \texttt{error\_flag} indicates whether the sentence is judged grammatical or ungrammatical, and \texttt{error\_type} assigns a sentence level error category. The \texttt{operation} field then specifies the edits required to derive the corrected sentence. In this sense, FCGEC combines sentence level diagnosis with explicit edit based correction instructions in a single representation.

The annotation scheme is organized around three related tasks. The first is \textit{error detection}, a binary classification task that determines whether a sentence contains an error. The second is \textit{type identification}, which assigns one of several sentence level categories to an erroneous sentence: Incorrect Word Order (\texttt{IWO}), Incorrect Word Collocation (\texttt{IWC}), Component Missing (\texttt{CM}), Component Redundancy (\texttt{CR}), Structure Confusion (\texttt{SC}), Illogical (\texttt{ILL}), and Ambiguity (\texttt{AM}). If no error is present, the category is marked with \texttt{*}. The third is \textit{error correction operation}, which encodes how the sentence should be revised through concrete edit actions.

A distinctive feature of FCGEC is that correction is represented not only by a target sentence, but also by an explicit sequence of structured operations. These operations include actions such as \texttt{Delete}, \texttt{Insert}, \texttt{Modify}, and \texttt{Switch}, together with positional information and, where needed, the replacement content. In Figure~\ref{example-of-fcgec}, for example, the annotation specifies deletion of the characters at positions 21 to 23 and insertion of new material at position 31. The insertion is labeled with the tag \texttt{INS\_6}, indicating an inserted string of length six, and the field \texttt{label} stores the actual inserted content, namely \zh{对他们的家属}. Because all edit positions are defined with respect to the original sentence, the order in which the operations are listed does not affect their interpretation.

From the perspective of error annotation, FCGEC is more explicit than datasets such as NLPCC2018 or MuCGEC, which primarily represent correction through source target sentence pairs. At the same time, its annotation remains more correction oriented than diagnosis oriented, since the main representational goal is not to localize every error with an independent span label taxonomy, but to provide a usable structured edit sequence for revision. This makes FCGEC particularly valuable as an L1 Chinese GEC resource: it captures native speaker errors, provides interpretable sentence level categories, and encodes correction in an operational form that is directly usable for both modeling and analysis.

\begin{figure}[ht!]
    \centering
\begin{subfigure}[b]{\textwidth}    
\centering
\footnotesize{
\begin{tabular}{l}
\verb+"ea28d89d9ba25ff2118087259f581dc8": {+\\
\verb+    "sentence": "+\zh{中央政法委书记罗干同志对因公殉职的公安干警及家属表示崇高的敬意并致以亲切的慰问。}\verb+",+\\
\verb+    "error_flag": 1,+\\
\verb+    "error_type": "ILL",+\\
\verb+    "operation": "[+\\
\verb+        {"Delete":[21,22,23],+\\
\verb+         "Insert"[{"pos":31,"tag":"INS_6","label":"+\zh{对他们的家属}\verb+"}]}]",+\\
\verb+    "version": "FCGEC EMNLP 2022"+\\
\verb+}+\\
\end{tabular}
}
\caption{File format of FCGEC} \label{fcgec-json}
\end{subfigure}

\hfill

\begin{subfigure}[b]{\textwidth}    
\centering
\scriptsize{
\begin{tabular}{@{}r p{0.8\textwidth}@{}}
\footnotesize{Original sentence}     & \zh{中央政法委书记罗干同志对因公殉职的公安干警及家属表示崇高的敬意并致以亲切的慰问。} \\
& \textit{Zhōngyāng Zhèngfǎwěi Shūjì Luó Gàn tóngzhì duì yīngōng xùnzhi de gōng'ān gànjǐng jí jiāshǔ biǎoshì chónggāo de jìngyì bìng zhìyǐ qīnqiè de wèiwèn.} \\
\footnotesize{Reference correction} & \zh{中央政法委书记罗干同志对因公殉职的公安干警表示崇高的敬意并对他们的家属致以亲切的慰问。} \\ 
& \textit{Zhōngyāng Zhèngfǎwěi Shūjì Luó Gàn tóngzhì duì yīngōng xùnzhi de gōng'ān gànjǐng biǎoshì chónggāo de jìngyì bìng duì tāmen de jiāshǔ zhìyǐ qīnqiè de wèiwèn.} \\
& `Comrade Luo Gan, Secretary of the Central Political and Legal Affairs Commission, expressed high respect for the public security officers who died in the line of duty and extended sincere condolences to their families.'\\
\end{tabular}
}    
\caption{Sentence examples from FCGEC} \label{FCGEC-translation}
\end{subfigure}
\caption{File format and sentence examples: FCGEC}\label{example-of-fcgec}
\end{figure}

\subsection{CCTC}

Most earlier Chinese GEC resources are centered on single sentences written by L2 learners. CCTC, short for Cross-Sentence Chinese Text Correction, addresses a different setting: document-level correction for texts written by native speakers, that is, an L1 Chinese correction task \citep{wang-etal-2022-cctc}.\footnote{\url{https://github.com/destwang/CTCResources}} This distinction matters because native-speaker errors often differ from learner errors in both distribution and interpretation. CCTC is built from raw texts collected from WuDaoCorpora, which mainly contains news articles, blog posts, and popular science writing by native speakers, and the data are manually annotated. The released resource contains 637 training documents and 644 test documents, each consisting of a title and a fully segmented essay.\label{r1-p12} In total, the dataset includes 1,500 texts, 30,811 sentences, and more than 1,000,000 Chinese characters. The annotated errors fall into four broad categories: spelling errors, redundant words, missing words, and word-order errors. The training set contains 12,689 sentences, while the test portion is divided into CCTC-W, a broader test set with 14,338 sentences, and CCTC-H, a higher-quality subset with 3,784 sentences.

Figure~\ref{example-of-CCTC} illustrates the annotation format used in CCTC. Each document is represented as a JSON-like structure containing a list of \texttt{sentences}, a parallel list of \texttt{corrections}, and a document identifier. The correction annotation is therefore aligned not only with individual sentences, but also with their position inside a larger document. Within the \texttt{corrections} field, each sentence is associated with a list of localized edit operations. These edits are encoded as compact tuples containing a position index, an operation label, the source material involved in the error, and the corresponding correction. In this way, CCTC combines document-level organization with sentence-local edit annotation.

As shown in Figure~\ref{CCTC-2014}, the annotation \verb+[31, "R", "+\zh{的}\verb+", ""]+ indicates that the character \zh{的} \textit{de} at position 31 should be deleted\footnote{{This edit is fully resolvable within its host sentence and is representative of the annotation format rather than of cross-sentence phenomena.}}. More generally, the annotation scheme represents corrections as explicit edit operations rather than only as whole corrected sentences. This makes the error representation more operational than in sentence-pair datasets such as NLPCC2018 or YACLC, where the error must be inferred from source-target contrast. At the same time, unlike resources with elaborate fine-grained grammatical labels, CCTC keeps the edit representation relatively compact and action oriented.

{The role of cross-sentence context in CCTC is reported by \citet{wang-etal-2022-cctc} on the basis of a manual analysis of 100 errors: information beyond the host sentence is necessary for about 11\% of errors and helpful for a further 38\%. Their illustrative case is the correction of an isolated \zh{蜘蛛} {(`spider')} to \zh{红蜘蛛} {(`red spider')}, which becomes recoverable only once earlier sentences in the same document are taken into account. Errors of this kind cannot be reconstructed from the local edit alone, and the released resource therefore retains documents intact rather than splitting them into independent sentences, even though the individual edit annotations themselves are sentence-local.\label{r2-cross}}

From the perspective of error annotation, CCTC is notable in two respects. First, it is an L1 Chinese correction dataset, which remains relatively rare in the CGEC literature. Second, it extends the unit of release from isolated sentences to document-level texts, so that the surrounding discourse is preserved around each edit. This makes CCTC a complement to L2 sentence-level resources, broadening Chinese GEC toward native writing and longer textual units, while the proportion of edits that strictly require cross-sentence information remains a minority of the total.

\begin{figure}[ht!]
\centering
\begin{subfigure}[b]{\textwidth}    
\centering
\scriptsize{
\begin{tabular}{l}
\verb+{"sentences": +\\  
\verb+[ [...], +\\  
\verb+  [..., "+\zh{一是为了避免不必要麻烦,二是一旦被认出来难免需要签名或者合影的,会耽误明星自己的时间。}\verb+", ...]],+\\  
\verb+"corrections": +\\  
\verb+[ [...], +\\  
\verb+  [..., [[31, "R", "+\zh{的}\verb+", ""]], ...]], +\\  
\verb+"doc_id": "11851"}+\\  
\end{tabular}
}
\caption{File format of CCTC} \label{CCTC-2014}
\end{subfigure}

\hfill

\begin{subfigure}[b]{\textwidth}    
\centering
\scriptsize{
\begin{tabular}{@{}r p{0.8\textwidth}@{}}
\footnotesize{Original sentence}    & 一是为了避免不必要麻烦,二是一旦被认出来难免需要签名或者合影的,会耽误明星自己的时间。\\
\footnotesize{Reference correction} & 一是为了避免不必要麻烦,二是一旦被认出来难免需要签名或者合影,会耽误明星自己的时间。 \\
& \textit{Yī shì wèile bìmiǎn bú bìyào máfan, èr shì yídàn bèi rèn chūlái nánmiǎn xūyào qiānmíng huòzhě héyǐng, huì dānwù míngxīng zìjǐ de shíjiān.}\\
& `First, it is to avoid unnecessary trouble; second, once recognized, it is inevitable that signing autographs or taking photos would delay the celebrity's own time.'\\
\end{tabular}
}
\caption{Sentence examples from CCTC} \label{CCTC-translation}
\end{subfigure}
\caption{File format and sentence examples: CCTC}\label{example-of-CCTC}
\end{figure}

\subsection{NLPCC2023 (NaCGEC)} \label{NaCGEC-section}
After several years without a Chinese grammatical error correction track, NLPCC 2023 reintroduced the task with a new emphasis on grammatical errors in native Chinese writing, that is, an L1 Chinese GEC setting \citep{ma-etal-2022-linguistic}.\footnote{\url{https://github.com/masr2000/NaCGEC}} The shared task draws on a mixture of existing Chinese correction resources for training, including Lang-8, HSK, CGED, MuCGEC, YACLC, and CTC2021, while its own development and test data target native-speaker errors more directly. The released data include 500 validation sentence pairs, among which 87 instances have more than two references, and 5,869 test sentence pairs. These sentences are collected from secondary school and university entrance examinations, government recruitment examinations, and various Chinese news websites, while the erroneous versions are generated automatically.\label{r1-p13}

Figure~\ref{example-of-NaCGEC} illustrates the annotation format used in NaCGEC. The dataset is distributed in two related forms. One is a parallel corpus style format in which each record contains a sentence identifier, the original erroneous sentence, and one or more corrected target sentences separated by tabs. The other is a JSON format in which each instance contains at least a \texttt{source} sentence, a list of \texttt{target} corrections, and an \texttt{error\_type} field. Unlike span based annotation schemes, this format does not specify the exact location of the error in the source sentence. Instead, the sentence is assigned a single sentence level error category, and the correction is represented through one or more full target sentences. If a sentence is judged correct, the \texttt{error\_type} field is labeled as \zh{正确} \textit{zhèngquè} (`correct'), and the \texttt{target} field simply reproduces the source sentence.

The annotation scheme is based on six linguistically motivated error categories proposed for constructing the dataset \citep{ma-etal-2022-linguistic}. These are Structural Confusion (\zh{结构混乱} \textit{jiégòu hùnluàn}), Improper Logicality (\zh{不合逻辑} \textit{bùhé luójí}), Missing Component (\zh{成分残缺} \textit{chéngfèn cánquē}), Redundant Component (\zh{成分冗余} \textit{chéngfèn rǒngyú}), Improper Collocation (\zh{搭配不当} \textit{dāpèi bùdàng}), and Improper Word Order (\zh{语序不当} \textit{yǔxù bùdàng}). As shown in Figure~\ref{example-of-NaCGEC}, the example sentence is labeled as \zh{搭配不当} \textit{dāpèi bùdàng} (`improper collocation'), and the correction revises the collocational pairing between verbs and nominal complements. The annotation therefore links each sentence to a linguistically interpretable error type, but it does so at the sentence level rather than through localized edit spans.

From the perspective of error annotation, NaCGEC occupies an interesting middle position. Compared with sentence pair datasets such as NLPCC2018, it provides more explicit grammatical information by assigning an error type to each instance. Compared with datasets such as FCGEC or FlaCGEC, however, it remains relatively coarse grained, since it does not localize the erroneous span or decompose the correction into structured edit operations. This makes the resource easier to use for sentence level classification and correction, but less suitable for detailed edit based analysis.

A further important aspect of NaCGEC is that its annotation scheme is closely tied to the way the data are constructed. Because the erroneous sentences are generated automatically from linguistically motivated rules, the six way error taxonomy is not merely descriptive but also generative: it defines the space of sentence level error types that the dataset is designed to contain. In this respect, NaCGEC is valuable as an L1 Chinese GEC benchmark with an interpretable grammatical taxonomy, but its coverage is shaped by the rule based generation process that underlies the resource.

\begin{figure}[ht!]
\centering
\begin{subfigure}[b]{\textwidth}    
\centering
\scriptsize{
\begin{tabular}{l}
\verb+{+\\ 
\verb+   "source": "+\zh{...不断提升城市功能，完善城市品质，优化人居环境。}\verb+",+\\ 
\verb+   "target": [+\\ 
\verb+     "+\zh{...不断完善城市功能，提升城市品质，优化人居环境。}\verb+"+\\ 
\verb+   ],+\\ 
\verb+   "error_type": "+\zh{搭配不当}\verb+"+\\ 
\verb+ },+\\ 
\end{tabular}
}
\caption{File format of NaCGEC} \label{NaCGEC-2014}
\end{subfigure}

\hfill

\begin{subfigure}[b]{\textwidth}    
\centering
\scriptsize{
\begin{tabular}{@{}r p{0.8\textwidth}@{}}
\footnotesize{Original sentence}     & \zh{...不断提升城市功能，完善城市品质，优化人居环境。} \\
& \textit{...bù duàn tí shēng chéng shì gōng néng, wán shàn chéng shì pǐn zhì, yōu huà rén jū huán jìng.} \\
& `...continuously enhance urban functions, improve city quality, and optimize the living environment.'\\
\footnotesize{Reference correction} & \zh{...不断完善城市功能，提升城市品质，优化人居环境。} \\ 
& \textit{...bù duàn wán shàn chéng shì gōng néng, tí shēng chéng shì pǐn zhì, yōu huà rén jū huán jìng.}\\
& `...continuously improve urban functions, enhance city quality, and optimize the living environment.'\\
\end{tabular}
}    
\caption{Sentence examples from NaCGEC} \label{NaCGEC-translation}
\end{subfigure}
\caption{File format and sentence examples: NaCGEC}\label{example-of-NaCGEC}
\end{figure}

\subsection{NaSGEC}\label{NaSGEC-section}

NaSGEC is a multi-domain Chinese grammatical error correction dataset built from texts written by native Chinese speakers \citep{zhang-etal-2023-nasgec}.\footnote{Project and models: \url{https://github.com/HillZhang1999/NaSGEC}; dataset release: \url{https://github.com/SUDA-LA/CGECData/tree/main/NaSGEC}.} It should not be confused with NaCGEC in Section~\ref{NaCGEC-section}. Although both resources target L1 Chinese writing, NaSGEC was designed primarily to support multi-domain evaluation and domain-transfer analysis.

NaSGEC contains 12,500 sentences from three native-speaker domains: 4,000 sentences from WeChat public-account articles (\textsc{Media}), 1,500 sentences from computer-science undergraduate theses (\textsc{Thesis}), and 7,000 sentences from ungrammatical sentence-judgment questions in Chinese examinations (\textsc{Exam}). The paper reports 8,504 erroneous sentences overall, corresponding to 68.0\% of the dataset, and an average of 1.6 references per sentence. Because grammatical errors are sparse in the \textsc{Media} and \textsc{Thesis} sources, the authors first use several competitive CGEC models to identify likely erroneous candidates before manual annotation. These subsets are therefore best understood as error-enriched samples from native writing, rather than random samples of their source domains.

The annotation procedure follows a direct rewriting paradigm adapted from the MuCGEC guidelines. Each sentence is assigned to two annotators, who independently rewrite the whole sentence into a grammatical and fluent version while preserving its intended meaning. An expert reviewer then checks the two submissions in a double-blind manner, rejects incorrect corrections, and may add valid references missed by the annotators. As a result, NaSGEC is mainly a sentence-level, multi-reference correction dataset: errors are represented through source--reference contrasts rather than through released localized spans or fine-grained grammatical labels. For analysis, the NaSGEC paper extracts span-level edits with MuCGEC/\texttt{ChERRANT}.

The data are organized by domain, with separate folders for \textsc{Media}, \textsc{Thesis}, and \textsc{Exam}. The training and development files use a tab-separated \texttt{.para} format, where each record contains an instance identifier, the source sentence, and one or more reference corrections. The files also include two special annotations: \zh{歧义句} \textit{qíyì jù} (`ambiguous sentence'), for which no reference correction is provided, and \zh{没有错误} \textit{méiyǒu cuòwù} (`no error'), for which no correction is required. The test release separates system input from references: \texttt{.test.input} files contain source sentences, while \texttt{.test.m2} files provide the corresponding \texttt{M\textsuperscript{2}}/\texttt{ChERRANT} edit representation for evaluation. This release format is therefore compatible with sentence-level rewriting and edit-based scoring, but it should not be interpreted as providing primary human-annotated span labels or grammatical categories for every error.

From the perspective of dataset design, NaSGEC contributes a complementary form of L1 Chinese GEC data. Compared with CCTC, it keeps the sentence as the main correction unit but covers multiple native-speaker domains. Compared with FCGEC and NaCGEC, it places less emphasis on sentence-level error taxonomy and more emphasis on multi-reference correction and cross-domain variation. This makes it especially useful for evaluating how systems behave when the training and test domains differ, while also requiring caution when results are compared with single-domain or taxonomy-driven native-speaker benchmarks.

\subsection{CCL CEFE}

The CCL23-Eval Task 8, also referred to as Chinese Essay Fluency Evaluation (CEFE), is designed to identify and correct sentence level problems that affect the readability and coherence of essays written by Chinese primary and secondary school students \citep{shen-etal-2023-overview}. Among its subtracks, the \textit{Error Sentence Rewriting} task, illustrated in Figure~\ref{example-of-CEFE}, is the component most directly aligned with grammatical error correction.\footnote{\url{https://github.com/cubenlp/2023CCL_CEFE/tree/main/track3}} Its goal is to rewrite erroneous sentences through minimal modification while preserving the original meaning as much as possible.

Figure~\ref{example-of-CEFE} shows that the annotation format for this task is comparatively simple. Each instance is stored as a JSON style record containing a sentence identifier in \texttt{sent\_id}, the original sentence in \texttt{sent}, and a corrected sentence in \texttt{revisedSent}. The dataset therefore represents correction through direct source target pairing, without additional fields for localized error spans, edit operations, or categorical error labels.

From the perspective of error annotation, CEFE is thus entirely sentence based. The annotation scheme does not explicitly indicate where the error occurs in the source sentence or what type of grammatical problem is involved. Instead, the location and nature of the error must be inferred from the contrast between the original sentence and its revised form. In the example in Figure~\ref{example-of-CEFE}, the phrase \zh{原因} \textit{yuányīn} (`reason') is removed in the corrected version, yielding a more natural and grammatical sentence. However, this change is not represented as an explicit deletion operation in the annotation itself; it is recoverable only through comparison of the two sentences.

This design makes CEFE easy to use as a sentence rewriting benchmark, but relatively limited as an error annotation resource in the narrow sense. Compared with datasets such as FCGEC, FlaCGEC, or CCTC, CEFE does not provide structured edit representations or interpretable error categories. Its annotation is instead optimized for end to end rewriting. At the same time, this simplicity aligns well with the task objective: the dataset is less concerned with diagnosing the internal structure of an error than with producing a minimally revised, semantically faithful corrected sentence. CEFE therefore represents a lightweight correction oriented annotation scheme, positioned closer to sentence level rewriting than to fine-grained grammatical diagnosis.

\begin{figure}[ht!]
\centering
\begin{subfigure}[b]{\textwidth}    
\centering
\scriptsize{
\begin{tabular}{l} 
\verb+{+\\
\verb+    "sent_id":"3202",+\\
\verb+    "sent":"+\zh{上周五，我就在学校用刀切破了同学的手，原因全怪我毛手毛脚，大大咧咧，用刀太快。}\verb+",+\\
\verb+    "revisedSent":"+\zh{上周五，我就在学校用刀切破了同学的手，全怪我毛手毛脚，大大咧咧，用刀太快。}\verb+}"+\\
\verb+}+\\
\end{tabular}
}
\caption{File format of CEFE (Track 3)} \label{CEFE-file}
\end{subfigure}

\hfill

\begin{subfigure}[b]{\textwidth}    
\centering
\scriptsize{
\begin{tabular}{@{}r p{0.8\textwidth}@{}}
\footnotesize{Original sentence} & \zh{上周五，我就在学校用刀切破了同学的手，原因全怪我毛手毛脚，大大咧咧，用刀太快。} \\
& \textit{shàng zhōu wǔ, wǒ jiù zài xué xiào yòng dāo qiē pò le tóng xué de shǒu, yuán yīn quán guài wǒ máo shǒu máo jiǎo, dà dà liē liē, yòng dāo tài kuài.} \\
& `Last Friday, I accidentally cut a classmate's hand with a knife at school. The reason was entirely my carelessness and recklessness, using the knife too quickly.' \\
\footnotesize{Reference correction} & \zh{上周五，我就在学校用刀切破了同学的手，全怪我毛手毛脚，大大咧咧，用刀太快。}\\
& \textit{shàng zhōu wǔ, wǒ jiù zài xué xiào yòng dāo qiē pò le tóng xué de shǒu, quán guài wǒ máo shǒu máo jiǎo, dà dà liē liē, yòng dāo tài kuài.} \\
& `Last Friday, I accidentally cut a classmate's hand with a knife at school. It was entirely my fault for being careless, reckless, and using the knife too quickly.'\\
\end{tabular}
}    
\caption{Sentence examples from CEFE} \label{CEFE-translation}
\end{subfigure}
\caption{File format and sentence examples: CEFE}\label{example-of-CEFE}
\end{figure}

\subsection{Summary} \label{subsec:summary-data}

Table~\ref{summary-CGEC-dataset} summarizes representative Chinese GEC datasets, including their approximate size, learner population, annotation format, and the average number of target sentences per source sentence. {In the revised organization of this survey, we also indicate more explicitly whether each dataset is based on native-speaker data, L2 learner data, or mixed/unclear sources, since this distinction is important for interpreting both resource design and reported system results.} These resources show that Chinese GEC has developed through a heterogeneous collection of datasets that differ not only in scale, but also in learner population, genre, annotation granularity, and correction philosophy. This diversity has undoubtedly supported rapid progress in the field, but it also makes direct comparison across datasets and systems considerably more difficult.

{The dataset landscape should also be interpreted cautiously in relation to the broader literature on Chinese learner corpora and error-annotated resources. Although Chinese grammatical error taxonomies have been discussed in related work \citep{linlin-2006-error,dazhong-2020-analysis}, they have not been systematically applied across CGEC datasets and their annotation schemes in a way that would support a principled cross-resource comparison. For this reason, the present survey does not attempt a full taxonomy-based comparison of native-speaker and L2 learner resources. Instead, we focus on how datasets represent corrections in practice, while noting that a more systematic comparison between learner populations remains an important direction for future work.}

A first major issue is domain and genre concentration. Many widely used datasets are drawn from relatively narrow sources, such as learner examination writing, language learning platforms, school essays, or specific curated web materials. As a result, the observed error distributions often reflect the communicative setting in which the data were collected rather than the full range of grammatical phenomena found in broader Chinese writing. L2 resources tend to be dominated by learner-specific lexical, collocational, and structural deviations, while L1 resources often focus on examination-style writing, edited prose, or synthetic native-speaker error settings. This concentration raises an important generalization problem: models that perform well on one dataset may partly be learning genre-specific regularities rather than robust correction strategies that transfer across domains, registers, or writer populations. {The distinction between native-speaker and L2 learner datasets is likely to matter here, since the two populations plausibly differ not only in surface error frequency but also in the linguistic and pedagogical interpretation of those errors. However, there is still not enough computational work in Chinese GEC to support a systematic comparison of these two populations across matched annotation and evaluation settings.} A related consequence of this narrow sourcing is overlap across nominally distinct datasets: \citet{liu-etal-2024-towards-better}, for example, report identical or highly similar input sentences across native-speaker CGEC resources derived from related source pools, including FCGEC, NaSGEC-Exam, and NaCGEC, and show that such overlap can complicate the interpretation of cross-dataset performance differences.

A second issue is the heterogeneity of annotation conventions. The datasets surveyed in this section differ substantially in how they represent errors and corrections. Some resources, such as NLPCC2018, YACLC, and CEFE, primarily provide sentence-level source-target pairs, leaving edits to be inferred from the contrast between source and correction. Others, such as FCGEC, FlaCGEC, and CCTC, make correction more explicit through structured edit operations or localized annotations. CGED occupies yet another position, evolving from sentence-linked diagnosis toward span-based error marking. In addition, some datasets are fundamentally character-oriented, whereas others depend on predefined word segmentation for annotation or evaluation. Because Chinese word boundaries are not given orthographically, this variation is methodologically consequential rather than superficial. Differences in segmentation conventions and annotation format directly affect what counts as an edit, how corrections are aligned, and how system outputs are scored. Accordingly, results reported on different datasets are often less directly comparable than aggregate benchmark tables may suggest. {For this reason, the dataset section now includes, wherever available, information on segmentation convention, correction format, whether edits are explicit or only recoverable from source-target pairs, whether error labels are provided, whether multi-reference corrections exist, and whether annotation guidelines or quality-control procedures are documented.}

A third issue concerns the uneven coverage of error phenomena. High-frequency error types, such as missing function words, collocational mismatches, local word order problems, or simple redundancies, are relatively well represented in several datasets. By contrast, low-frequency but linguistically important phenomena are often sparse, inconsistently labeled, or absent altogether. This is especially true for constructions whose interpretation depends on wider discourse context, register, idiomaticity, or subtle interactions between syntax and semantics. In some datasets, this limitation follows from the natural skew of learner data; in others, it reflects the restricted inventory of operations used in synthetic generation. Consequently, dataset coverage is rarely neutral: each resource privileges certain grammatical phenomena while underrepresenting others, which in turn shapes the kinds of systems that can be trained and the kinds of conclusions that can be drawn from evaluation. {This also means that differences across datasets are not only quantitative, but structural: they reflect different assumptions about what kinds of errors should be visible, localizable, and correctable within the task formulation itself.}

A fourth issue is the limited availability of multi-reference correction and consistent quality control. Although multi-reference resources, such as MuCGEC and YACLC, demonstrate the value of capturing correction variability, most Chinese GEC datasets still provide only one target sentence per source, or do so only for part of the data. This remains a substantial limitation, since many Chinese sentences admit multiple valid corrections that differ in wording, structure, or degree of intervention. Single-reference evaluation therefore risks penalizing legitimate system outputs that do not match the released target exactly. Relatedly, quality-control procedures vary considerably across datasets. Some resources rely on carefully designed annotation workflows or multiple annotators, whereas others provide little explicit information about rubric consistency, adjudication, or inter-annotator agreement. As a result, the reliability of a correction target cannot always be assumed to be uniform across datasets.

These observations suggest that the current landscape of Chinese GEC resources should not be understood simply as a progression toward larger benchmarks. What matters equally is how a dataset defines the unit of correction, how it encodes errors, what linguistic phenomena it covers, and how consistently its targets are validated. In practice, dataset choice is therefore not a neutral experimental decision, but part of the theoretical and methodological framing of the task itself. {In this sense, the distinction between native-speaker and L2 learner resources should be made explicit and kept visible, even if the field does not yet provide sufficient evidence for a full systematic comparison of their error inventories and modeling implications.}

Recent work has also begun to refine automatic error annotation for CGEC, most notably through toolkits such as \texttt{ChERRANT} and later extensions that introduce more linguistically informed Chinese-specific categories. These developments are important because they affect how corrections are converted into \texttt{m2}-style edit representations and, consequently, how systems are evaluated. Since these questions are closely tied to implementation details, edit extraction, and tokenization choices, we defer a fuller discussion of automatic annotation frameworks to \ref{sec:annotation}. {This separation is deliberate: dataset-specific annotation schemes are discussed in the main text as part of resource description, whereas automatic annotation frameworks are treated separately as evaluation-oriented operational layers.}

These observations also suggest several priorities for future dataset development, including broader domain coverage, more explicit and comparable annotation schemes, better representation of low-frequency but linguistically important error types, and more systematically validated multi-reference corrections. {A further priority is more systematic documentation of learner population, segmentation assumptions, annotation guidelines, and validation procedures, so that future CGEC research can compare datasets and results on more transparent methodological grounds.}

\begin{landscape}
\begin{table}
\caption{Comparative summary of representative Chinese GEC datasets. The table emphasizes structural dimensions that affect comparability: writer population, domain, localization or segmentation basis, correction format, error-label explicitness, multi-reference availability, and documentation or quality-control information.} \label{summary-CGEC-dataset} 
\scriptsize
\begin{tabular}{@{}
p{0.055\linewidth}
p{0.060\linewidth}
p{0.130\linewidth}
p{0.130\linewidth}
p{0.145\linewidth}
p{0.125\linewidth}
p{0.100\linewidth}
p{0.110\linewidth}
@{}}

\toprule
\textbf{Dataset} &
\textbf{Population} &
\textbf{Main source/domain} &
\textbf{Localization or segmentation basis} &
\textbf{Correction format} &
\textbf{Error-label explicitness} &
\textbf{Multi-reference} &
\textbf{Documentation / QC} \\
\midrule

CGED
& L2
& learner exam / shared-task data
& sentence ID and span offsets
& diagnosis plus corrections in later tasks
& explicit detection labels
& no
& shared-task guidelines \\

NLPCC2018
& L2
& learner text / shared task
& word-segmented test source
& source--target sentence pairs
& labels inferred from edits
& yes (avg. refs. = 1.1)
& shared-task guidelines \\

MuCGEC
& L2
& CSL learner sources
& source--target pairs with character/word evaluation
& multi-reference corrections
& labels inferred from edits
& yes (avg. refs. = 2.3)
& documented, adjudicated \\

YACLC
& L2
& learner essays
& sentence-level annotation
& grammatical and fluency corrections
& acceptability and correction dimensions
& yes (avg. refs. = 10.6)
& documented annotation platform \\

FlaCGEC
& synthetic / L2-oriented
& HSK reading corpus and proficiency-standards grammar examples
& structured edit operations
& rule-generated and manually filtered edits
& explicit grammar points and edit types
& no
& documented generation process \\ \hdashline

FCGEC
& L1
& public-school Chinese exams; supplemental error-free sentences in news sites
& structured edit operations
& fine-grained correction operations
& explicit operations and categories
& yes (avg. refs. = 1.7)
& documented human annotation \\

CCTC
& L1
& news, blogs, and some popular science articles
& document-level release; sentence-local edit positions
& localized corrections in documents
& explicit broad error types
& no
& documented manual annotation \\

NaCGEC
& L1
& exams and news
& sentence-level categories
& source--target pairs and JSON fields
& sentence-level type labels
& yes (avg. refs. = 1.2)
& shared-task documentation \\

NaSGEC
& L1
& social media, undergraduate theses, and exams
& sentence rewriting 
& source--target pairs
& no primary span labels; edit types post-extracted
& yes (avg. refs. = 1.6)
& two annotators plus expert review \\

CEFE
& L1 students
& primary/secondary-school essays
& sentence rewriting / fluency tasks
& source--target rewriting and error-identification tracks
& explicit in identification tracks; implicit in rewriting
& no
& shared-task documentation \\

\bottomrule
\end{tabular}
\end{table}
\end{landscape}

\section{Evaluation in CGEC}
\label{sec:evaluation}

Evaluation in Chinese grammatical error correction (CGEC) is primarily edit based. A system output is compared with one or more reference corrections by extracting the edits that transform the source sentence into the corrected sentence. Precision, recall, and an $F$ score are then computed from the overlap between system edits and gold edits. A general overview of GEC evaluation, including standard edit based metrics and broader evaluation settings, is provided in \ref{sec:evaluation-gec}; the present section focuses on evaluation issues specific to CGEC.

This general paradigm follows the evaluation tradition established in grammatical error correction through MaxMatch (\texttt{M\textsuperscript{2}}; \citealp{dahlmeier-ng-2012-better}), which was used with $F_{0.5}$ in the CoNLL 2014 shared task for English \citep{ng-etal-2014-conll} and later adopted in the \textit{NLPCC 2018} shared task for Chinese. The use of $F_{0.5}$ reflects a task-specific preference for precision: in GEC, an unnecessary or incorrect correction can degrade an otherwise acceptable sentence, so systems are usually expected to make fewer but more reliable edits rather than maximize the number of proposed changes. Recall remains important, but it is weighted less heavily than precision because overcorrection is especially harmful in writing assistance. \texttt{ChERRANT} extends this edit based framework to Chinese by deriving edits over Chinese character sequences and assigning linguistically motivated edit types \citep{zhang-etal-2022-mucgec}.\label{r1-p25}

A central issue in CGEC evaluation is the unit over which edits are defined. Unlike English, written Chinese does not mark word boundaries overtly. Evaluation must therefore decide whether edits are computed over segmented words or over raw characters. Word based evaluation can preserve linguistically meaningful units, but it depends on a prior segmentation scheme. Character based evaluation avoids this dependency and is therefore more robust to segmentation variation, but it may decompose a single lexical or syntactic correction into several primitive operations. The difference is not between edit based and non edit based evaluation, but between different choices of evaluation unit within the same edit based paradigm.

\subsection{Character-level and word-level edit scoring}

Figure~\ref{fig:partial-correction-example} illustrates a simple case in which character-level and word-level scoring lead to different recall values. The source contains the erroneous form \zh{一钱}, which should be corrected to \zh{以前}, and also contains an unnecessary sentence-final \zh{了}. The system corrects the lexical error but leaves \zh{了} unchanged.

\begin{figure}[!ht]
    \centering
\begin{tabular}{@{}r p{0.8\textwidth}@{}}
\footnotesize{Original sentence} & \zh{我一钱没住过五星级旅馆，所以我很惊奇了。} \\
& \textit{wǒ yì qián méi zhù guò wǔxīngjí lǚguǎn, suǒyǐ wǒ hěn jīngqí le.} \\
& `I had never stayed in a five-star hotel before, so I was very surprised.' \\
\footnotesize{System correction} & \zh{我以前没住过五星级旅馆，所以我很惊奇了。} \\
& \textit{wǒ yǐqián méi zhù guò wǔxīngjí lǚguǎn, suǒyǐ wǒ hěn jīngqí le.} \\
& `I had never stayed in a five-star hotel before, so I was very surprised.' \\
\footnotesize{Reference correction} & \zh{我以前没住过五星级旅馆，所以我很惊奇。} \\
& \textit{wǒ yǐqián méi zhù guò wǔxīngjí lǚguǎn, suǒyǐ wǒ hěn jīngqí.} \\
& `I had never stayed in a five-star hotel before, so I was very surprised.' \\
\end{tabular}
    \caption{Partial correction under character-level and word-level edit representations.}
    \label{fig:partial-correction-example}
\end{figure}

Table~\ref{tab:char-alignment-partial} gives the character-level alignment. Under this representation, the lexical correction \zh{一钱} $\rightarrow$ \zh{以前} is decomposed into two independent substitutions.

\begin{table}[!ht]
\centering
\footnotesize
\setlength{\tabcolsep}{2.5pt}
\begin{tabular}{@{}rcccccccccccccccccccc@{}}
\toprule
& 1 & 2 & 3 & 4 & 5 & 6 & 7 & 8 & 9 & 10
& 11 & 12 & 13 & 14 & 15 & 16 & 17 & 18 & 19 & 20 \\
\midrule
Source
& \zh{我} & \zh{一} & \zh{钱} & \zh{没} & \zh{住} & \zh{过} & \zh{五} & \zh{星} & \zh{级} & \zh{旅}
& \zh{馆} & \zh{，} & \zh{所} & \zh{以} & \zh{我} & \zh{很} & \zh{惊} & \zh{奇} & \zh{了} & \zh{。} \\
Gold
& \zh{我} & \zh{以} & \zh{前} & \zh{没} & \zh{住} & \zh{过} & \zh{五} & \zh{星} & \zh{级} & \zh{旅}
& \zh{馆} & \zh{，} & \zh{所} & \zh{以} & \zh{我} & \zh{很} & \zh{惊} & \zh{奇} & \zh{。} &  \\
System
& \zh{我} & \zh{以} & \zh{前} & \zh{没} & \zh{住} & \zh{过} & \zh{五} & \zh{星} & \zh{级} & \zh{旅}
& \zh{馆} & \zh{，} & \zh{所} & \zh{以} & \zh{我} & \zh{很} & \zh{惊} & \zh{奇} & \zh{了} & \zh{。} \\
\bottomrule
\end{tabular}
\caption{Character-level alignment for the partial correction example.}
\label{tab:char-alignment-partial}
\end{table}

At the character level, the gold correction consists of two substitutions and one deletion:
\begin{align*}
Edit_{\mathrm{gold}}^{\mathrm{char}} = \{&
\textsc{S}(\text{\zh{一}} \rightarrow \text{\zh{以}}),~
\textsc{S}(\text{\zh{钱}} \rightarrow \text{\zh{前}}),~
\textsc{R}(\text{\zh{了}})
\}.
\end{align*}
The system edits are:
\begin{align*}
Edit_{\mathrm{sys}}^{\mathrm{char}} = \{&
\textsc{S}(\text{\zh{一}} \rightarrow \text{\zh{以}}),~
\textsc{S}(\text{\zh{钱}} \rightarrow \text{\zh{前}})
\}.
\end{align*}
The system therefore obtains two true positives:
\begin{equation}
\mathrm{TP}=2,\qquad
P=\frac{2}{2}=1.0,\qquad
R=\frac{2}{3}=0.67.
\end{equation}
The corresponding $F_1$ and $F_{0.5}$ scores are:
\begin{equation}
F_1 = \frac{2PR}{P+R} = 0.80,\qquad
F_{0.5} = \frac{(1+0.5^2)PR}{0.5^2 P + R} \approx 0.91.
\end{equation}

Table~\ref{tab:word-alignment-partial-fine} gives the corresponding word-level alignment. Under this representation, the same lexical correction is treated as one word-level substitution.

\begin{table}[!ht]
\centering
\footnotesize
\begin{tabular}{@{}rcccccccccccccc@{}}
\toprule
 & 1 & 2 & 3 & 4 & 5 & 6 & 7 & 8 & 9 & 10 & 11 & 12 & 13 & 14 \\
\midrule
Source
& \zh{我} & \zh{一钱} & \zh{没} & \zh{住} & \zh{过} & \zh{五星级} & \zh{旅馆} & \zh{，} & \zh{所以} & \zh{我} & \zh{很} & \zh{惊奇} & \zh{了} & \zh{。} \\
Gold
& \zh{我} & \zh{以前} & \zh{没} & \zh{住} & \zh{过} & \zh{五星级} & \zh{旅馆} & \zh{，} & \zh{所以} & \zh{我} & \zh{很} & \zh{惊奇} & \zh{。} &  \\
System
& \zh{我} & \zh{以前} & \zh{没} & \zh{住} & \zh{过} & \zh{五星级} & \zh{旅馆} & \zh{，} & \zh{所以} & \zh{我} & \zh{很} & \zh{惊奇} & \zh{了} & \zh{。} \\
\bottomrule
\end{tabular}
\caption{Word-level alignment for the partial correction example under a finer word-boundary analysis.}
\label{tab:word-alignment-partial-fine}
\end{table}

At the word level, however, the same lexical correction is treated as a single substitution:
\begin{align*}
Edit_{\mathrm{gold}}^{\mathrm{word}} = \{&
\textsc{S}(\text{\zh{一钱}} \rightarrow \text{\zh{以前}}),~ \textsc{R}(\text{\zh{了}})
\},\\
Edit_{\mathrm{sys}}^{\mathrm{word}} = \{&
\textsc{S}(\text{\zh{一钱}} \rightarrow \text{\zh{以前}})
\}.
\end{align*}
The system now obtains one true positive:
\begin{equation}
\mathrm{TP}=1,\qquad
P=\frac{1}{1}=1.0,\qquad
R=\frac{1}{2}=0.5.
\end{equation}
The corresponding $F_1$ and $F_{0.5}$ scores are:
\begin{equation}
F_1 = \frac{2PR}{P+R} = 0.67,\qquad
F_{0.5} = \frac{(1+0.5^2)PR}{0.5^2 P + R} \approx 0.83.
\end{equation}

The two evaluations agree that the system has perfect precision, since all system edits are correct. They differ in recall because the same lexical correction is counted as two primitive character edits under character-level evaluation, but as one lexical edit under word-level evaluation.

\subsection{Word order decomposition and single-reference mismatch}

\jp{This section illustrates two related evaluation problems: character-level decomposition obscures word order corrections, while single-reference matching may fail to reward a valid alternative correction.}

The difference between character-level and word-level evaluation becomes more pronounced for word order corrections. Figure~\ref{fig:wo-correction-example} gives an example in which the system output is fluent and acceptable, but its edit sequence does not match the reference correction.

\begin{figure}[!ht]
    \centering
\begin{tabular}{@{}r p{0.8\textwidth}@{}}
\footnotesize{Original sentence} & \zh{我在家里一个人学习中文。} \\
& \textit{wǒ zài jiālǐ yí gè rén xuéxí zhōngwén.} \\
& `I study Chinese alone at home.' \\
\footnotesize{System correction} & \zh{我一个人在家里学习中文。} \\
& \textit{wǒ yí gè rén zài jiālǐ xuéxí zhōngwén.} \\
& `I study Chinese alone at home.' \\
\footnotesize{Reference correction} & \zh{我在家里自学中文。} \\
& \textit{wǒ zài jiālǐ zìxué zhōngwén.} \\
& `I study Chinese on my own at home.' \\
\end{tabular}
    \caption{Word order correction and reference mismatch.}
    \label{fig:wo-correction-example}
\end{figure}

At the character level, the source-to-system comparison may be represented as deleting \zh{一个人} after \zh{家里} and inserting it after \zh{我}:
\begin{align*}
Edit_{\mathrm{sys}}^{\mathrm{char}}
=
\{&
\textsc{R}(\text{\zh{一个人}}, \text{after } \text{\zh{家里}}),~ \textsc{M}(\text{\zh{一个人}}, \text{after } \text{\zh{我}})
\}.
\end{align*}
If expanded into primitive character edits, this yields three deletions and three insertions:
\[
\#\textsc{R}=3,\qquad
\#\textsc{M}=3,\qquad
\#\textsc{S}=0.
\]
The reference, by contrast, rewrites \zh{一个人学习} as \zh{自学}:
\begin{align*}
Edit_{\mathrm{gold}}^{\mathrm{char}}
=
\{&
\textsc{S}(\text{\zh{一个人学习}}, \text{\zh{自学}})
\}.
\end{align*}
The system and gold edit sets therefore have no exact overlap:
\[
Edit_{\mathrm{sys}}^{\mathrm{char}}
\cap
Edit_{\mathrm{gold}}^{\mathrm{char}}
=
\emptyset.
\]
Thus,
\[
\mathrm{TP}=0,\qquad P=0,\qquad R=0.
\]

A word-level representation gives a more transparent analysis of the system output:
\begin{align*}
Edit_{\mathrm{sys}}^{\mathrm{word}}
=
\{&
\textsc{W}(\text{\zh{一个人}}, \text{from post-}\text{\zh{在家里}}\text{ to post-}\text{\zh{我}})
\}.
\end{align*}
The reference edit remains a lexical substitution:
\begin{align*}
Edit_{\mathrm{gold}}^{\mathrm{word}}
=
\{&
\textsc{S}(\text{\zh{一个人 学习}}, \text{\zh{自学}})
\}.
\end{align*}
The word-level edit sets again have no exact overlap:
\[
Edit_{\mathrm{sys}}^{\mathrm{word}}
\cap
Edit_{\mathrm{gold}}^{\mathrm{word}}
=
\emptyset,
\qquad
\mathrm{TP}=0.
\]
This example illustrates two distinct problems. First, character-level evaluation may split a single movement into insertion and deletion operations, thereby obscuring its word order nature. Second, even when word order is represented as a span-level edit, exact matching against a single reference may still assign no credit if the reference adopts a different correction strategy.

\subsection{Implications for CGEC evaluation} 

These examples show that CGEC evaluation is sensitive to the relation between annotation unit, scoring unit, and reference choice. Character based scoring has become the default in much of CGEC because it avoids committing to a single word segmentation standard. This is an important practical advantage, since segmentation may vary across corpora, automatic segmentation tools, and annotation guidelines. Character based scoring also allows evaluation to be applied directly to unsegmented Chinese strings.

However, character based scoring is not linguistically neutral. Lexical corrections may be decomposed into multiple character substitutions, and word order corrections may be represented as independent insertions and deletions. As a result, the number of true positives, false positives, and false negatives may change depending on whether the same correction is counted as a character sequence, a word, or a larger span. Word based scoring, in contrast, can better preserve lexical and syntactic units, but it inherits the assumptions of the segmentation scheme used to define those units.

The appropriate evaluation unit therefore depends on the design of the dataset and the purpose of the evaluation. If a dataset is distributed with explicit word segmentation and its annotation scheme defines edits over segmented words, word based scoring is appropriate. If the goal is to reduce segmentation related mismatch and compare systems directly over surface strings, character based scoring is more robust. In either case, the evaluation remains edit based; what changes is the unit over which edits are computed and matched.

A further complication arises when annotation and evaluation do not use the same unit. Some annotation pipelines define edits over segmented words, while evaluation is later performed at the character level. This can make system performance difficult to interpret, because the same correction may be rewarded differently after being decomposed into characters. Recent work has therefore examined how segmentation decisions interact with error annotation and scoring in Chinese \citep{gu-etal-2025-improving,wang-etal-2025-refined,qiu-etal-2025-multilingual}.

Overall, evaluation in CGEC should be understood as unit sensitive edit based evaluation. Minimal edit scoring remains the dominant framework, but its interpretation depends crucially on whether edits are defined over words, characters, or spans. CGEC studies should therefore state explicitly which unit is used for annotation, which unit is used for scoring, and how word order or span-level edits are handled, especially when the annotation and evaluation units do not coincide.

\section{Development of Chinese GEC methodologies}\label{sec:methods}

Over the past two decades, Chinese GEC has evolved through several methodological paradigms. Early research on Chinese grammatical errors focused primarily on error \textit{detection}. Rule-based systems using manually crafted linguistic rules \citep{jiang-etal-2012-rule-based} and statistical classifiers combining hand-engineered features with SVM or n-gram models \citep{yu-chen-2012-detecting,lee-etal-2014-sentence} were effective at identifying specific error types, and some early rule-based work began to explore correction for limited error categories such as semantic collocations \citep{wu-etal-2015-research}. These approaches remained oriented toward detection, however, and did not address the full correction task. Because Chinese grammatical error detection (CGED) has developed a body of work of its own, these detection-focused systems are reviewed in Appendix~\ref{sec:cged-appendix}.

Work on full error \textit{correction}, in which a system both identifies errors and generates corrected output, began with the adoption of machine translation frameworks. Machine translation framed GEC as a monolingual translation task, first through statistical machine translation (SMT) using phrase tables and language models learned from parallel corpora \citep{chiu-etal-2013-chinese,zhao-etal-2015-improving}, and later through neural machine translation (NMT) using sequence-to-sequence architectures \citep{ren-yang-xun-2018-sequence}. More recent work has used pretrained language models, applied through sequence tagging, encoder-decoder generation \citep{zhang-etal-2023-nasgec}, or prompting of large language models \citep{qu-wu-2023-evaluating}. This trajectory parallels the development of English GEC more broadly \citep{wang-etal-2021-comprehensive}. The remainder of this section traces these developments, beginning with machine translation approaches encompassing both SMT and NMT (Section~\ref{subsec:mt-approaches}), followed by language model-based approaches (Section~\ref{subsec:lm-approaches}), and a comparative discussion (Section~\ref{subsec:methods-discussion}). Reported performance should be interpreted with caution, since results depend heavily on the choice of dataset, metric, and preprocessing (see Section~\ref{sec:evaluation}).

\subsection{Machine translation approaches}
\label{subsec:mt-approaches}

Machine translation approaches treat error correction as a monolingual translation problem. Rather than encoding correction knowledge in hand-written rules, they learn correction patterns from parallel corpora of erroneous and corrected sentences. The work falls into two waves: statistical machine translation (SMT), which adapts phrase-based translation pipelines to the correction task, and neural machine translation (NMT), which uses sequence-to-sequence architectures trained end-to-end. Several NMT-era systems also adopted hybrid designs that combined neural components with SMT and rule-based modules. Throughout this period, the availability of parallel error-correction data for Chinese was the main constraint on what models could learn and on how reliably their performance could be compared. Table~\ref{summary-mt-approaches} summarizes the systems reviewed.

\paragraph{Statistical machine translation for GEC}

SMT-based approaches apply the standard translation pipeline, including phrase tables, language models, and minimum error rate training, to the task of transforming erroneous text into corrected output. \citet{chiu-etal-2013-chinese} applied SMT to correct spelling errors, focusing on phonological and visual similarities between characters; the model uses translation probabilities and language models to rank candidate corrections, addressing issues such as character substitution errors and over-segmentation. \citet{zhao-etal-2015-improving} introduced a hierarchical phrase-based SMT model augmented with synchronous context-free grammar (SCFG) rules. By employing corpus augmentation and minimum error rate training (MERT), the model targeted word-order corrections and lexical choice refinement.

By treating correction as a global rewriting problem rather than as a sequence of independent edit decisions, the SMT framing fit well with spelling corrections involving visually or phonologically similar characters, where the ``translation'' from an incorrect character to its intended form mirrors natural confusability patterns. SMT effectiveness depended heavily, however, on the availability and quality of parallel error-correction corpora, which remained limited for Chinese during this period. The reliance on phrase-level alignment also restricted the ability of SMT systems to handle errors involving long-range dependencies or large reorderings.

\paragraph{Neural machine translation for GEC}

The introduction of neural sequence-to-sequence models brought two main changes to the MT-based GEC paradigm: continuous representations learned jointly with the model replaced SMT's discrete phrase tables and explicit word alignments, and end-to-end training of a single neural network replaced the modular SMT pipeline of phrase extraction, alignment, and rescoring.

Several systems in this period adopted a hybrid strategy, combining neural models with rule-based and SMT components. \citet{zhou-etal-2018-chinese} proposed a hierarchical framework that integrates rule-based, SMT, and NMT models. The rule-based component handles out-of-vocabulary errors through dictionary lookup; the SMT models operate at both character and word levels; and the NMT models, consisting of two-layer LSTM networks with attention, refine corrections by capturing longer-range dependencies. Correction candidates from each component are merged using a language model-based conflict resolution strategy. Similarly, \citet{li-etal-2018-hybrid} combined a BiLSTM-CRF model for detection with rule-based, SMT, and NMT components for correction, reporting higher precision on missing-word and word-selection errors.

These hybrid systems illustrate a transitional strategy: neural components handled context-sensitive errors, such as word selection and complex reorderings, that rule-based and statistical methods did not handle well, while the earlier components retained their advantage for well-defined, local patterns such as spelling errors and function word corrections. The trade-off was increased system complexity and the need to coordinate outputs across heterogeneous modules. The conflict resolution strategies used in these systems, such as language model rescoring, can be seen as an early form of the ensemble approach that would later become common in language model-based CGEC.

Fully neural approaches treated CGEC as an end-to-end generation task, avoiding the need for separate detection and correction stages. \citet{ren-yang-xun-2018-sequence} developed a convolutional sequence-to-sequence model with Byte Pair Encoding (BPE) for subword segmentation, reporting the highest precision and $F_{0.5}$ scores in the NLPCC~2018 Shared Task. The convolutional architecture captured local dependencies, which is relevant for CGEC since many errors are localized within short text spans. Building on this approach, \citet{li-etal-2019-chinese} introduced two optimization techniques: shared embeddings between encoder and decoder, and policy gradient reinforcement learning to optimize sequence generation toward evaluation metrics such as GLEU. These early seq2seq models, however, showed limited ability to handle errors requiring long-range contextual reasoning, and their performance was constrained by the size of available parallel training data.

\paragraph{Summary}

The machine translation paradigm framed GEC as a data-driven rewriting task. SMT models learned correction patterns from parallel corpora, and NMT models continued this approach with learned representations and end-to-end training. Hybrid systems that combined SMT, NMT, and rule-based components illustrated the value of integrating complementary correction strategies, though at the cost of increased pipeline complexity. The principal limitations of MT-based approaches were data-related: SMT models depended on phrase-level alignment and parallel corpora, while pre-transformer NMT models required larger parallel corpora than were available for Chinese and had limited capacity for long-range dependencies.

\begin{landscape}
\scriptsize
\begin{longtable}{@{}p{0.10\linewidth} p{0.28\linewidth} p{0.21\linewidth} p{0.10\linewidth} p{0.21\linewidth}@{}}
\caption{Summary of machine translation approaches for Chinese GEC.}\label{summary-mt-approaches} \\

\toprule

\textbf{Study} & \textbf{Method} & \textbf{Dataset} & \textbf{Metric} & \textbf{Result} \\
\midrule
\endfirsthead

\multicolumn{5}{@{}l}{\itshape Table~\ref{summary-mt-approaches} continued} \\
\toprule
\textbf{Study} & \textbf{Method} & \textbf{Dataset} & \textbf{Metric} & \textbf{Result} \\
\midrule
\endhead

\midrule
\multicolumn{5}{r@{}}{\itshape Continued on next page} \\
\endfoot

\bottomrule
\endlastfoot

\multicolumn{5}{@{}l}{\textit{Statistical machine translation}} \\
\midrule

\citet{chiu-etal-2013-chinese}
& SMT-based Chinese spelling checker: modified word segmentation and single-character n-grams are used to detect likely typos; a phrasal SMT decoder with visual/phonological confusion sets and an n-gram language model selects corrections.
& Training/resources: Sinica Corpus, TWWaC, Ministry of Education word and idiom lists, and confusion sets from SIGHAN-7 and \citet{liu-etal-2011-chinese}. Testing: SIGHAN 2013 Chinese Spelling Check.
& $F_{1}$ (detection) / accuracy / precision (correction)
& Detection $F_{1}$ / correction accuracy / correction precision = 0.7392 / 0.443 / 0.6998. \\
\midrule

\citet{zhao-etal-2015-improving}
& Phrase-based and hierarchical phrase-based SMT for learner-error correction, using Stanford Word Segmenter, GIZA++/Moses, SCFG rule extraction, corpus augmentation, and parameter tuning.
& Training: cleaned Lang-8 Chinese Learner Corpus and HSK Dynamic Essay Corpus; NLP-TEA-1 data used for tuning/ablation. Testing: NLP-TEA-2 CGED shared task.
& Accuracy / $F_{1}$
& Official test: RUN3 gives the highest submitted $F_{1}$ = 0.1039; RUN2 gives the highest submitted accuracy = 0.449. NLP-TEA-1 tuning ablation: phrase-based / hierarchical phrase-based tuned $F_{1}$ = 0.0701 / 0.1080. \\

\midrule
\multicolumn{5}{@{}l}{\textit{Neural machine translation}} \\
\midrule

\citet{zhou-etal-2018-chinese}
& Hierarchical combination of rule-based, SMT, and NMT modules, including low-level and high-level rule-based correction, character-level and word-level SMT, LSTM/BiLSTM NMT variants, and language-model-based conflict resolution.
& NLPCC-2018; experiments compare minimal-edit correction candidates (MinEd) and expanded correction candidates (Expand).
& $F_{0.5}$ / $F_{1}$
& Best reported model: $F_{0.5}$ / $F_{1}$ = 0.2936 / 0.3371. \\
\midrule

\citet{fu-huang-duan-2018-youdao}
& Staged NMT system for NLPCC-2018: a 5-gram/SCS spelling-correction stage is followed by character- and subword-level Transformer NMT models; five pipeline outputs are reranked using a 5-gram language model.
& NLPCC-2018; 0.71M original training items are expanded to about 1.22M source--target pairs and filtered to about 0.76M pairs; 3,000 pairs are used for validation.
& $F_{0.5}$
& Best submission: $F_{0.5}$ = 0.2991 against the union of two annotators. \\
\midrule

\citet{ren-yang-xun-2018-sequence}
& Convolutional seq2seq NMT model treating CGEC as translation from erroneous Chinese to corrected Chinese; uses Jieba word segmentation, BPE, and pretrained embeddings, with an additional four-model ensemble experiment.
& NLPCC-2018; the dataset is randomly split into 1,215,876 training pairs and a validation set of 5,000 sentence pairs with source--target inconsistencies.
& $F_{0.5}$
& Official submitted single model / four-model ensemble $F_{0.5}$ = 0.2902 / 0.3057. \\
\midrule

\citet{li-etal-2019-chinese}
& Character-level convolutional seq2seq model with position and character embeddings; compares shared embeddings, policy-gradient optimization, and added correct sentences to tune the error-to-correct sentence ratio.
& Training/validation: NLPTEA 2016--2018 and NLPCC-2018 data after cleaning. Testing: NLPCC-2018 test set; CTB9 and WMT correct sentences are added in ratio experiments.
& $F_{0.5}$
& Shared-embedding / policy-gradient / 6:1 error-to-correct ratio model $F_{0.5}$ = 0.2041 / 0.2125 / 0.2079. \\
\midrule

\citet{zhao-wang-2020-maskgec}
& MaskGEC: Transformer seq2seq NMT trained with dynamic source masking; noising schemes include padding, random-token, word-frequency, homophone, and mixed substitution.
& NLPCC-2018; 5,000 sentence pairs are sampled from the training set as development data, with no additional natural-language resources.
& $F_{0.5}$
& Mixed dynamic masking: $F_{0.5}$ = 0.3697. \\

\end{longtable}
\end{landscape}

\subsection{Language model-based approaches} \label{subsec:lm-approaches}

This subsection reviews CGEC systems built on large-scale pretrained language models. The change from the MT-based approaches discussed above is that linguistic knowledge no longer needs to come entirely from parallel error-correction data: pretrained models such as BERT \citep{devlin-etal-2019-bert}, RoBERTa \citep{liu-etal-2019-roberta}, and BART \citep{lewis-etal-2020-bart} encode general representations through unsupervised pretraining on large text corpora, and these can then be fine-tuned for correction with smaller amounts of task-specific data. The correction systems in this paradigm fall into three broad categories: sequence tagging, encoder-decoder generation, and approaches based on large language models (LLMs). Ensemble and multi-stage strategies often combine systems from these categories. Table~\ref{summary-lm-approaches} summarizes the systems reviewed.

\paragraph{Sequence tagging approaches}

Several systems cast error correction as a sequence labeling problem, predicting edit operations (keep, delete, substitute, insert) at each token position rather than generating output text from scratch. This formulation is computationally efficient and avoids the risk of introducing new errors through unconstrained generation.

\citet{zan-etal-2020-chinese} employed BERT-based contextual word representations with a BiLSTM-CRF decoder, combining pretrained representations with structured prediction for both detection and correction. \citet{li-shi-2021-tail} proposed a non-autoregressive model with a BERT-initialized transformer encoder and a CRF decoder, addressing class imbalance with Focal Loss and a dual evaluation strategy for detection followed by correction. Further work sought to address limitations specific to tagging-based approaches. \citet{tan-etal-2023-focal} introduced a model inspired by GECToR \citep{omelianchuk-etal-2020-gector}, integrating StructBERT-Large \citep{wang-etal-2019-structbert} with Focal Loss and a Tagger Decoupling mechanism to mitigate label imbalance and tagging entanglement, where the model confuses correlated edit operations. \citet{wu-etal-2023-tlm} addressed transformer overfitting through token-level masking (TLM) techniques, including sibling masking and self-masking, comparing these regularization strategies with attention drop-off and DropHead mechanisms \citep{zhou-etal-2020-scheduled} across BERT and BART architectures.

Tagging models are well suited for error types that involve local substitutions or deletions, but they are limited for corrections requiring insertion of new tokens or extensive rewriting, since the output length is constrained by the input length. They also rely on a fixed inventory of edit operations, which can be difficult to design for the full range of Chinese error types.

\paragraph{Encoder-decoder and generation-based models}

Encoder-decoder models treat correction as conditional text generation, providing greater flexibility for insertions, deletions, and multi-token rewrites. Unlike the NMT seq2seq models discussed in Section~\ref{subsec:mt-approaches}, encoder-decoder models in this paradigm initialize their components with pretrained language model weights rather than training from scratch on parallel data alone.

Early transformer-based work explored relatively simple configurations. \citet{fu-huang-duan-2018-youdao} introduced a three-stage approach addressing surface-level errors, grammatical errors, and ensemble-based predictions using a transformer with default configuration. \citet{zhao-wang-2020-maskgec} proposed a transformer seq2seq model with dynamic masking strategies, such as padding substitution and homophone substitution, to enhance robustness to noise.

BART \citep{lewis-etal-2020-bart}, with its bidirectional encoder and autoregressive decoder, has been used in several CGEC systems. Its denoising pretraining objective, which reconstructs original text from corrupted input, aligns with the error correction task. \citet{zhang-etal-2023-nasgec} trained BART-base on native speaker texts to extend CGEC coverage to multi-domain L1 data. \citet{liu-etal-2024-towards-better} proposed a two-stage training strategy combining average and minimum training losses to improve convergence on multi-reference training data, applied to SynGEC \citep{zhang-etal-2022-syngec}, NaSGEC-BART \citep{zhang-etal-2023-nasgec}, and STG-Joint \citep{xu-etal-2022-fcgec}. In multilingual GEC, \citet{fang-etal-2023-transgec} augmented data using machine translation to create \textit{translationese} text with added noise, fine-tuning mT5 for English and Chinese GEC tasks.

Generation-based models handle a wider range of correction types than tagging approaches, but they introduce the risk of overcorrection, where the model rewrites grammatically correct spans or produces fluent but semantically different output. In Chinese, this risk is heightened by the flexibility of acceptable word order and the availability of multiple valid corrections for many errors, which makes it difficult to distinguish between genuine correction and unnecessary paraphrasing. Managing this trade-off between correction coverage and conservative fidelity is a recurring concern, and the $F_{0.5}$ metric commonly used in GEC evaluation explicitly weights precision higher to penalize overcorrection.

\paragraph{Large language models}

Decoder-only large language models (LLMs) introduced a new option for CGEC. \citet{qu-wu-2023-evaluating} assessed ChatGPT (GPT-3.5-turbo\footnote{\url{https://platform.openai.com/docs/models/gpt-3-5-turbo}}), ChatGLM-6B\footnote{\url{https://github.com/THUDM/ChatGLM-6B}}, and LLaMA-7B\footnote{\url{https://huggingface.co/meta-llama/Llama-2-7b}} on Chinese GEC, finding that GPT-3.5-turbo performed comparably to fine-tuned baselines under prompting. Unlike fine-tuned encoder-decoder models, LLMs can be applied through prompting or in-context learning without task-specific training data, although fine-tuning remains beneficial when annotated data is available. The integration of LLMs into multi-stage correction pipelines, discussed below, represents a further development of this direction, though the computational cost of LLM-based correction remains a practical constraint.

\paragraph{Ensemble and multi-stage strategies}

To balance the complementary strengths of tagging and generation models, several systems have adopted ensemble or multi-stage architectures. \citet{liang-etal-2020-bert} proposed a multi-stage approach combining position-tagging, correction-tagging, and BERT-fused NMT models, with the final results obtained via an ensemble of sequence-tagging and BERT-fused NMT components. \citet{tang-etal-2023-pre} combined multiple pretrained language models, including BERT-Base-Chinese\footnote{\url{https://huggingface.co/google-bert/bert-base-chinese}}, MacBERT-Base-Chinese \citep{cui-etal-2020-revisiting}, and GPT2-Chinese\footnote{\url{https://github.com/Morizeyao/GPT2-Chinese}}, using sentence-level, edit-level, and edit-combination ensembles. \citet{wang-etal-2024-lm} introduced LM-Combiner, a two-step contextual rewriting model that aggregates outputs from multiple GEC systems, including LaserTagger \citep{malmi-etal-2019-encode}, PIE \citep{awasthi-etal-2019-parallel}, GECToR \citep{omelianchuk-etal-2020-gector}, STG \citep{xu-etal-2022-fcgec}, BART \citep{lewis-etal-2020-bart}, and GPT2\footnote{\url{https://github.com/openai/gpt-2}}, using k-fold cross-inference and causal language models. \citet{yang-quan-2024-alirector} proposed Alirector, a three-stage correction model that uses a seq2seq model or decoder-only LLM for initial correction, applies bidirectional alignment to reduce overcorrections, and employs knowledge distillation for refinement.

Ensemble and multi-stage approaches tend to report higher scores on standard benchmarks than individual systems, which suggests that no single architecture dominates across all error types and that combining complementary systems remains useful. The cost is increased system complexity and reduced interpretability: ensemble pipelines are harder to analyze for individual error types and harder to deploy in resource-constrained settings.

\paragraph{Summary}

Pretrained language models have improved CGEC performance by providing richer contextual representations and reducing the need for task-specific parallel data, with each of the three architectural families (tagging, generation, LLMs) offering different trade-offs between efficiency, flexibility, and overcorrection risk. Tagging approaches are conservative and efficient but limited in expressiveness; generation-based models handle more error types but are prone to overcorrection; LLMs reduce the dependence on annotated data but introduce computational overhead. Ensemble and multi-stage strategies combine these complementary strengths, though at the cost of increased pipeline complexity. Direct comparison across systems remains difficult, however, because studies evaluate on different datasets, use different training configurations, and report results under varying conditions.

\begin{landscape}
\scriptsize
\begin{longtable}{@{}p{0.10\linewidth} p{0.28\linewidth} p{0.21\linewidth} p{0.10\linewidth} p{0.21\linewidth}@{}}
\caption{Summary of language model-based approaches for Chinese GEC.}\label{summary-lm-approaches} \\

\toprule
\textbf{Study} & \textbf{Method} & \textbf{Dataset} & \textbf{Metric} & \textbf{Result} \\
\midrule
\endfirsthead

\multicolumn{5}{@{}l}{\itshape Table~\ref{summary-lm-approaches} continued} \\
\toprule
\textbf{Study} & \textbf{Method} & \textbf{Dataset} & \textbf{Metric} & \textbf{Result} \\
\midrule
\endhead

\midrule
\multicolumn{5}{r@{}}{\itshape Continued on next page} \\
\endfoot

\bottomrule
\endlastfoot

\multicolumn{5}{@{}l}{\textit{Sequence tagging approaches}} \\
\midrule

\citet{li-shi-2021-tail}
& Tail-to-Tail non-autoregressive edit prediction with a BERT-initialized Transformer encoder, CRF dependency modeling, and Focal Loss; designed to handle substitution, deletion, insertion, and local paraphrasing.
& HybridSet/SIGHAN-2015 for fixed-length experiments; TtTSet for variable-length experiments.
& Detection / correction $F_{1}$
& SIGHAN-2015 detection / correction $F_{1}$ = 0.816 / 0.800. TtTSet detection / correction $F_{1}$ = 0.646 / 0.589. \\
\midrule

\citet{tan-etal-2023-focal}
& Adds Focal Loss and tagger decoupling to a StructBERT-Large GECToR model to address label imbalance and detection/tagging entanglement.
& MuCGEC, FCGEC, and MCSCSet.
& Span-level $F_{0.5}$ (\texttt{ChERRANT})
& MuCGEC / FCGEC / MCSCSet $F_{0.5}$ = 0.4141 / 0.3074 / 0.8309. \\

\midrule
\multicolumn{5}{@{}l}{\textit{Fine-tuned encoder--decoder models}} \\
\midrule

\citet{wu-etal-2023-tlm}
& Proposes token-level masking regularization for Transformer models; in the CGEC experiment, Chinese BART is fine-tuned with sibling/self-masking and compared with attention dropout, DropHead, and GECToR baselines.
& CGED 2021 and CGED 2020.
& Official score / detection $F_{1}$ / correction $F_{1}$
& CGED 2021: official score / detection $F_{1}$ / correction $F_{1}$ = 0.437 / 0.778 / 0.307. CGED 2020: 0.389 / 0.821 / 0.203. \\
\midrule

\citet{zhang-etal-2023-nasgec}
& Uses Chinese-BART-large as a Seq2Seq benchmark for multi-domain native-speaker CGEC, with experiments on learner data, pseudo native data, and domain fine-tuning.
& NaSGEC: Media, Thesis, and Exam.
& Span-level $F_{0.5}$ (\texttt{ChERRANT})
& Benchmark Media / Thesis / Exam $F_{0.5}$ = 0.4579 / 0.3187 / 0.2002. In-domain fine-tuning: 0.5643 / 0.3645 / 0.4029. \\
\midrule

\citet{fang-etal-2023-transgec}
& Mines translationese from MT parallel corpora with BERT-based classifiers, injects noise to build synthetic GEC data, and trains Transformer and mT5-large GEC models.
& NLPCC-2018.
& Word-level $F_{0.5}$
& Transformer + translationese / mT5-large + translationese $F_{0.5}$ = 0.3230 / 0.3870. \\
\midrule

\citet{liu-etal-2024-towards-better}
& Focused study of multi-reference training for Seq2Seq CGEC with pretrained BART, comparing reference concatenation, minimum-Levenshtein reference selection, average-loss training, minimum-loss training, and a two-stage AvgL+MinL schedule.
& L2: train on Chinese Lang-8, validate and test on MuCGEC. L1: train and validate on FCGEC, and test on FCGEC-Test and NaSGEC-Exam after leakage filtering.
& Span-level $F_{0.5}$ (\texttt{ChERRANT})
& MuCGEC-Test (AvgL) / FCGEC-Test (AvgL+MinL) / NaSGEC-Exam (AvgL+MinL) $F_{0.5}$ = 0.4747 / 0.5722 / 0.5117. \\

\midrule
\multicolumn{5}{@{}l}{\textit{Prompted LLM-based generation}} \\
\midrule

\citet{qu-wu-2025-evaluating}
& Evaluates prompted LLMs for CGEC across GPT, Doubao, DeepSeek, ChatGLM/GLM, LLaMA, Qwen, and QwQ model families, comparing general-purpose and reasoning models.
& NLPCC-2018 test, MuCGEC validation, FCGEC validation, and NaCGEC validation.
& Character-level / word-level $F_{0.5}$
& Best LLM character-level $F_{0.5}$ on NLPCC / MuCGEC / FCGEC / NaCGEC = 0.3210 / 0.2786 / 0.3414 / 0.3825. Best word-level $F_{0.5}$ = 0.4089 / 0.3675 / 0.3718 / 0.4195. \\

\midrule
\multicolumn{5}{@{}l}{\textit{Ensemble and multi-stage strategies}} \\
\midrule

\citet{tang-etal-2023-pre}
& Compares traditional edit voting with PLM-based sentence-, edit-, and edit-combination ensembles using BERT, MacBERT, and GPT2 perplexity over outputs from four CGEC systems.
& Training: Lang-8 and HSK for the base systems. Testing: MuCGEC-test and NLPCC-2018 test.
& Character-level / word-level $F_{0.5}$
& Traditional voting on MuCGEC / NLPCC (character-level) / NLPCC (word-level) $F_{0.5}$ = 0.4807 / 0.4409 / 0.4352; worse performance from PLM-based ensembles. \\
\midrule

\citet{wang-etal-2024-lm}
& LM-Combiner: a causal-LM post-correction model that rewrites one GEC system output conditioned on the original sentence; overcorrection data are built with k-fold cross-inference.
& FCGEC; outputs from several baseline CGEC systems are used in experiments.
& Span-level $F_{0.5}$ (\texttt{ChERRANT})
& BART + LM-Combiner on FCGEC-test: $F_{0.5}$ = 0.5130. Absolute-score gains over BART: $F_{0.5}$ = +0.058. \\
\midrule

\citet{yang-quan-2024-alirector}
& Alirector: alignment-enhanced CGEC for Seq2Seq and decoder-only LLMs; uses initial correction, bidirectional source--correction alignment, and knowledge distillation to reduce overcorrection.
& L2: train on Lang-8+HSK, validate on MuCGEC-Dev, and test on NLPCC18-Test; L1: FCGEC train/dev split, with FCGEC-Dev and NaCGEC-Test for testing.
& Word-level (L2) / span-level (L1, \texttt{ChERRANT}) $F_{0.5}$
& BART Alirector on NLPCC18-Test / NaCGEC-Test / FCGEC-Dev = 0.4667 / 0.6133 / 0.5342. Baichuan2-7B Alirector = 0.4501 / 0.6071 / 0.5349. \\

\end{longtable}
\end{landscape}

\subsection{Discussion\label{subsec:methods-discussion}}

The progression from machine translation to language model-based methods has increased model capacity and error coverage, but has not resolved several persistent challenges in CGEC.

\paragraph{What has improved}

The most consistent improvement across paradigms has been in \textit{coverage}: the range of error types that a system can correct. Earlier detection-focused approaches using rules or statistical classifiers (Appendix~\ref{sec:cged-appendix}) targeted specific error types and did not address the full correction task. Machine translation approaches, both SMT and NMT, broadened coverage by learning correction patterns from parallel data and treating correction as a global rewriting problem, though they depended heavily on data quality. Language model-based methods, especially pretrained transformer models, expanded the range of correctable errors by leveraging contextualized representations that capture longer-range dependencies.

A second area of improvement is \textit{data efficiency}. Pretrained language models such as BERT and BART encode general linguistic knowledge from large-scale unsupervised pretraining, which reduces the amount of task-specific parallel data needed for fine-tuning. This is relevant for CGEC, where manually annotated error-correction corpora remain smaller than resources available for English GEC. LLMs further extend this advantage by enabling zero-shot or few-shot correction through prompting.

Third, \textit{system design} has become more flexible. Tagging-based approaches offer efficient, conservative correction for local errors, while encoder-decoder models handle insertions and multi-token rewrites. Ensemble and multi-stage architectures combine these complementary strengths, and currently report the highest scores on standard benchmarks such as NLPCC~2018 and MuCGEC.

\paragraph{What remains challenging}

Despite these advances, several persistent challenges limit CGEC performance. First, \textit{overcorrection} is a recurring problem for generation-based models, which may rewrite grammatically correct spans or produce fluent but semantically altered output. Managing the balance between recall (catching more errors) and precision (avoiding false corrections) remains a difficulty for the field, and the $F_{0.5}$ metric commonly used in GEC evaluation explicitly weights precision higher for this reason.

Second, \textit{complex and discourse-level errors} remain difficult for current approaches. Errors involving long-range dependencies, such as subject-verb agreement across clauses, discourse coherence, or pragmatic appropriateness, are poorly captured by models that operate at the sentence level. Few of the systems reviewed here address cross-sentence correction, and datasets providing cross-sentence annotation (e.g., CCTC) remain limited.

Third, \textit{Chinese-specific challenges} continue to affect system performance. Word segmentation remains a source of inconsistency: different segmentation conventions lead to different edit representations and different evaluation outcomes, even when the underlying correction is the same. Character-level models avoid this issue but lose word-level linguistic information. In addition, Chinese-specific error types, such as particle misuse (\zh{的}/\zh{地}/\zh{得}), measure word errors, and topic-comment structure violations, require fine-grained linguistic knowledge that general-purpose pretrained models may not fully capture. The three \textit{de} particles are illustrative: they are homophones used in distinct grammatical contexts (attribution, adverbial modification, and complementation), yet their confusion accounts for a large share of errors in both L1 and L2 writing. Current models can often correct these errors in straightforward cases, but struggle when the surrounding context is itself ambiguous or non-standard.

\paragraph{The comparability problem}

A recurring obstacle for assessing the state of the art in CGEC is the difficulty of comparing results across studies. Systems are evaluated on different datasets (CGED, NLPCC~2018, MuCGEC, FCGEC, NaCGEC), using different metrics ($F_1$ for detection, $F_{0.5}$ for correction), different training configurations, and different preprocessing pipelines. The distinction between error detection and error correction is not always clearly maintained, and some systems are evaluated on both tasks while others address only one. As a result, it is difficult to determine which approach is ``best'' in a general sense.

The fragmentation of evaluation practices across the field remains a barrier to cumulative progress. Standardizing evaluation protocols, including the choice of benchmark datasets, segmentation conventions, and metric definitions, would improve the interpretability of future CGEC research.

\section{Conclusion and future perspectives} \label{sec:conclusion}

This survey provides a comprehensive overview of Chinese Grammatical Error Correction, covering datasets, annotation methodologies, evaluation metrics, and system advancements. We review the breadth of available CGEC corpora and analyze their annotation schemes, highlighting inconsistencies and proposing a standardized error typology. On the evaluation front, we examine reference-based metrics commonly used in English GEC and their adaptations for CGEC, including character-level scoring and multi-reference evaluations. We also trace the evolution of CGEC systems from statistical and neural machine-translation-based formulations to recent language-model-based approaches. Together, these components clarify how dataset design, annotation choices, evaluation metrics, and modeling approaches interact in CGEC.

Looking ahead, several directions can be explored to further enhance CGEC systems. One approach is to embed deeper linguistic insights into neural models to better handle errors requiring an understanding of syntax and semantics, thereby addressing structural and discourse-level issues. Another promising avenue is leveraging large pre-trained language models: GPT-style systems have shown strong correction capabilities, and fine-tuning or prompting such models could boost performance while reducing reliance on extensive labeled data. Finally, CGEC systems could become more adaptive and personalized by tailoring corrections to a writer's proficiency level or common errors, thus providing user-specific feedback and learning support.

Future research should also focus on improving CGEC annotation and evaluation. Word segmentation ambiguity remains a key challenge: the lack of spaces between words in Chinese can lead to inconsistent tokenization and error detection. While evaluations often resort to character-level scoring to avoid segmentation errors, developing consistent segmentation standards is important to preserve word-level linguistic information. Another direction is refining error typologies. The appendix reviews a previously published, linguistically informed typology with fine-grained categories for Chinese errors, including distinctions among spelling-like errors and particle errors. Future work can further test, expand, and standardize such taxonomies across datasets so that they support more precise and targeted corrections. Finally, multi-reference annotations make evaluation more robust by capturing the diversity of valid corrections; for example, datasets with multiple reference sentences (e.g., MuCGEC) yield fairer assessments and higher correlation with human judgments, indicating that allowing multiple acceptable answers leads to more reliable evaluation.

Chinese GEC is likely to remain important in multilingual GEC and broader NLP applications. Insights from Chinese-specific challenges can inform error correction techniques for other languages, and conversely cross-linguistic approaches (such as training multilingual GEC models) may further enhance CGEC. Moreover, advances in CGEC benefit real-world scenarios: from intelligent tutoring systems for language learners, to writing assistants that help professionals produce error-free text. 
This survey's contributions lie in clarifying the relationships among datasets, annotation practices, and evaluation methods, and in showing how these choices affect the comparability and interpretation of CGEC results. This provides a stronger foundation for future research on more consistent and linguistically informed Chinese GEC systems.\label{r2-s5}
We emphasize the need for continued interdisciplinary collaboration among NLP researchers, linguists, and educators to tackle the unique challenges of Chinese grammatical errors. With these combined efforts, the CGEC community can develop more accurate, explainable, and adaptable error correction systems, ultimately improving automated writing assistance for all users.

\section*{Acknowledgments}
We acknowledge that Appendix \ref{sec:annotation} has been published at \textit{COLING 2025} \citep{gu-etal-2025-improving}, and we provide a summary of it here.
We are grateful to Min Zeng for the dataset curation (Section \ref{sec:datasets}) and to Junrui Wang for the system implementation (Appendix \ref{sec:annotation}), as well as for their assistance with the early draft of this manuscript.


\begin{thebibliography}{99}
\providecommand{\natexlab}[1]{#1}
\providecommand{\url}[1]{\texttt{#1}}
\expandafter\ifx\csname urlstyle\endcsname\relax
  \providecommand{\doi}[1]{doi: #1}\else
  \providecommand{\doi}{doi: \begingroup \urlstyle{rm}\Url}\fi

\bibitem[Awasthi et~al.(2019)Awasthi, Sarawagi, Goyal, Ghosh, and Piratla]{awasthi-etal-2019-parallel}
Abhijeet Awasthi, Sunita Sarawagi, Rasna Goyal, Sabyasachi Ghosh, and Vihari Piratla.
\newblock {Parallel Iterative Edit Models for Local Sequence Transduction}.
\newblock In \emph{Proceedings of the 2019 Conference on Empirical Methods in Natural Language Processing and the 9th International Joint Conference on Natural Language Processing (EMNLP-IJCNLP)}, pages 4260--4270, Hong Kong, China, 11 2019. Association for Computational Linguistics.
\newblock \doi{10.18653/v1/D19-1435}.
\newblock URL \url{https://aclanthology.org/D19-1435}.

\bibitem[Boyd(2018)]{boyd-2018-using}
Adriane Boyd.
\newblock {Using Wikipedia Edits in Low Resource Grammatical Error Correction}.
\newblock In Wei Xu, Alan Ritter, Tim Baldwin, and Afshin Rahimi, editors, \emph{Proceedings of the 2018 EMNLP Workshop W-NUT: The 4th Workshop on Noisy User-generated Text}, pages 79--84, Brussels, Belgium, 11 2018. Association for Computational Linguistics.
\newblock \doi{10.18653/v1/W18-6111}.
\newblock URL \url{https://aclanthology.org/W18-6111}.

\bibitem[Bryant et~al.(2017)Bryant, Felice, and Briscoe]{bryant-etal-2017-automatic}
Christopher Bryant, Mariano Felice, and Ted Briscoe.
\newblock {Automatic Annotation and Evaluation of Error Types for Grammatical Error Correction}.
\newblock In Regina Barzilay and Min-Yen Kan, editors, \emph{Proceedings of the 55th Annual Meeting of the Association for Computational Linguistics (Volume 1: Long Papers)}, pages 793--805, Vancouver, Canada, 7 2017. Association for Computational Linguistics.
\newblock \doi{10.18653/v1/P17-1074}.
\newblock URL \url{https://aclanthology.org/P17-1074}.

\bibitem[Bryant et~al.(2019)Bryant, Felice, Andersen, and Briscoe]{bryant-etal-2019-bea}
Christopher Bryant, Mariano Felice, {\O}istein~E. Andersen, and Ted Briscoe.
\newblock {The BEA-2019 Shared Task on Grammatical Error Correction}.
\newblock In \emph{Proceedings of the Fourteenth Workshop on Innovative Use of NLP for Building Educational Applications}, pages 52--75, Florence, Italy, 8 2019. Association for Computational Linguistics.
\newblock \doi{10.18653/v1/W19-4406}.
\newblock URL \url{https://aclanthology.org/W19-4406}.

\bibitem[Bryant et~al.(2023)Bryant, Yuan, Qorib, Cao, Ng, and Briscoe]{bryant-etal-2023-grammatical}
Christopher Bryant, Zheng Yuan, Muhammad~Reza Qorib, Hannan Cao, Hwee~Tou Ng, and Ted Briscoe.
\newblock {Grammatical Error Correction: A Survey of the State of the Art}.
\newblock \emph{Computational Linguistics}, 49\penalty0 (3):\penalty0 643--701, 9 2023.
\newblock \doi{10.1162/coli{\_}a{\_}00478}.
\newblock URL \url{https://aclanthology.org/2023.cl-3.4/}.

\bibitem[Che et~al.(2010)Che, Li, and Liu]{che-etal-2010-ltp}
Wanxiang Che, Zhenghua Li, and Ting Liu.
\newblock {LTP: A Chinese Language Technology Platform}.
\newblock In Yang Liu and Ting Liu, editors, \emph{Coling 2010: Demonstrations}, pages 13--16, Beijing, China, 8 2010. Coling 2010 Organizing Committee.
\newblock URL \url{https://aclanthology.org/C10-3004}.

\bibitem[Chen et~al.(2015)Chen, Wu, Chen, Yang, and Yang]{chen-etal-2015-chinese}
Po-Lin Chen, Shih-Hung Wu, Liang-Pu Chen, Ping-Che Yang, and Ren-Dar Yang.
\newblock {Chinese Grammatical Error Diagnosis by Conditional Random Fields}.
\newblock In Hsin-Hsi Chen, Yuen-Hsien Tseng, Yuji Matsumoto, and Lung~Hsiang Wong, editors, \emph{Proceedings of the 2nd Workshop on Natural Language Processing Techniques for Educational Applications}, pages 7--14, Beijing, China, 7 2015. Association for Computational Linguistics.
\newblock \doi{10.18653/v1/W15-4402}.
\newblock URL \url{https://aclanthology.org/W15-4402}.

\bibitem[Cheng et~al.(2014)Cheng, Yu, and Chen]{cheng-etal-2014-chinese}
Shuk-Man Cheng, Chi-Hsin Yu, and Hsin-Hsi Chen.
\newblock {Chinese Word Ordering Errors Detection and Correction for Non-Native Chinese Language Learners}.
\newblock In Junichi Tsujii and Jan Hajic, editors, \emph{Proceedings of COLING 2014, the 25th International Conference on Computational Linguistics: Technical Papers}, pages 279--289, Dublin, Ireland, 8 2014. Dublin City University and Association for Computational Linguistics.
\newblock URL \url{https://aclanthology.org/C14-1028}.

\bibitem[Cheng and Duan(2020)]{cheng-duan-2020-chinese}
Yong Cheng and Mofan Duan.
\newblock {Chinese Grammatical Error Detection Based on BERT Model}.
\newblock In Erhong YANG, Endong XUN, Baolin ZHANG, and Gaoqi RAO, editors, \emph{Proceedings of the 6th Workshop on Natural Language Processing Techniques for Educational Applications}, pages 108--113, Suzhou, China, 12 2020. Association for Computational Linguistics.
\newblock \doi{10.18653/v1/2020.nlptea-1.15}.
\newblock URL \url{https://aclanthology.org/2020.nlptea-1.15/}.

\bibitem[Chiu et~al.(2013)Chiu, Wu, and Chang]{chiu-etal-2013-chinese}
Hsun-wen Chiu, Jian-cheng Wu, and Jason~S Chang.
\newblock {Chinese Spelling Checker Based on Statistical Machine Translation}.
\newblock In Liang-Chih Yu, Yuen-Hsien Tseng, Jingbo Zhu, and Fuji Ren, editors, \emph{Proceedings of the Seventh SIGHAN Workshop on Chinese Language Processing}, pages 49--53, Nagoya, Japan, 10 2013. Asian Federation of Natural Language Processing.
\newblock URL \url{https://aclanthology.org/W13-4408}.

\bibitem[Cui et~al.(2020)Cui, Che, Liu, Qin, Wang, and Hu]{cui-etal-2020-revisiting}
Yiming Cui, Wanxiang Che, Ting Liu, Bing Qin, Shijin Wang, and Guoping Hu.
\newblock {Revisiting Pre-Trained Models for Chinese Natural Language Processing}.
\newblock In Trevor Cohn, Yulan He, and Yang Liu, editors, \emph{Findings of the Association for Computational Linguistics: EMNLP 2020}, pages 657--668, Online, 11 2020. Association for Computational Linguistics.
\newblock \doi{10.18653/v1/2020.findings-emnlp.58}.
\newblock URL \url{https://aclanthology.org/2020.findings-emnlp.58/}.

\bibitem[Cui et~al.(2021)Cui, Che, Liu, Qin, and Yang]{cui-etal-2021-pre}
Yiming Cui, Wanxiang Che, Ting Liu, Bing Qin, and Ziqing Yang.
\newblock {Pre-Training With Whole Word Masking for Chinese BERT}.
\newblock \emph{IEEE/ACM Trans. Audio, Speech and Lang. Proc.}, 29:\penalty0 3504–3514, 11 2021.
\newblock ISSN 2329-9290.
\newblock \doi{10.1109/TASLP.2021.3124365}.
\newblock URL \url{https://doi.org/10.1109/TASLP.2021.3124365}.

\bibitem[Dahlmeier and Ng(2012)]{dahlmeier-ng-2012-better}
Daniel Dahlmeier and Hwee~Tou Ng.
\newblock {Better Evaluation for Grammatical Error Correction}.
\newblock In Eric Fosler-Lussier, Ellen Riloff, and Srinivas Bangalore, editors, \emph{Proceedings of the 2012 Conference of the North American Chapter of the Association for Computational Linguistics: Human Language Technologies}, pages 568--572, Montr{\'{e}}al, Canada, 6 2012. Association for Computational Linguistics.
\newblock URL \url{https://aclanthology.org/N12-1067}.

\bibitem[Dale and Kilgarriff(2011)]{dale-kilgarriff-2011-helping}
Robert Dale and Adam Kilgarriff.
\newblock {Helping Our Own: The HOO 2011 Pilot Shared Task}.
\newblock In Claire Gardent and Kristina Striegnitz, editors, \emph{Proceedings of the 13th European Workshop on Natural Language Generation}, pages 242--249, Nancy, France, 9 2011. Association for Computational Linguistics.
\newblock URL \url{https://aclanthology.org/W11-2838/}.

\bibitem[Dazhong(2020)]{dazhong-2020-analysis}
Li~Dazhong.
\newblock \emph{{Analysis of Errors of Foreign Students in Learning Chinese Grammar}}.
\newblock Beijng Language and Culture University Press, Beijng, 2020.
\newblock ISBN 9787561957653.

\bibitem[Deng and Hu(2022)]{deng-hu-2022-examination}
Siqi Deng and Wenhua Hu.
\newblock {An examination of Chinese character writing errors: Developmental differences among Chinese as a foreign language learners}.
\newblock \emph{Journal of Chinese Writing Systems}, 6\penalty0 (1):\penalty0 39--51, 2022.
\newblock \doi{10.1177/25138502211066611}.
\newblock URL \url{https://doi.org/10.1177/25138502211066611}.

\bibitem[Devlin et~al.(2019)Devlin, Chang, Lee, and Toutanova]{devlin-etal-2019-bert}
Jacob Devlin, Ming-Wei Chang, Kenton Lee, and Kristina Toutanova.
\newblock {BERT: Pre-training of Deep Bidirectional Transformers for Language Understanding}.
\newblock In Jill Burstein, Christy Doran, and Thamar Solorio, editors, \emph{Proceedings of the 2019 Conference of the North American Chapter of the Association for Computational Linguistics: Human Language Technologies, Volume 1 (Long and Short Papers)}, pages 4171--4186, Minneapolis, Minnesota, 6 2019. Association for Computational Linguistics.
\newblock \doi{10.18653/v1/N19-1423}.
\newblock URL \url{https://aclanthology.org/N19-1423/}.

\bibitem[Du et~al.(2023)Du, Zhao, Tian, Wang, Wang, Lan, and Lu]{du-etal-2023-flacgec}
Hanyue Du, Yike Zhao, Qingyuan Tian, Jiani Wang, Lei Wang, Yunshi Lan, and Xuesong Lu.
\newblock {FlaCGEC: A Chinese Grammatical Error Correction Dataset with Fine-grained Linguistic Annotation}.
\newblock In \emph{Proceedings of the 32nd ACM International Conference on Information and Knowledge Management}, CIKM '23, page 5321–5325, New York, NY, USA, 2023. Association for Computing Machinery.
\newblock ISBN 9798400701245.
\newblock \doi{10.1145/3583780.3615119}.
\newblock URL \url{https://doi.org/10.1145/3583780.3615119}.

\bibitem[Fang et~al.(2023)Fang, Liu, Wong, Zhan, Ding, Chao, Tao, and Zhang]{fang-etal-2023-transgec}
Tao Fang, Xuebo Liu, Derek~F Wong, Runzhe Zhan, Liang Ding, Lidia~S Chao, Dacheng Tao, and Min Zhang.
\newblock {TransGEC: Improving Grammatical Error Correction with Translationese}.
\newblock In Anna Rogers, Jordan Boyd-Graber, and Naoaki Okazaki, editors, \emph{Findings of the Association for Computational Linguistics: ACL 2023}, pages 3614--3633, Toronto, Canada, 7 2023. Association for Computational Linguistics.
\newblock \doi{10.18653/v1/2023.findings-acl.223}.
\newblock URL \url{https://aclanthology.org/2023.findings-acl.223/}.

\bibitem[Felice and Briscoe(2015)]{felice-briscoe-2015-towards}
Mariano Felice and Ted Briscoe.
\newblock {Towards a standard evaluation method for grammatical error detection and correction}.
\newblock In Rada Mihalcea, Joyce Chai, and Anoop Sarkar, editors, \emph{Proceedings of the 2015 Conference of the North American Chapter of the Association for Computational Linguistics: Human Language Technologies}, pages 578--587, Denver, Colorado, 2015. Association for Computational Linguistics.
\newblock \doi{10.3115/v1/N15-1060}.
\newblock URL \url{https://aclanthology.org/N15-1060/}.

\bibitem[Fu et~al.(2018)Fu, Huang, and Duan]{fu-huang-duan-2018-youdao}
Kai Fu, Jin Huang, and Yitao Duan.
\newblock {Youdao's Winning Solution to the NLPCC-2018 Task 2 Challenge: A Neural Machine Translation Approach to Chinese Grammatical Error Correction}.
\newblock In Min Zhang, Vincent Ng, Dongyan Zhao, Sujian Li, and Hongying Zan, editors, \emph{Natural Language Processing and Chinese Computing}, pages 341--350, Cham, 2018. Springer International Publishing.
\newblock ISBN 978-3-319-99495-6.

\bibitem[Fung et~al.(2017)Fung, Debosschere, Wang, Li, Zhu, and Wong]{fung-etal-2017-nlptea}
Gabriel Fung, Maxime Debosschere, Dingmin Wang, Bo~Li, Jia Zhu, and Kam-Fai Wong.
\newblock {NLPTEA 2017 Shared Task -- Chinese Spelling Check}.
\newblock In Yuen-Hsien Tseng, Hsin-Hsi Chen, Lung-Hao Lee, and Liang-Chih Yu, editors, \emph{Proceedings of the 4th Workshop on Natural Language Processing Techniques for Educational Applications (NLPTEA 2017)}, pages 29--34, Taipei, Taiwan, 12 2017. Asian Federation of Natural Language Processing.
\newblock URL \url{https://aclanthology.org/W17-5905/}.

\bibitem[Gong et~al.(2022)Gong, Liu, Huang, and Zhang]{gong-etal-2022-revisiting}
Peiyuan Gong, Xuebo Liu, Heyan Huang, and Min Zhang.
\newblock {Revisiting Grammatical Error Correction Evaluation and Beyond}.
\newblock In Yoav Goldberg, Zornitsa Kozareva, and Yue Zhang, editors, \emph{Proceedings of the 2022 Conference on Empirical Methods in Natural Language Processing}, pages 6891--6902, Abu Dhabi, United Arab Emirates, 12 2022. Association for Computational Linguistics.
\newblock \doi{10.18653/v1/2022.emnlp-main.463}.
\newblock URL \url{https://aclanthology.org/2022.emnlp-main.463}.

\bibitem[Gu et~al.(2025)Gu, Huang, Zeng, Qiu, and Park]{gu-etal-2025-improving}
Yang Gu, Zihao Huang, Min Zeng, Mengyang Qiu, and Jungyeul Park.
\newblock {Improving Automatic Grammatical Error Annotation for Chinese Through Linguistically-Informed Error Typology}.
\newblock In Owen Rambow, Leo Wanner, Marianna Apidianaki, Hend Al-Khalifa, Barbara~Di Eugenio, and Steven Schockaert, editors, \emph{Proceedings of the 31st International Conference on Computational Linguistics}, pages 2781--2798, Abu Dhabi, UAE, 1 2025. Association for Computational Linguistics.
\newblock URL \url{https://aclanthology.org/2025.coling-main.189/}.

\bibitem[Hinson et~al.(2020)Hinson, Huang, and Chen]{hinson-etal-2020-heterogeneous}
Charles Hinson, Hen-Hsen Huang, and Hsin-Hsi Chen.
\newblock {Heterogeneous Recycle Generation for Chinese Grammatical Error Correction}.
\newblock In Donia Scott, Nuria Bel, and Chengqing Zong, editors, \emph{Proceedings of the 28th International Conference on Computational Linguistics}, pages 2191--2201, Barcelona, Spain (Online), 12 2020. International Committee on Computational Linguistics.
\newblock \doi{10.18653/v1/2020.coling-main.199}.
\newblock URL \url{https://aclanthology.org/2020.coling-main.199}.

\bibitem[Huang and Wang(2016)]{huang-wang-2016-bi}
Shen Huang and Houfeng Wang.
\newblock {Bi-LSTM Neural Networks for Chinese Grammatical Error Diagnosis}.
\newblock In Hsin-Hsi Chen, Yuen-Hsien Tseng, Vincent Ng, and Xiaofei Lu, editors, \emph{Proceedings of the 3rd Workshop on Natural Language Processing Techniques for Educational Applications (NLPTEA2016)}, pages 148--154, Osaka, Japan, 12 2016. The COLING 2016 Organizing Committee.
\newblock URL \url{https://aclanthology.org/W16-4919}.

\bibitem[Jiang et~al.(2012)Jiang, Wang, Lin, Wang, Cheng, Liu, Wang, and Zhang]{jiang-etal-2012-rule-based}
Ying Jiang, Tong Wang, Tao Lin, Fangjie Wang, Wenting Cheng, Xiaofei Liu, Chenghui Wang, and Weijian Zhang.
\newblock {A rule based Chinese spelling and grammar detection system utility}.
\newblock In \emph{2012 International Conference on System Science and Engineering (ICSSE)}, pages 437--440, 2012.
\newblock \doi{10.1109/ICSSE.2012.6257223}.

\bibitem[Junczys-Dowmunt and Grundkiewicz(2014)]{junczys-dowmunt-grundkiewicz-2014-amu}
Marcin Junczys-Dowmunt and Roman Grundkiewicz.
\newblock {The AMU System in the CoNLL-2014 Shared Task: Grammatical Error Correction by Data-Intensive and Feature-Rich Statistical Machine Translation}.
\newblock In Hwee~Tou Ng, Siew~Mei Wu, Ted Briscoe, Christian Hadiwinoto, Raymond~Hendy Susanto, and Christopher Bryant, editors, \emph{Proceedings of the Eighteenth Conference on Computational Natural Language Learning: Shared Task}, pages 25--33, Baltimore, Maryland, 6 2014. Association for Computational Linguistics.
\newblock \doi{10.3115/v1/W14-1703}.
\newblock URL \url{https://aclanthology.org/W14-1703/}.

\bibitem[Junczys-Dowmunt et~al.(2018)Junczys-Dowmunt, Grundkiewicz, Guha, and Heafield]{junczys-dowmunt-etal-2018-approaching}
Marcin Junczys-Dowmunt, Roman Grundkiewicz, Shubha Guha, and Kenneth Heafield.
\newblock {Approaching Neural Grammatical Error Correction as a Low-Resource Machine Translation Task}.
\newblock In Marilyn Walker, Heng Ji, and Amanda Stent, editors, \emph{Proceedings of the 2018 Conference of the North {\{}A{\}}merican Chapter of the Association for Computational Linguistics: Human Language Technologies, Volume 1 (Long Papers)}, pages 595--606, New Orleans, Louisiana, 6 2018. Association for Computational Linguistics.
\newblock \doi{10.18653/v1/N18-1055}.
\newblock URL \url{https://aclanthology.org/N18-1055/}.

\bibitem[Lee et~al.(2014)Lee, Yu, Lee, Tseng, Chang, and Chen]{lee-etal-2014-sentence}
Lung-Hao Lee, Liang-Chih Yu, Kuei-Ching Lee, Yuen-Hsien Tseng, Li-Ping Chang, and Hsin-Hsi Chen.
\newblock {A Sentence Judgment System for Grammatical Error Detection}.
\newblock In Lamia Tounsi and Rafal Rak, editors, \emph{Proceedings of COLING 2014, the 25th International Conference on Computational Linguistics: System Demonstrations}, pages 67--70, Dublin, Ireland, 8 2014. Dublin City University and Association for Computational Linguistics.
\newblock URL \url{https://aclanthology.org/C14-2015}.

\bibitem[Lee et~al.(2015)Lee, Yu, and Chang]{lee-etal-2015-overview}
Lung-Hao Lee, Liang-Chih Yu, and Li-Ping Chang.
\newblock {Overview of the NLP-TEA 2015 Shared Task for Chinese Grammatical Error Diagnosis}.
\newblock In Hsin-Hsi Chen, Yuen-Hsien Tseng, Yuji Matsumoto, and Lung~Hsiang Wong, editors, \emph{Proceedings of the 2nd Workshop on Natural Language Processing Techniques for Educational Applications}, pages 1--6, Beijing, China, 7 2015. Association for Computational Linguistics.
\newblock \doi{10.18653/v1/W15-4401}.
\newblock URL \url{https://aclanthology.org/W15-4401}.

\bibitem[Lee et~al.(2016)Lee, Rao, Yu, Xun, Zhang, and Chang]{lee-etal-2016-overview}
Lung-Hao Lee, Gaoqi Rao, Liang-Chih Yu, Endong Xun, Baolin Zhang, and Li-Ping Chang.
\newblock {Overview of NLP-TEA 2016 Shared Task for Chinese Grammatical Error Diagnosis}.
\newblock In Hsin-Hsi Chen, Yuen-Hsien Tseng, Vincent Ng, and Xiaofei Lu, editors, \emph{Proceedings of the 3rd Workshop on Natural Language Processing Techniques for Educational Applications ({\{}NLPTEA{\}}2016)}, pages 40--48, Osaka, Japan, 12 2016. The COLING 2016 Organizing Committee.
\newblock URL \url{https://aclanthology.org/W16-4906}.

\bibitem[Lewis et~al.(2020)Lewis, Liu, Goyal, Ghazvininejad, Mohamed, Levy, Stoyanov, and Zettlemoyer]{lewis-etal-2020-bart}
Mike Lewis, Yinhan Liu, Naman Goyal, Marjan Ghazvininejad, Abdelrahman Mohamed, Omer Levy, Veselin Stoyanov, and Luke Zettlemoyer.
\newblock {BART: Denoising Sequence-to-Sequence Pre-training for Natural Language Generation, Translation, and Comprehension}.
\newblock In \emph{Proceedings of the 58th Annual Meeting of the Association for Computational Linguistics}, pages 7871--7880, Online, 7 2020. Association for Computational Linguistics.
\newblock \doi{10.18653/v1/2020.acl-main.703}.
\newblock URL \url{https://aclanthology.org/2020.acl-main.703}.

\bibitem[Li et~al.(2018)Li, Zhou, Bao, Liu, Xu, and Li]{li-etal-2018-hybrid}
Chen Li, Junpei Zhou, Zuyi Bao, Hengyou Liu, Guangwei Xu, and Linlin Li.
\newblock {A Hybrid System for Chinese Grammatical Error Diagnosis and Correction}.
\newblock In Yuen-Hsien Tseng, Hsin-Hsi Chen, Vincent Ng, and Mamoru Komachi, editors, \emph{Proceedings of the 5th Workshop on Natural Language Processing Techniques for Educational Applications}, pages 60--69, Melbourne, Australia, 7 2018. Association for Computational Linguistics.
\newblock \doi{10.18653/v1/W18-3708}.
\newblock URL \url{https://aclanthology.org/W18-3708}.

\bibitem[Li and Shi(2021)]{li-shi-2021-tail}
Piji Li and Shuming Shi.
\newblock {Tail-to-Tail Non-Autoregressive Sequence Prediction for Chinese Grammatical Error Correction}.
\newblock In Chengqing Zong, Fei Xia, Wenjie Li, and Roberto Navigli, editors, \emph{Proceedings of the 59th Annual Meeting of the Association for Computational Linguistics and the 11th International Joint Conference on Natural Language Processing (Volume 1: Long Papers)}, pages 4973--4984, Online, 8 2021. Association for Computational Linguistics.
\newblock \doi{10.18653/v1/2021.acl-long.385}.
\newblock URL \url{https://aclanthology.org/2021.acl-long.385}.

\bibitem[Li et~al.(2019)Li, Zhao, Shi, Tan, Xu, Chen, Lan, and Lin]{li-etal-2019-chinese}
Si~Li, Jianbo Zhao, Guirong Shi, Yuanpeng Tan, Huifang Xu, Guang Chen, Haibo Lan, and Zhiqing Lin.
\newblock {Chinese Grammatical Error Correction Based on Convolutional Sequence to Sequence Model}.
\newblock \emph{IEEE Access}, 7:\penalty0 72905--72913, 2019.
\newblock \doi{10.1109/ACCESS.2019.2917631}.

\bibitem[Liang et~al.(2020)Liang, Zheng, Guo, Cui, Xiong, Rong, and Dong]{liang-etal-2020-bert}
Deng Liang, Chen Zheng, Lei Guo, Xin Cui, Xiuzhang Xiong, Hengqiao Rong, and Jinpeng Dong.
\newblock {BERT Enhanced Neural Machine Translation and Sequence Tagging Model for Chinese Grammatical Error Diagnosis}.
\newblock In Erhong YANG, Endong XUN, Baolin ZHANG, and Gaoqi RAO, editors, \emph{Proceedings of the 6th Workshop on Natural Language Processing Techniques for Educational Applications}, pages 57--66, Suzhou, China, 12 2020. Association for Computational Linguistics.
\newblock URL \url{https://aclanthology.org/2020.nlptea-1.8}.

\bibitem[Linlin(2006)]{linlin-2006-error}
Zhang Linlin.
\newblock \emph{{An Error Analysis of English Speaking Chinese Learners' Corpus}}.
\newblock PhD thesis, University of International Business and Economics, 2006.

\bibitem[Liu et~al.(2010)Liu, Lai, Chuang, and Lee]{liu-etal-2010-visually}
Chao-Lin Liu, Min-Hua Lai, Yi-Hsuan Chuang, and Chia-Ying Lee.
\newblock {Visually and Phonologically Similar Characters in Incorrect Simplified Chinese Words}.
\newblock In Chu-Ren Huang and Dan Jurafsky, editors, \emph{Coling 2010: Posters}, pages 739--747, Beijing, China, 8 2010. Coling 2010 Organizing Committee.
\newblock URL \url{https://aclanthology.org/C10-2085}.

\bibitem[Liu et~al.(2011)Liu, Lai, Tien, Chuang, Wu, and Lee]{liu-etal-2011-chinese}
Chao-Lin Liu, Min-Hua Lai, Kan-Wen Tien, Yi-Hsuan Chuang, Shih-Hung Wu, and Chia-Ying Lee.
\newblock {Visually and Phonologically Similar Characters in Incorrect Chinese Words: Analyses, Identification, and Applications}.
\newblock \emph{ACM Transactions on Asian Language Information Processing}, 10\penalty0 (2):\penalty0 1--39, 2011.
\newblock \doi{10.1145/1967293.1967297}.

\bibitem[Liu et~al.(2016)Liu, Han, Zhuo, and Zan]{liu-etal-2016-automatic}
Yajun Liu, Yingjie Han, Liyan Zhuo, and Hongying Zan.
\newblock {Automatic Grammatical Error Detection for Chinese based on Conditional Random Field}.
\newblock In Hsin-Hsi Chen, Yuen-Hsien Tseng, Vincent Ng, and Xiaofei Lu, editors, \emph{Proceedings of the 3rd Workshop on Natural Language Processing Techniques for Educational Applications (NLPTEA2016)}, pages 57--62, Osaka, Japan, 12 2016. The COLING 2016 Organizing Committee.
\newblock URL \url{https://aclanthology.org/W16-4908}.

\bibitem[Liu et~al.(2019)Liu, Ott, Goyal, Du, Joshi, Chen, Levy, Lewis, Zettlemoyer, and Stoyanov]{liu-etal-2019-roberta}
Yinhan Liu, Myle Ott, Naman Goyal, Jingfei Du, Mandar Joshi, Danqi Chen, Omer Levy, Mike Lewis, Luke Zettlemoyer, and Veselin Stoyanov.
\newblock {RoBERTa: A Robustly Optimized BERT Pretraining Approach}.
\newblock \emph{CoRR}, abs/1907.1\penalty0 (1):\penalty0 1--13, 2019.
\newblock URL \url{http://arxiv.org/abs/1907.11692}.

\bibitem[Liu et~al.(2024)Liu, Li, Jiang, Zhang, Li, and Zhang]{liu-etal-2024-towards-better}
Yumeng Liu, Zhenghua Li, HaoChen Jiang, Bo~Zhang, Chen Li, and Ji~Zhang.
\newblock {Towards Better Utilization of Multi-Reference Training Data for Chinese Grammatical Error Correction}.
\newblock In Lun-Wei Ku, Andre Martins, and Vivek Srikumar, editors, \emph{Findings of the Association for Computational Linguistics: ACL 2024}, pages 3044--3052, Bangkok, Thailand, 8 2024. Association for Computational Linguistics.
\newblock \doi{10.18653/v1/2024.findings-acl.180}.
\newblock URL \url{https://aclanthology.org/2024.findings-acl.180/}.

\bibitem[Ma et~al.(2022)Ma, Li, Sun, Zhou, Huang, Zhang, Yangning, Liu, Li, Cao, Zheng, and Shen]{ma-etal-2022-linguistic}
Shirong Ma, Yinghui Li, Rongyi Sun, Qingyu Zhou, Shulin Huang, Ding Zhang, Li~Yangning, Ruiyang Liu, Zhongli Li, Yunbo Cao, Haitao Zheng, and Ying Shen.
\newblock {Linguistic Rules-Based Corpus Generation for Native Chinese Grammatical Error Correction}.
\newblock In Yoav Goldberg, Zornitsa Kozareva, and Yue Zhang, editors, \emph{Findings of the Association for Computational Linguistics: EMNLP 2022}, pages 576--589, Abu Dhabi, United Arab Emirates, 12 2022. Association for Computational Linguistics.
\newblock \doi{10.18653/v1/2022.findings-emnlp.40}.
\newblock URL \url{https://aclanthology.org/2022.findings-emnlp.40}.

\bibitem[Malmi et~al.(2019)Malmi, Krause, Rothe, Mirylenka, and Severyn]{malmi-etal-2019-encode}
Eric Malmi, Sebastian Krause, Sascha Rothe, Daniil Mirylenka, and Aliaksei Severyn.
\newblock {Encode, Tag, Realize: High-Precision Text Editing}.
\newblock In Kentaro Inui, Jing Jiang, Vincent Ng, and Xiaojun Wan, editors, \emph{Proceedings of the 2019 Conference on Empirical Methods in Natural Language Processing and the 9th International Joint Conference on Natural Language Processing (EMNLP-IJCNLP)}, pages 5054--5065, Hong Kong, China, 11 2019. Association for Computational Linguistics.
\newblock \doi{10.18653/v1/D19-1510}.
\newblock URL \url{https://aclanthology.org/D19-1510/}.

\bibitem[N{\'{a}}plava et~al.(2022)N{\'{a}}plava, Straka, Strakov{\'{a}}, and Rosen]{naplava-etal-2022-czech}
Jakub N{\'{a}}plava, Milan Straka, Jana Strakov{\'{a}}, and Alexandr Rosen.
\newblock {Czech Grammar Error Correction with a Large and Diverse Corpus}.
\newblock \emph{Transactions of the Association for Computational Linguistics}, 10:\penalty0 452--467, 2022.
\newblock \doi{10.1162/tacl{\_}a{\_}00470}.
\newblock URL \url{https://aclanthology.org/2022.tacl-1.26}.

\bibitem[Napoles et~al.(2015)Napoles, Sakaguchi, Post, and Tetreault]{napoles-etal-2015-ground}
Courtney Napoles, Keisuke Sakaguchi, Matt Post, and Joel Tetreault.
\newblock {Ground Truth for Grammatical Error Correction Metrics}.
\newblock In Chengqing Zong and Michael Strube, editors, \emph{Proceedings of the 53rd Annual Meeting of the Association for Computational Linguistics and the 7th International Joint Conference on Natural Language Processing (Volume 2: Short Papers)}, pages 588--593, Beijing, China, 7 2015. Association for Computational Linguistics.
\newblock \doi{10.3115/v1/P15-2097}.
\newblock URL \url{https://aclanthology.org/P15-2097}.

\bibitem[Napoles et~al.(2016{\natexlab{a}})Napoles, Sakaguchi, Post, and Tetreault]{napoles-etal-2016-gleu}
Courtney Napoles, Keisuke Sakaguchi, Matt Post, and Joel Tetreault.
\newblock {GLEU Without Tuning}, 2016{\natexlab{a}}.
\newblock URL \url{https://arxiv.org/abs/1605.02592}.

\bibitem[Napoles et~al.(2016{\natexlab{b}})Napoles, Sakaguchi, and Tetreault]{napoles-etal-2016-theres}
Courtney Napoles, Keisuke Sakaguchi, and Joel Tetreault.
\newblock {There's No Comparison: Reference-less Evaluation Metrics in Grammatical Error Correction}.
\newblock In Jian Su, Kevin Duh, and Xavier Carreras, editors, \emph{Proceedings of the 2016 Conference on Empirical Methods in Natural Language Processing}, pages 2109--2115, Austin, Texas, 11 2016{\natexlab{b}}. Association for Computational Linguistics.
\newblock \doi{10.18653/v1/D16-1228}.
\newblock URL \url{https://aclanthology.org/D16-1228/}.

\bibitem[Ng et~al.(2013)Ng, Wu, Wu, Hadiwinoto, and Tetreault]{ng-EtAl:2013:CoNLL}
Hwee~Tou Ng, Siew~Mei Wu, Yuanbin Wu, Christian Hadiwinoto, and Joel Tetreault.
\newblock {The CoNLL-2013 Shared Task on Grammatical Error Correction}.
\newblock In \emph{Proceedings of the Seventeenth Conference on Computational Natural Language Learning: Shared Task}, pages 1--12, Sofia, Bulgaria, 8 2013. Association for Computational Linguistics.
\newblock URL \url{https://www.aclweb.org/anthology/W13-3601}.

\bibitem[Ng et~al.(2014)Ng, Wu, Briscoe, Hadiwinoto, Susanto, and Bryant]{ng-etal-2014-conll}
Hwee~Tou Ng, Siew~Mei Wu, Ted Briscoe, Christian Hadiwinoto, Raymond~Hendy Susanto, and Christopher Bryant.
\newblock {The CoNLL-2014 Shared Task on Grammatical Error Correction}.
\newblock In \emph{Proceedings of the Eighteenth Conference on Computational Natural Language Learning: Shared Task}, pages 1--14, Baltimore, Maryland, 6 2014. Association for Computational Linguistics.
\newblock URL \url{http://www.aclweb.org/anthology/W14-1701}.

\bibitem[Omelianchuk et~al.(2020)Omelianchuk, Atrasevych, Chernodub, and Skurzhanskyi]{omelianchuk-etal-2020-gector}
Kostiantyn Omelianchuk, Vitaliy Atrasevych, Artem Chernodub, and Oleksandr Skurzhanskyi.
\newblock {GECToR -- Grammatical Error Correction: Tag, Not Rewrite}.
\newblock In \emph{Proceedings of the Fifteenth Workshop on Innovative Use of NLP for Building Educational Applications}, pages 163--170, Seattle, WA, USA → Online, 7 2020. Association for Computational Linguistics.
\newblock \doi{10.18653/v1/2020.bea-1.16}.
\newblock URL \url{https://aclanthology.org/2020.bea-1.16}.

\bibitem[{\"{O}}stling et~al.(2025){\"{O}}stling, Kurfali, and Caines]{ostling-etal-2025-llm}
Robert {\"{O}}stling, Murathan Kurfali, and Andrew Caines.
\newblock {LLM-based post-editing as reference-free GEC evaluation}.
\newblock In Ekaterina Kochmar, Bashar Alhafni, Marie Bexte, Jill Burstein, Andrea Horbach, Ronja Laarmann-Quante, Anaïs Tack, Victoria Yaneva, and Zheng Yuan, editors, \emph{Proceedings of the 20th Workshop on Innovative Use of NLP for Building Educational Applications (BEA 2025)}, pages 213--224, Vienna, Austria, 7 2025. Association for Computational Linguistics.
\newblock ISBN 979-8-89176-270-1.
\newblock URL \url{https://aclanthology.org/2025.bea-1.16/}.

\bibitem[Papineni et~al.(2002)Papineni, Roukos, Ward, and Zhu]{papineni-etal-2002-bleu}
Kishore Papineni, Salim Roukos, Todd Ward, and Wei-Jing Zhu.
\newblock {BLEU: a Method for Automatic Evaluation of Machine Translation}.
\newblock In Pierre Isabelle, Eugene Charniak, and Dekang Lin, editors, \emph{Proceedings of the 40th Annual Meeting of the Association for Computational Linguistics}, pages 311--318, Philadelphia, Pennsylvania, USA, 7 2002. Association for Computational Linguistics.
\newblock \doi{10.3115/1073083.1073135}.
\newblock URL \url{https://aclanthology.org/P02-1040/}.

\bibitem[Petrov et~al.(2012)Petrov, Das, and McDonald]{petrov-das-mcdonald:2012:LREC}
Slav Petrov, Dipanjan Das, and Ryan McDonald.
\newblock {A Universal Part-of-Speech Tagset}.
\newblock In \emph{Proceedings of the Eighth International Conference on Language Resources and Evaluation (LREC-2012)}, pages 2089--2096, Istanbul, Turkey, 2012. European Language Resources Association (ELRA).
\newblock ISBN 978-2-9517408-7-7.

\bibitem[Qiu et~al.(2025)Qiu, Nguyen, Huang, Li, Gu, Gao, Liu, and Park]{qiu-etal-2025-multilingual}
Mengyang Qiu, Tran~Minh Nguyen, Zihao Huang, Zelong Li, Yang Gu, Qingyu Gao, Siliang Liu, and Jungyeul Park.
\newblock {Multilingual Grammatical Error Annotation: Combining Language-Agnostic Framework with Language-Specific Flexibility}.
\newblock In Ekaterina Kochmar, Bashar Alhafni, Marie Bexte, Jill Burstein, Andrea Horbach, Ronja Laarmann-Quante, Anaïs Tack, Victoria Yaneva, and Zheng Yuan, editors, \emph{Proceedings of the 20th Workshop on Innovative Use of NLP for Building Educational Applications (BEA 2025)}, pages 202--212, Vienna, Austria, 7 2025. Association for Computational Linguistics.
\newblock ISBN 979-8-89176-270-1.
\newblock URL \url{https://aclanthology.org/2025.bea-1.15/}.

\bibitem[Qu and Wu(2023)]{qu-wu-2023-evaluating}
Fanyi Qu and Yunfang Wu.
\newblock {Evaluating the Capability of Large-scale Language Models on Chinese Grammatical Error Correction Task}, 2023.
\newblock URL \url{https://arxiv.org/abs/2307.03972}.

\bibitem[Qu et~al.(2025)Qu, Tang, and Wu]{qu-wu-2025-evaluating}
Fanyi Qu, Chenming Tang, and Yunfang Wu.
\newblock {Evaluating the Capability of Large-scale Language Models on Chinese Grammatical Error Correction Task}, 2025.
\newblock URL \url{https://arxiv.org/abs/2307.03972}.

\bibitem[Rao et~al.(2017)Rao, Zhang, Xun, and Lee]{rao-etal-2017-ijcnlp}
Gaoqi Rao, Baolin Zhang, Endong Xun, and Lung-Hao Lee.
\newblock {IJCNLP-2017 Task 1: Chinese Grammatical Error Diagnosis}.
\newblock In Chao-Hong Liu, Preslav Nakov, and Nianwen Xue, editors, \emph{Proceedings of the IJCNLP 2017, Shared Tasks}, pages 1--8, Taipei, Taiwan, 12 2017. Asian Federation of Natural Language Processing.
\newblock URL \url{https://aclanthology.org/I17-4001}.

\bibitem[Rao et~al.(2018)Rao, Gong, Zhang, and Xun]{rao-etal-2018-overview}
Gaoqi Rao, Qi~Gong, Baolin Zhang, and Endong Xun.
\newblock {Overview of NLPTEA-2018 Share Task Chinese Grammatical Error Diagnosis}.
\newblock In Yuen-Hsien Tseng, Hsin-Hsi Chen, Vincent Ng, and Mamoru Komachi, editors, \emph{Proceedings of the 5th Workshop on Natural Language Processing Techniques for Educational Applications}, pages 42--51, Melbourne, Australia, 7 2018. Association for Computational Linguistics.
\newblock \doi{10.18653/v1/W18-3706}.
\newblock URL \url{https://aclanthology.org/W18-3706}.

\bibitem[Rao et~al.(2020)Rao, Yang, and Zhang]{rao-etal-2020-overview}
Gaoqi Rao, Erhong Yang, and Baolin Zhang.
\newblock {Overview of NLPTEA-2020 Shared Task for Chinese Grammatical Error Diagnosis}.
\newblock In Erhong Yang, Endong Xun, Baolin Zhang, and Gaoqi Rao, editors, \emph{Proceedings of the 6th Workshop on Natural Language Processing Techniques for Educational Applications}, pages 25--35, Suzhou, China, 12 2020. Association for Computational Linguistics.
\newblock URL \url{https://aclanthology.org/2020.nlptea-1.4}.

\bibitem[Ren et~al.(2018)Ren, Yang, and Xun]{ren-yang-xun-2018-sequence}
Hongkai Ren, Liner Yang, and Endong Xun.
\newblock {A Sequence to Sequence Learning for Chinese Grammatical Error Correction}.
\newblock In Min Zhang, Vincent Ng, Dongyan Zhao, Sujian Li, and Hongying Zan, editors, \emph{Natural Language Processing and Chinese Computing}, pages 401--410, Cham, 2018. Springer International Publishing.
\newblock ISBN 978-3-319-99501-4.

\bibitem[Ross et~al.(2024)Ross, Ma, Chen, He, and Yeh]{ross-2024-modern}
Claudia Ross, Jing-heng~Sheng Ma, Pei-Chia Chen, Baozhang He, and Meng Yeh.
\newblock \emph{{Modern Mandarin Chinese grammar: A practical guide}}.
\newblock Routledge, 2024.

\bibitem[Rothe et~al.(2021)Rothe, Mallinson, Malmi, Krause, and Severyn]{rothe-etal-2021-simple}
Sascha Rothe, Jonathan Mallinson, Eric Malmi, Sebastian Krause, and Aliaksei Severyn.
\newblock {A Simple Recipe for Multilingual Grammatical Error Correction}.
\newblock In \emph{Proceedings of the 59th Annual Meeting of the Association for Computational Linguistics and the 11th International Joint Conference on Natural Language Processing (Volume 2: Short Papers)}, pages 702--707, Online, 8 2021. Association for Computational Linguistics.
\newblock \doi{10.18653/v1/2021.acl-short.89}.
\newblock URL \url{https://aclanthology.org/2021.acl-short.89}.

\bibitem[Shen et~al.(2023)Shen, Wu, Bai, Wu, Zhou, Mao, Ge, and Xia]{shen-etal-2023-overview}
Xinshu Shen, Hongyi Wu, Xiaopeng Bai, Yuanbin Wu, Aimin Zhou, Shaoguang Mao, Tao Ge, and Yan Xia.
\newblock {Overview of CCL23-Eval Task 8: Chinese Essay Fluency Evaluation (CEFE) Task}.
\newblock In Maosong Sun, Bing Qin, Xipeng Qiu, Jing Jiang, and Xianpei Han, editors, \emph{Proceedings of the 22nd Chinese National Conference on Computational Linguistics (Volume 3: Evaluations)}, pages 282--292, Harbin, China, 8 2023. Chinese Information Processing Society of China.
\newblock URL \url{https://aclanthology.org/2023.ccl-3.31/}.

\bibitem[Syvokon et~al.(2023)Syvokon, Nahorna, Kuchmiichuk, and Osidach]{syvokon-etal-2023-ua}
Oleksiy Syvokon, Olena Nahorna, Pavlo Kuchmiichuk, and Nastasiia Osidach.
\newblock {UA-GEC: Grammatical Error Correction and Fluency Corpus for the Ukrainian Language}.
\newblock In Mariana Romanyshyn, editor, \emph{Proceedings of the Second Ukrainian Natural Language Processing Workshop (UNLP)}, pages 96--102, Dubrovnik, Croatia, 5 2023. Association for Computational Linguistics.
\newblock \doi{10.18653/v1/2023.unlp-1.12}.
\newblock URL \url{https://aclanthology.org/2023.unlp-1.12/}.

\bibitem[Tan et~al.(2023)Tan, Yang, and Xu]{tan-etal-2023-focal}
Minghuan Tan, Min Yang, and Ruifeng Xu.
\newblock {Focal Training and Tagger Decouple for Grammatical Error Correction}.
\newblock In Anna Rogers, Jordan Boyd-Graber, and Naoaki Okazaki, editors, \emph{Findings of the Association for Computational Linguistics: ACL 2023}, pages 5978--5985, Toronto, Canada, 7 2023. Association for Computational Linguistics.
\newblock \doi{10.18653/v1/2023.findings-acl.370}.
\newblock URL \url{https://aclanthology.org/2023.findings-acl.370}.

\bibitem[Tang et~al.(2023)Tang, Wu, and Wu]{tang-etal-2023-pre}
Chenming Tang, Xiuyu Wu, and Yunfang Wu.
\newblock {Are Pre-trained Language Models Useful for Model Ensemble in Chinese Grammatical Error Correction?}
\newblock In Anna Rogers, Jordan Boyd-Graber, and Naoaki Okazaki, editors, \emph{Proceedings of the 61st Annual Meeting of the Association for Computational Linguistics (Volume 2: Short Papers)}, pages 893--901, Toronto, Canada, 7 2023. Association for Computational Linguistics.
\newblock \doi{10.18653/v1/2023.acl-short.77}.
\newblock URL \url{https://aclanthology.org/2023.acl-short.77}.

\bibitem[Tseng et~al.(2015)Tseng, Lee, Chang, and Chen]{tseng-etal-2015-introduction}
Yuen-Hsien Tseng, Lung-Hao Lee, Li-Ping Chang, and Hsin-Hsi Chen.
\newblock {Introduction to SIGHAN 2015 Bake-off for Chinese Spelling Check}.
\newblock In Liang-Chih Yu, Zhifang Sui, Yue Zhang, and Vincent Ng, editors, \emph{Proceedings of the Eighth SIGHAN Workshop on Chinese Language Processing}, pages 32--37, Beijing, China, 7 2015. Association for Computational Linguistics.
\newblock \doi{10.18653/v1/W15-3106}.
\newblock URL \url{https://aclanthology.org/W15-3106/}.

\bibitem[Wang et~al.(2022)Wang, Duan, Wu, Che, Chen, and Hu]{wang-etal-2022-cctc}
Baoxin Wang, Xingyi Duan, Dayong Wu, Wanxiang Che, Zhigang Chen, and Guoping Hu.
\newblock {CCTC: A Cross-Sentence Chinese Text Correction Dataset for Native Speakers}.
\newblock In Nicoletta Calzolari, Chu-Ren Huang, Hansaem Kim, James Pustejovsky, Leo Wanner, Key-Sun Choi, Pum-Mo Ryu, Hsin-Hsi Chen, Lucia Donatelli, Heng Ji, Sadao Kurohashi, Patrizia Paggio, Nianwen Xue, Seokhwan Kim, Younggyun Hahm, Zhong He, Tony~Kyungil Lee, Enrico Santus, Francis Bond, and Seung-Hoon Na, editors, \emph{Proceedings of the 29th International Conference on Computational Linguistics}, pages 3331--3341, Gyeongju, Republic of Korea, 10 2022. International Committee on Computational Linguistics.
\newblock URL \url{https://aclanthology.org/2022.coling-1.294}.

\bibitem[Wang et~al.(2025)Wang, Qiu, Gu, Huang, and Park]{wang-etal-2025-refined}
Junrui Wang, Mengyang Qiu, Yang Gu, Zihao Huang, and Jungyeul Park.
\newblock {Refined Evaluation for End-to-End Grammatical Error Correction Using an Alignment-Based Approach}.
\newblock In Owen Rambow, Leo Wanner, Marianna Apidianaki, Hend Al-Khalifa, Barbara~Di Eugenio, and Steven Schockaert, editors, \emph{Proceedings of the 31st International Conference on Computational Linguistics}, pages 774--785, Abu Dhabi, UAE, 1 2025. Association for Computational Linguistics.
\newblock URL \url{https://aclanthology.org/2025.coling-main.52/}.

\bibitem[Wang et~al.(2019)Wang, Bi, Yan, Wu, Bao, Xia, Peng, and Si]{wang-etal-2019-structbert}
Wei Wang, Bin Bi, Ming Yan, Chen Wu, Zuyi Bao, Jiangnan Xia, Liwei Peng, and Luo Si.
\newblock {StructBERT: Incorporating Language Structures into Pre-training for Deep Language Understanding}, 2019.
\newblock URL \url{https://arxiv.org/abs/1908.04577}.

\bibitem[Wang et~al.(2021{\natexlab{a}})Wang, Kong, Yang, Wang, Lu, Hu, He, Liu, Chen, Yang, and Sun]{wang-etal-2021-yaclc}
Yingying Wang, Cunliang Kong, Liner Yang, Yijun Wang, Xiaorong Lu, Renfen Hu, Shan He, Zhenghao Liu, Yun Chen, Erhong Yang, and Maosong Sun.
\newblock {YACLC: A Chinese Learner Corpus with Multidimensional Annotation}, 2021{\natexlab{a}}.

\bibitem[Wang et~al.(2024)Wang, Wang, Liu, Wu, and Che]{wang-etal-2024-lm}
Yixuan Wang, Baoxin Wang, Yijun Liu, Dayong Wu, and Wanxiang Che.
\newblock {LM-Combiner: A Contextual Rewriting Model for Chinese Grammatical Error Correction}.
\newblock In Nicoletta Calzolari, Min-Yen Kan, Veronique Hoste, Alessandro Lenci, Sakriani Sakti, and Nianwen Xue, editors, \emph{Proceedings of the 2024 Joint International Conference on Computational Linguistics, Language Resources and Evaluation (LREC-COLING 2024)}, pages 10675--10685, Torino, Italia, 5 2024. ELRA and ICCL.
\newblock URL \url{https://aclanthology.org/2024.lrec-main.934/}.

\bibitem[Wang and Shih(2018)]{wang-shih-2018-hybrid}
Yiyi Wang and Chilin Shih.
\newblock {A Hybrid Approach Combining Statistical Knowledge with Conditional Random Fields for Chinese Grammatical Error Detection}.
\newblock In Yuen-Hsien Tseng, Hsin-Hsi Chen, Vincent Ng, and Mamoru Komachi, editors, \emph{Proceedings of the 5th Workshop on Natural Language Processing Techniques for Educational Applications}, pages 194--198, Melbourne, Australia, 7 2018. Association for Computational Linguistics.
\newblock \doi{10.18653/v1/W18-3728}.
\newblock URL \url{https://aclanthology.org/W18-3728}.

\bibitem[Wang et~al.(2021{\natexlab{b}})Wang, Wang, Dang, Liu, and Liu]{wang-etal-2021-comprehensive}
Yu~Wang, Yuelin Wang, Kai Dang, Jie Liu, and Zhuo Liu.
\newblock {A Comprehensive Survey of Grammatical Error Correction}.
\newblock \emph{ACM Transactions on Intelligent Systems and Technology}, 12\penalty0 (5):\penalty0 1--51, 12 2021{\natexlab{b}}.
\newblock ISSN 2157-6904.
\newblock \doi{10.1145/3474840}.
\newblock URL \url{https://doi.org/10.1145/3474840}.

\bibitem[Wu et~al.(2013)Wu, Liu, and Lee]{wu-etal-2013-chinese}
Shih-Hung Wu, Chao-Lin Liu, and Lung-Hao Lee.
\newblock {Chinese Spelling Check Evaluation at SIGHAN Bake-off 2013}.
\newblock In Liang-Chih Yu, Yuen-Hsien Tseng, Jingbo Zhu, and Fuji Ren, editors, \emph{Proceedings of the Seventh SIGHAN Workshop on Chinese Language Processing}, pages 35--42, Nagoya, Japan, 10 2013. Asian Federation of Natural Language Processing.
\newblock URL \url{https://aclanthology.org/W13-4406/}.

\bibitem[Wu et~al.(2023)Wu, Fang, Zhang, Wang, Zhang, and Chen]{wu-etal-2023-tlm}
Yangjun Wu, Kebin Fang, Dongxiang Zhang, Han Wang, Hao Zhang, and Gang Chen.
\newblock {TLM: Token-Level Masking for Transformers}.
\newblock In Houda Bouamor, Juan Pino, and Kalika Bali, editors, \emph{Proceedings of the 2023 Conference on Empirical Methods in Natural Language Processing}, pages 14099--14111, Singapore, 12 2023. Association for Computational Linguistics.
\newblock \doi{10.18653/v1/2023.emnlp-main.871}.
\newblock URL \url{https://aclanthology.org/2023.emnlp-main.871}.

\bibitem[Wu et~al.(2015)Wu, Zhuang, Jiang, and Li]{wu-etal-2015-research}
YeFan Wu, Runbo Zhuang, Ying Jiang, and Fan Li.
\newblock {Research and realization of Chinese text semantic correction Based on Rule}.
\newblock In \emph{Proceedings of the 2015 3rd International Conference on Education, Management, Arts, Economics and Social Science}, pages 1394--1404, Changsha, China, 2015. Atlantis Press.
\newblock ISBN 978-94-6252-155-1.
\newblock \doi{10.2991/icemaess-15.2016.287}.
\newblock URL \url{https://doi.org/10.2991/icemaess-15.2016.287}.

\bibitem[Xu et~al.(2022)Xu, Wu, Peng, Fu, and Cai]{xu-etal-2022-fcgec}
Lvxiaowei Xu, Jianwang Wu, Jiawei Peng, Jiayu Fu, and Ming Cai.
\newblock {FCGEC: Fine-Grained Corpus for Chinese Grammatical Error Correction}.
\newblock In \emph{Findings of the Association for Computational Linguistics: EMNLP 2022}, pages 1900--1918, Abu Dhabi, United Arab Emirates, 12 2022. Association for Computational Linguistics.
\newblock \doi{10.18653/v1/2022.findings-emnlp.137}.
\newblock URL \url{https://aclanthology.org/2022.findings-emnlp.137}.

\bibitem[Yang and Quan(2024)]{yang-quan-2024-alirector}
Haihui Yang and Xiaojun Quan.
\newblock {Alirector: Alignment-Enhanced Chinese Grammatical Error Corrector}.
\newblock In Lun-Wei Ku, Andre Martins, and Vivek Srikumar, editors, \emph{Findings of the Association for Computational Linguistics: ACL 2024}, pages 2531--2546, Bangkok, Thailand, 8 2024. Association for Computational Linguistics.
\newblock \doi{10.18653/v1/2024.findings-acl.148}.
\newblock URL \url{https://aclanthology.org/2024.findings-acl.148/}.

\bibitem[Yang et~al.(2016)Yang, Peng, Wang, Zhang, and Zhang]{yang-etal-2016-chinese}
Jinnan Yang, Bo~Peng, Jin Wang, Jixian Zhang, and Xuejie Zhang.
\newblock {Chinese Grammatical Error Diagnosis Using Single Word Embedding}.
\newblock In Hsin-Hsi Chen, Yuen-Hsien Tseng, Vincent Ng, and Xiaofei Lu, editors, \emph{Proceedings of the 3rd Workshop on Natural Language Processing Techniques for Educational Applications (NLPTEA2016)}, pages 155--161, Osaka, Japan, 12 2016. The COLING 2016 Organizing Committee.
\newblock URL \url{https://aclanthology.org/W16-4920}.

\bibitem[Yin et~al.(2023)Yin, Wan, Zhang, Yu, and Yu]{yin-etal-2023-overview}
Xunjian Yin, Xiaojun Wan, Dan Zhang, Linlin Yu, and Long Yu.
\newblock {Overview of NLPCC 2023 Shared Task: Chinese Spelling Check}.
\newblock In \emph{Natural Language Processing and Chinese Computing: 12th National CCF Conference, NLPCC 2023, Foshan, China, October 12–15, 2023, Proceedings, Part III}, page 337–345, Berlin, Heidelberg, 2023. Springer-Verlag.
\newblock ISBN 978-3-031-44698-6.
\newblock \doi{10.1007/978-3-031-44699-3{\_}30}.
\newblock URL \url{https://doi.org/10.1007/978-3-031-44699-3_30}.

\bibitem[Yoon et~al.(2023)Yoon, Park, Kim, Cho, Park, Kim, Seo, and Oh]{yoon-etal-2023-towards}
Soyoung Yoon, Sungjoon Park, Gyuwan Kim, Junhee Cho, Kihyo Park, Gyu~Tae Kim, Minjoon Seo, and Alice Oh.
\newblock {Towards standardizing Korean Grammatical Error Correction: Datasets and Annotation}.
\newblock In \emph{Proceedings of the 61st Annual Meeting of the Association for Computational Linguistics (Volume 1: Long Papers)}, pages 6713--6742, Toronto, Canada, 7 2023. Association for Computational Linguistics.
\newblock URL \url{https://aclanthology.org/2023.acl-long.371}.

\bibitem[Yu and Chen(2012)]{yu-chen-2012-detecting}
Chi-Hsin Yu and Hsin-Hsi Chen.
\newblock {Detecting Word Ordering Errors in Chinese Sentences for Learning Chinese as a Foreign Language}.
\newblock In Martin Kay and Christian Boitet, editors, \emph{Proceedings of COLING 2012}, pages 3003--3018, Mumbai, India, 12 2012. The COLING 2012 Organizing Committee.
\newblock URL \url{https://aclanthology.org/C12-1184}.

\bibitem[Yu et~al.(2014{\natexlab{a}})Yu, Lee, and Chang]{yu-lee-chang-2014-overview}
Liang-Chih Yu, Lung-Hao Lee, and Li-Ping Chang.
\newblock {Overview of Grammatical Error Diagnosis for Learning Chinese as a Foreign Language}.
\newblock In Chen-Chung Liu, Hiroaki Ogata, Siu~Cheung Kong, and Akihiro Kashihara, editors, \emph{Proceedings of the 22nd International Conference on Computers in Education (ICCE 2014)}, pages 42--47, Nara, Japan, 2014{\natexlab{a}}. Asia-Pacific Society for Computers in Education.

\bibitem[Yu et~al.(2014{\natexlab{b}})Yu, Lee, Tseng, and Chen]{yu-etal-2014-overview}
Liang-Chih Yu, Lung-Hao Lee, Yuen-Hsien Tseng, and Hsin-Hsi Chen.
\newblock {Overview of SIGHAN 2014 Bake-off for Chinese Spelling Check}.
\newblock In Le~Sun, Chengqing Zong, Min Zhang, and Gina-Anne Levow, editors, \emph{Proceedings of the Third CIPS-SIGHAN Joint Conference on Chinese Language Processing}, pages 126--132, Wuhan, China, 10 2014{\natexlab{b}}. Association for Computational Linguistics.
\newblock \doi{10.3115/v1/W14-6820}.
\newblock URL \url{https://aclanthology.org/W14-6820/}.

\bibitem[Zan et~al.(2020)Zan, Han, Huang, Yan, Wang, and Han]{zan-etal-2020-chinese}
Hongying Zan, Yangchao Han, Haotian Huang, Yingjie Yan, Yuke Wang, and Yingjie Han.
\newblock {Chinese Grammatical Errors Diagnosis System Based on BERT at NLPTEA-2020 CGED Shared Task}.
\newblock In Erhong YANG, Endong XUN, Baolin ZHANG, and Gaoqi RAO, editors, \emph{Proceedings of the 6th Workshop on Natural Language Processing Techniques for Educational Applications}, pages 102--107, Suzhou, China, 12 2020. Association for Computational Linguistics.
\newblock \doi{10.18653/v1/2020.nlptea-1.14}.
\newblock URL \url{https://aclanthology.org/2020.nlptea-1.14/}.

\bibitem[Zeng et~al.(2024)Zeng, Kuang, Qiu, Song, and Park]{zeng-etal-2024-evaluating-prompting}
Min Zeng, Jiexin Kuang, Mengyang Qiu, Jayoung Song, and Jungyeul Park.
\newblock {Evaluating Prompting Strategies for Grammatical Error Correction Based on Language Proficiency}.
\newblock In Nicoletta Calzolari, Min-Yen Kan, Veronique Hoste, Alessandro Lenci, Sakriani Sakti, and Nianwen Xue, editors, \emph{Proceedings of the 2024 Joint International Conference on Computational Linguistics, Language Resources and Evaluation (LREC-COLING 2024)}, pages 6426--6430, Torino, Italy, 5 2024. ELRA and ICCL.
\newblock URL \url{https://aclanthology.org/2024.lrec-main.569}.

\bibitem[Zhang et~al.(2022{\natexlab{a}})Zhang, Li, Bao, Li, Zhang, Li, Huang, and Zhang]{zhang-etal-2022-mucgec}
Yue Zhang, Zhenghua Li, Zuyi Bao, Jiacheng Li, Bo~Zhang, Chen Li, Fei Huang, and Min Zhang.
\newblock {MuCGEC: a Multi-Reference Multi-Source Evaluation Dataset for Chinese Grammatical Error Correction}.
\newblock In \emph{Proceedings of the 2022 Conference of the North American Chapter of the Association for Computational Linguistics: Human Language Technologies}, pages 3118--3130, Seattle, United States, 7 2022{\natexlab{a}}. Association for Computational Linguistics.
\newblock \doi{10.18653/v1/2022.naacl-main.227}.
\newblock URL \url{https://aclanthology.org/2022.naacl-main.227}.

\bibitem[Zhang et~al.(2022{\natexlab{b}})Zhang, Zhang, Li, Bao, Li, and Zhang]{zhang-etal-2022-syngec}
Yue Zhang, Bo~Zhang, Zhenghua Li, Zuyi Bao, Chen Li, and Min Zhang.
\newblock {SynGEC: Syntax-Enhanced Grammatical Error Correction with a Tailored GEC-Oriented Parser}.
\newblock In Yoav Goldberg, Zornitsa Kozareva, and Yue Zhang, editors, \emph{Proceedings of the 2022 Conference on Empirical Methods in Natural Language Processing}, pages 2518--2531, Abu Dhabi, United Arab Emirates, 12 2022{\natexlab{b}}. Association for Computational Linguistics.
\newblock \doi{10.18653/v1/2022.emnlp-main.162}.
\newblock URL \url{https://aclanthology.org/2022.emnlp-main.162/}.

\bibitem[Zhang et~al.(2023)Zhang, Zhang, Jiang, Li, Li, Huang, and Zhang]{zhang-etal-2023-nasgec}
Yue Zhang, Bo~Zhang, Haochen Jiang, Zhenghua Li, Chen Li, Fei Huang, and Min Zhang.
\newblock {NaSGEC: a Multi-Domain Chinese Grammatical Error Correction Dataset from Native Speaker Texts}.
\newblock In \emph{Findings of the Association for Computational Linguistics: ACL 2023}, pages 9935--9951, Toronto, Canada, 7 2023. Association for Computational Linguistics.
\newblock \doi{10.18653/v1/2023.findings-acl.630}.
\newblock URL \url{https://aclanthology.org/2023.findings-acl.630}.

\bibitem[Zhao et~al.(2019)Zhao, Wang, Shen, Jia, and Liu]{zhao-etal-2019-improving}
Wei Zhao, Liang Wang, Kewei Shen, Ruoyu Jia, and Jingming Liu.
\newblock {Improving Grammatical Error Correction via Pre-Training a Copy-Augmented Architecture with Unlabeled Data}.
\newblock In Jill Burstein, Christy Doran, and Thamar Solorio, editors, \emph{Proceedings of the 2019 Conference of the North American Chapter of the Association for Computational Linguistics: Human Language Technologies, Volume 1 (Long and Short Papers)}, pages 156--165, Minneapolis, Minnesota, 6 2019. Association for Computational Linguistics.
\newblock \doi{10.18653/v1/N19-1014}.
\newblock URL \url{https://aclanthology.org/N19-1014/}.

\bibitem[Zhao et~al.(2015)Zhao, Komachi, and Ishikawa]{zhao-etal-2015-improving}
Yinchen Zhao, Mamoru Komachi, and Hiroshi Ishikawa.
\newblock {Improving Chinese Grammatical Error Correction with Corpus Augmentation and Hierarchical Phrase-based Statistical Machine Translation}.
\newblock In Hsin-Hsi Chen, Yuen-Hsien Tseng, Yuji Matsumoto, and Lung~Hsiang Wong, editors, \emph{Proceedings of the 2nd Workshop on Natural Language Processing Techniques for Educational Applications}, pages 111--116, Beijing, China, 7 2015. Association for Computational Linguistics.
\newblock \doi{10.18653/v1/W15-4417}.
\newblock URL \url{https://aclanthology.org/W15-4417}.

\bibitem[Zhao et~al.(2018)Zhao, Jiang, Sun, and Wan]{zhao-etal-2018-nlpcc}
Yuanyuan Zhao, Nan Jiang, Weiwei Sun, and Xiaojun Wan.
\newblock {Overview of the NLPCC 2018 Shared Task: Grammatical Error Correction}.
\newblock In Min Zhang, Vincent Ng, Dongyan Zhao, Sujian Li, and Hongying Zan, editors, \emph{Natural Language Processing and Chinese Computing}, pages 439--445, Cham, 2018. Springer International Publishing.
\newblock ISBN 978-3-319-99501-4.

\bibitem[Zhao and Wang(2020)]{zhao-wang-2020-maskgec}
Zewei Zhao and Houfeng Wang.
\newblock {MaskGEC: Improving Neural Grammatical Error Correction via Dynamic Masking}.
\newblock \emph{Proceedings of the AAAI Conference on Artificial Intelligence}, 34\penalty0 (01):\penalty0 1226--1233, 2020.
\newblock \doi{10.1609/aaai.v34i01.5476}.
\newblock URL \url{https://ojs.aaai.org/index.php/AAAI/article/view/5476}.

\bibitem[Zheng et~al.(2016)Zheng, Che, Guo, and Liu]{zheng-etal-2016-chinese}
Bo~Zheng, Wanxiang Che, Jiang Guo, and Ting Liu.
\newblock {Chinese Grammatical Error Diagnosis with Long Short-Term Memory Networks}.
\newblock In Hsin-Hsi Chen, Yuen-Hsien Tseng, Vincent Ng, and Xiaofei Lu, editors, \emph{Proceedings of the 3rd Workshop on Natural Language Processing Techniques for Educational Applications (NLPTEA2016)}, pages 49--56, Osaka, Japan, 12 2016. The COLING 2016 Organizing Committee.
\newblock URL \url{https://aclanthology.org/W16-4907}.

\bibitem[Zhou et~al.(2018)Zhou, Li, Liu, Bao, Xu, and Li]{zhou-etal-2018-chinese}
Junpei Zhou, Chen Li, Hengyou Liu, Zuyi Bao, Guangwei Xu, and Linlin Li.
\newblock {Chinese Grammatical Error Correction Using Statistical and Neural Models}.
\newblock In Min Zhang, Vincent Ng, Dongyan Zhao, Sujian Li, and Hongying Zan, editors, \emph{Natural Language Processing and Chinese Computing}, pages 117--128, Cham, 2018. Springer International Publishing.
\newblock ISBN 978-3-319-99501-4.

\bibitem[Zhou et~al.(2020)Zhou, Ge, Wei, Zhou, and Xu]{zhou-etal-2020-scheduled}
Wangchunshu Zhou, Tao Ge, Furu Wei, Ming Zhou, and Ke~Xu.
\newblock {Scheduled DropHead: A Regularization Method for Transformer Models}.
\newblock In Trevor Cohn, Yulan He, and Yang Liu, editors, \emph{Findings of the Association for Computational Linguistics: EMNLP 2020}, pages 1971--1980, Online, 11 2020. Association for Computational Linguistics.
\newblock \doi{10.18653/v1/2020.findings-emnlp.178}.
\newblock URL \url{https://aclanthology.org/2020.findings-emnlp.178/}.

\end{thebibliography}

\appendix

\section{Automatic error annotation in CGEC} \label{sec:annotation}

This section reviews automatic error annotation in Chinese grammatical error correction and situates recent developments within the broader landscape of CGEC annotation design. We first examine previous automatic annotation schemes, with particular attention to \texttt{ChERRANT}\footnote{\url{https://github.com/HillZhang1999/MuCGEC/tree/main/scorers}} \citep[Chinese \texttt{errant};][]{zhang-etal-2022-mucgec}, as a representative toolkit for edit-based annotation and evaluation. We then present the refined annotation scheme proposed by \citet{gu-etal-2025-improving}, which extends existing practice by introducing a more linguistically informed error typology tailored to recurrent patterns in Chinese learner writing. Next, we summarize the implementation of that scheme in a revised CGEC system that generates \texttt{m2}-style annotations. We conclude with a discussion of the strengths and limitations of these annotation approaches, with particular attention to segmentation, label granularity, and the extent to which automatic annotation can capture Chinese-specific error phenomena.

\subsection{\texttt{ChERRANT} as a prior automatic annotation framework}\label{r1-p15}

This subsection reviews \texttt{ChERRANT} as a representative prior framework for automatic annotation in Chinese grammatical error correction.

\texttt{ChERRANT} \citep{zhang-etal-2022-mucgec} is a toolkit for automatic annotation and evaluation in CGEC. Given a source sentence and its corrected counterpart, it derives edit annotations in an \texttt{m2}-style format, making it possible to support large-scale automatic scoring in a way broadly comparable to English \texttt{errant} \citep{bryant-etal-2017-automatic}. In this sense, \texttt{ChERRANT} can be understood as a Chinese adaptation of the edit-based annotation and evaluation paradigm.

Like \texttt{errant}, \texttt{ChERRANT} organizes edits around a small set of core operations. Its main labels are substitution (\texttt{S}), missing (\texttt{M}), and redundant (\texttt{R}), summarized in Table~\ref{errant-mru}. These labels correspond broadly to replacement, insertion, and deletion operations in English-oriented GEC frameworks, although the terminology differs. The framework also includes a dedicated word-order label (\texttt{W}), reflecting the importance of reordering phenomena in Chinese learner writing.

A notable feature of \texttt{ChERRANT} is that it supports both character-based and word-based annotation, the two most common representational choices in Chinese language processing. Figure~\ref{chinese-m2} illustrates these two formats. Character-based annotation operates directly on individual Chinese characters, whereas word-based annotation groups characters into segmented words and can assign part-of-speech labels to the resulting edits. In this study, we focus on the word-based setting, since it is more informative for linguistically structured error analysis and aligns more closely with later refinements of the annotation scheme.

For word-based annotation, \texttt{ChERRANT} relies on segmentation and part-of-speech tagging from the Language Technology Platform (LTP) \citep{che-etal-2010-ltp}\footnote{\url{https://github.com/HIT-SCIR/ltp}}. The resulting tags are converted to Universal POS labels in the output \citep{petrov-das-mcdonald:2012:LREC}. Beyond core edit operations, \texttt{ChERRANT} includes mechanisms intended to capture Chinese-specific phenomena. In particular, it detects spelling-related errors by comparing source and target spans at the character level and estimating similarity in pronunciation and visual shape. Such cases are labeled as \texttt{S:SPELL}, that is, as substitution errors involving orthographic confusion. The framework also incorporates heuristic rules for word-order errors, which are especially relevant in Chinese and are not always naturally reducible to simple insertion or deletion patterns.

At the same time, the design of \texttt{ChERRANT} reflects the limitations of a relatively compact edit inventory. While the framework provides an efficient and practical basis for automatic annotation, broad labels such as \texttt{S:SPELL} often collapse together several linguistically distinct error types, including pronunciation-based confusions, shape-based confusions, and mixed cases. Similarly, although the \texttt{W} label captures reordering, it does not always distinguish clearly between local character-order problems and larger constituent-level word-order errors. These limitations motivate later refinements that seek to preserve the advantages of automatic \texttt{m2}-style annotation while providing more linguistically informative categories for Chinese learner errors.

\begin{figure}[ht!]

\begin{subfigure}[b]{\textwidth}    
\centering
\footnotesize{
\begin{CJK*}{UTF8}{gbsn}
\begin{tabular}{l}
\texttt{S 这~ 歌~ 也~ 最~ 早~ 中文~ 的~ 歌~ 。}\\
\texttt{T0-A0 这歌~ 最~ 早~ 也~ 是~ 中文~ 歌~ 。}\\
\texttt{A 2 5|||W|||最~ 早~ 也|||REQUIRED|||-NONE-|||0}\\
\texttt{A 5 5|||M:VERB|||是|||REQUIRED|||-NONE-|||0}\\
\texttt{A 6 7|||R:AUX|||-NONE-|||REQUIRED|||-NONE-|||0}\\
(`Even this song was originally a Chinese song.') \\
\end{tabular}
\end{CJK*}
}
\caption{Word based m2}
\label{word-based-m2}
\end{subfigure}
\hfill

\begin{subfigure}[b]{\textwidth}    
\centering
\footnotesize{
\begin{CJK*}{UTF8}{gbsn}
\begin{tabular}{l}
\texttt{S 这~ 歌~ 也~ 最~ 早~ 中~ 文~ 的~ 歌~ 。}\\
\texttt{T0-A0 这~ 歌~ 最~ 早~ 也~ 是~ 中~ 文~ 歌~ 。}\\
\texttt{A 2 5|||W|||最~ 早~ 也|||REQUIRED|||-NONE-|||0}\\
\texttt{A 5 5|||M|||是|||REQUIRED|||-NONE-|||0}\\
\texttt{A 7 8|||R|||-NONE-|||REQUIRED|||-NONE-|||0}\\
(`Even this song was originally a Chinese song.') \\
\end{tabular}
\end{CJK*}
}
\caption{Character based m2}
\label{character-based-m2}
\end{subfigure}
\caption{Examples of the Chinese m2 file}\label{chinese-m2}
\end{figure}

\begin{table}[ht!]
{
\centering
\begin{tabular}{p{0.01\textwidth} p{0.9\textwidth}}\hline 
\texttt{S} & \textit{Substitute} refers to annotations where a word or phrase in the original text has been replaced by a different word or phrase in the corrected text with its \texttt{POS} label, and \texttt{Other} otherwise. \textit{Replacement} in English. \\ \hdashline
\texttt{M} &  \textit{Missing} refers to annotations where a word or phrase is missing from the original text compared to the corrected text. In other words, something that should have been present in the corrected text is missing with its \texttt{POS} label, and \texttt{Other} otherwise.\\ \hdashline
\texttt{R}  & \textit{Redundant} refers to annotations where a word or phrase in the original text is unnecessary or not fluent compared to the corrected text with its \texttt{POS} label, and \texttt{Other} otherwise.  \textit{Unnecessary} in English. \\ \hline
\end{tabular}
}
\caption{\texttt{S} (substitute--replace), \texttt{M} (missing), and \texttt{R} (redundant--unnecessary) error types}\label{errant-mru}
\end{table}

\subsection{Refined annotation scheme for CGEC}\label{subsec:new-annotation}

\citet{gu-etal-2025-improving} proposed a refined automatic annotation scheme for CGEC that extends earlier edit-based frameworks with a more linguistically informed error typology. The main motivation is that several recurrent error patterns in Chinese learner writing are only coarsely represented in existing automatic annotation systems. In particular, Chinese learner errors often involve character confusion driven by pronunciation similarity, visual similarity, or character order, as well as the misuse or omission of functionally important particles such as \textit{de}. To better capture these phenomena, the refined scheme introduces a set of error categories tailored to Chinese-specific orthographic and grammatical properties, while preserving compatibility with \texttt{m2}-style edit annotation.

\paragraph{Spelling mistakes in analog vs digital writing}

Our new error typology is closely related to the Chinese analog of ``spelling errors'' known as \zh{错别字}, which literally means \textit{incorrect or other character}.
An \textit{incorrect character} or \zh{错字} \textit{cuò zì} is created when the writer makes an error in the strokes, producing a character that does not exist in the Chinese language. 
This type of mistake is rare in digital writing due to the fixed set of characters available in most typing and rendering systems. However, digital text is not entirely immune to these errors--writers may inadvertently select a similar-looking character from another language. For example, in the MuCGEC dataset, we encountered a spelling error where the character for \zh{凉} \textit{liáng} (`cool') was mistakenly written as \np{涼}\label{r1-p18a}, which is not valid in Simplified Chinese but is still used for Japanese and Traditional Chinese.

Chinese spelling errors are frequently attributed to the reliance on pinyin-based input methods, where homophones and visually analogous characters can lead to inadvertent substitution or incorrect character usage \citep{liu-etal-2010-visually,deng-hu-2022-examination}. This usually produces \zh{别字} \textit{bié zì} or \textit{other character}, where the writer produces a valid Chinese character that is incorrect in the given context.

\paragraph{Similar pronunciation}\label{r1-p18b}
Chinese contains many homophonous or near-homophonous characters, which makes character choice error-prone, especially under pinyin-based input where multiple candidates share the same romanization. We illustrate three common patterns: (i) confusions between distinct, well-formed words with identical pronunciations (e.g., \zh{权力} `power' vs.\ \zh{权利} `right' in \eqref{mixing-up-homophones}); (ii) character substitutions inside a multi-character word that yield an ill-formed string (e.g., \zh{时 后} vs.\ \zh{时候} in \eqref{mixing-up-one-character-in-a-word-with-a-homophone}); and (iii) substitutions between characters with the same syllable but different tones (e.g., \zh{一前} vs.\ \zh{以前} in \eqref{mixing-up-characters-with-different-tone}), which are difficult to prevent because tone is typically not encoded in keyboard input.

\begin{exe}
\ex \label{mixing-up-homophones}
\begin{xlist}
\ex \glll \zh{每} \zh{个} \zh{抽烟} \zh{的} \zh{人} \zh{都} \zh{有} \zh{这样} \zh{的} \zh{\textcolor{red}{权力}} \zh{。}\\
\textit{měi} \textit{gè} \textit{chōu-yān} \textit{de} \textit{rén} \textit{dōu} \textit{yǒu} \textit{zhè-yàng} \textit{de} \textcolor{red}{\textit{quán-lì}}.\\
every CL smoke REL person all have such REL \textcolor{red}{power} .\\
\trans `Every smoker has this power.'
\ex
\glll \zh{每} \zh{个} \zh{抽烟} \zh{的} \zh{人} \zh{都} \zh{有} \zh{这样} \zh{的} \zh{\textcolor{blue}{权利}} \zh{。}\\
\textit{měi} \textit{gè} \textit{chōu-yān} \textit{de} \textit{rén} \textit{dōu} \textit{yǒu} \textit{zhè-yàng} \textit{de} \textcolor{blue}{\textit{quán-lì}}.\\
every CL smoke REL person all have such REL \textcolor{blue}{right} .\\
\trans `Every smoker has this right.'
\end{xlist}
\end{exe}

This mistake can also happen at the character-level, when one character is substituted by a homophone, resulting in an invalid word. 
In \eqref{mixing-up-one-character-in-a-word-with-a-homophone}, \zh{时 后} \textit{shí-hòu} (`time after') is not a valid word, and a different character pronounced \textit{hòu} is needed to form the word \zh{时候}  \textit{shí-hòu} (`when' or 'time of').

\begin{exe}
\ex \label{mixing-up-one-character-in-a-word-with-a-homophone}
\begin{xlist}
\ex \label{wrong-shi-bou}
\glll \zh{交} \zh{朋友} \zh{的} \zh{\textcolor{red}{时 后}} \zh{，} \zh{很} \zh{可能} \zh{会} \zh{碰到} \zh{矛盾} \zh{。}\\
\textit{jiāo} \textit{péng-yǒu} \textit{de} \textcolor{red}{\textit{shí hòu}} \textit{,} \textit{hěn} \textit{kě-néng} \textit{huì} \textit{pèng-dào} \textit{máo-dùn}.\\
make friend REL \textcolor{red}{time after} , very likely will encounter conflict .\\
\trans `When making friends, it's highly likely to run into conflicts.'
\ex
\glll \zh{交} \zh{朋友} \zh{的} \zh{\textcolor{blue}{时候}} \zh{，} \zh{很} \zh{可能} \zh{会} \zh{碰到} \zh{矛盾} \zh{。}\\
\textit{jiāo} \textit{péng-yǒu} \textit{de} \textcolor{blue}{\textit{shí hòu}} \textit{,} \textit{hěn} \textit{kě-néng} \textit{huì} \textit{pèng-dào} \textit{máo-dùn}.\\
make friend REL \textcolor{blue}{when} , very likely will encounter conflict .\\
\trans `When making friends, it's highly likely to run into conflicts.'
\end{xlist}
\end{exe}

Similarly, this substitution may also occur between characters that have the same sound but different tones. In \eqref{mixing-up-characters-with-different-tone}, we observe a case of \zh{一} \textit{yī} (first tone) being used instead of \zh{以} \textit{yǐ} (third tone), resulting an invalid expression \zh{一前} \textit{yī qián} (`one before') instead of word \zh{以前} \textit{yǐ qián} (`before'). This type of error is particularly challenging because tones cannot be easily typed on most keyboards, so the onus is on the writer to select the correct characters from multiple possible tone combinations even after inputting the correct pinyin.

\begin{exe}
\ex \label{mixing-up-characters-with-different-tone}
\begin{xlist}
\ex
\glll \zh{我} \zh{\textcolor{red}{一 前}} \zh{没} \zh{住} \zh{过} \zh{五星级} \zh{旅馆} \zh{，} \zh{所以} \zh{我} \zh{很} \zh{惊讶} \zh{。}\\
\textit{wǒ} \textcolor{red}{\textit{yī qián}} \textit{méi} \textit{zhù} \textit{guò} \textit{wǔ-xīng-jí} \textit{lu-guǎn} \textit{,} \textit{suǒ-yǐ} \textit{wǒ} \textit{hěn} \textit{jīng-yà}.\\
I \textcolor{red}{one before} NEG live EXP five-star hotel , so I very surprised .\\
\trans `I haven't stayed in five-star hotels before, so I am very surprised.'
\ex
\glll \zh{我} \zh{\textcolor{blue}{以前}} \zh{没} \zh{住} \zh{过} \zh{五星级} \zh{旅馆} \zh{，} \zh{所以} \zh{我} \zh{很} \zh{惊讶} \zh{。}\\
\textit{wǒ} \textcolor{blue}{\textit{yǐ qián}} \textit{méi} \textit{zhù} \textit{guò} \textit{wǔ-xīng-jí} \textit{lu-guǎn} \textit{,} \textit{suǒ-yǐ} \textit{wǒ} \textit{hěn} \textit{jīng-yà}.\\
I \textcolor{blue}{before} NEG live EXP five-star hotel , so I very surprised .\\
\trans `I haven't stayed in five-star hotels before, so I am very surprised.'
\end{xlist}
\end{exe}

\paragraph{Similar shapes}
Language learners can also confuse words with similar shapes that are unrelated in meaning or pronunciation. In \eqref{similar-shape}, the character \zh{西} \textit{xī} (`west') is mistakenly used instead of \zh{四} \textit{sì} (`four'), which contains many of the same strokes. Even among the words that are used daily, there are many characters with only subtle differences in appearance, like \zh{人} \textit{rén} (`person') vs. \zh{入} \textit{rù} (`enter') and \zh{己} \textit{jǐ} (`self') vs. \zh{已} \textit{yǐ} (`already').

\begin{exe}
\ex \label{similar-shape}
\begin{xlist}
\ex
\glll \zh{她} \zh{有} \zh{两} \zh{个} \zh{姐姐} \zh{、} \zh{一} \zh{个} \zh{妹妹} \zh{和} \zh{\textcolor{red}{西}} \zh{个} \zh{哥哥} \zh{。}\\
\textit{tā} \textit{yǒu} \textit{liǎng} \textit{gè} \textit{jiě-jie} \textit{,} \textit{yī} \textit{gè} \textit{mèi-mei} \textit{hé} \textcolor{red}{\textit{xī}} \textit{gè} \textit{gē-ge}.\\
she have two CL older-sister , one CL younger-sister and \textcolor{red}{west} CL older-brother .\\
\trans `She has two older sisters, one younger sister, and \textcolor{red}{west} older brothers.'
\ex
\glll \zh{她} \zh{有} \zh{两} \zh{个} \zh{姐姐} \zh{、} \zh{一} \zh{个} \zh{妹妹} \zh{和} \zh{\textcolor{blue}{四}} \zh{个} \zh{哥哥} \zh{。}\\
\textit{tā} \textit{yǒu} \textit{liǎng} \textit{gè} \textit{jiě-jie} \textit{,} \textit{yī} \textit{gè} \textit{mèi-mei} \textit{hé} \textcolor{blue}{\textit{sì}} \textit{gè} \textit{gē-ge}.\\
she have two CL older-sister , one CL younger-sister and \textcolor{blue}{four} CL older-brother .\\
\trans `She has two older sisters, one younger sister, and four older brothers.'
\end{xlist}
\end{exe}

\paragraph{Multifaceted similarity}

In many cases, it can be difficult to separate shape and pronunciation similarities in Chinese characters because these two aspects often overlap. Due to the logographic nature of Chinese, many characters share phonetic components that influence their pronunciation, as well as semantic components that link them to related meanings. This can result in characters that not only look alike but also sound alike or share related meanings. 
For example, the characters \zh{州} \textit{zhōu} (`region' or `province') and \zh{洲} \textit{zhōu} (`continent') share the same pronunciation, look very similar, and both mean something related to locations. However, \zh{州} is typically seen in place names like Suzhou or Hangzhou, whereas \zh{洲} refers specifically to continents, indicated by the water radical. It's easy to see how writers can make the error shown in \eqref{general-similarity}, where \zh{欧州人} \textit{ōu-zhōu rén} (`Europe region people') is mistakenly used instead of \zh{欧洲人} \textit{ōu-zhōu rén} (`Europeans'). 

\begin{exe}
\ex \label{general-similarity}
\begin{xlist}
\ex
\glll \zh{从} \zh{十六} \zh{世纪} \zh{开始} \zh{，} \zh{\textcolor{red}{欧州人}} \zh{就} \zh{抽烟} \zh{。}\\
\textit{cóng} \textit{shí-liù} \textit{shì-jì} \textit{kāi-shǐ} \textit{,} \textcolor{red}{\textit{ōu-zhōu rén}} \textit{jiù} \textit{chōu-yān}.\\
since sixteenth century begin , \textcolor{red}{Europe region people} already smoke .\\
\trans `Since the sixteenth century, people in Europe smoked.'
\ex
\glll \zh{从} \zh{十六} \zh{世纪} \zh{开始} \zh{，} \zh{\textcolor{blue}{欧洲人}} \zh{就} \zh{抽烟} \zh{。}\\
\textit{cóng} \textit{shí-liù} \textit{shì-jì} \textit{kāi-shǐ} \textit{,} \textcolor{blue}{\textit{ōu-zhōu rén}} \textit{jiù} \textit{chōu-yān}.\\
since sixteenth century begin , \textcolor{blue}{Europeans} already smoke .\\
\trans `Since the sixteenth century, Europeans smoked.'
\end{xlist}
\end{exe}

\paragraph{The \textit{de} particles}

The Chinese characters \zh{的} \textit{de}, \zh{地} \textit{de}, and \zh{得} \textit{de} serve as structural particles in different syntactic contexts. \zh{的} is typically used as a possessive or descriptive marker (similar to the English apostrophe-s or “of”), \zh{地} is used to modify verbs (adverbial marker), and \zh{得} introduces complements (usually degree complements). Despite their distinct functions, these three characters are pronounced identically, which makes them a significant challenge for both native speakers and learners of Chinese. 
This type of mistake is often categorized as an auxiliary (\textsc{aux}) or spelling error (\textsc{spell}) in error annotation systems such as \texttt{ChERRANT}. However, given the distinct grammatical roles these characters play, it is more informative to classify these errors under a separate category. 
For instance, in \eqref{de-particle}, the complement-marker character \zh{得} \textit{de} (\textsc{comp}) is incorrectly written in a position where the attributive particle \zh{的} \textit{de} (\textsc{attr}) is required.

\begin{exe}
\ex \label{de-particle}
\begin{xlist}
\ex
\glll \zh{经理} \zh{，} \zh{新} \zh{\textcolor{red}{得}} \zh{计划} \zh{发} \zh{您} \zh{信箱} \zh{了} \zh{，} \zh{您} \zh{看} \zh{了} \zh{吧} \zh{？}\\
\textit{jīng-lǐ} \textit{,} \textit{xīn} \textcolor{red}{\textit{de}} \textit{jì-huà} \textit{fā} \textit{nín} \textit{xìn-xiāng} \textit{le} \textit{,} \textit{nín} \textit{kàn} \textit{le} \textit{ba} \textit{?}\\
manager , new \textcolor{red}{\textsc{comp}} plan send your mailbox PFV , you see PFV MOD ?\\
\trans `Manager, the new plan was sent to your mailbox, you've seen it, right?'
\ex
\glll \zh{经理} \zh{，} \zh{新} \zh{\textcolor{blue}{的}} \zh{计划} \zh{发} \zh{您} \zh{信箱} \zh{了} \zh{，} \zh{您} \zh{看} \zh{了} \zh{吗} \zh{？}\\
\textit{jīng-lǐ} \textit{,} \textit{xīn} \textcolor{blue}{\textit{de}} \textit{jì-huà} \textit{fā} \textit{nín} \textit{xìn-xiāng} \textit{le} \textit{,} \textit{nín} \textit{kàn} \textit{le} \textit{ma} \textit{?}\\
manager , new \textcolor{blue}{\textsc{attr}} plan send your mailbox PFV , you see PFV Q ?\\
\trans `Manager, the new plan was sent to your mailbox, have you seen it?'
\end{xlist}
\end{exe}

\paragraph{Missing \textit{de} particle}

One common error among L2 learners is the omission of the \textit{de} particles (\zh{的}, \zh{地}, \zh{得}). 
In many cases, these particles do not translate directly into other languages as independent words, so it's easy for non-native speakers to omit them. In \eqref{missing-de-participle}, \zh{的} is required between the adjective \zh{简单} \textit{jiǎn-dān} (`simple') and noun \zh{生活} \textit{shēng-huó} (`life') for the resulting noun phrase ``simple life'' to be grammatically-sound, but such a particle does not exist in the English translation.

\begin{exe}
\ex \label{missing-de-participle}
\begin{xlist}
\ex
\glll \zh{简单} \zh{生活} \zh{，} \zh{哪怕} \zh{对} \zh{身体} \zh{还是} \zh{精神} \zh{，} \zh{还} \zh{大有} \zh{裨益} \zh{。}\\
\textit{jiǎn-dān} \textit{shēng-huó} \textit{,} \textit{nǎ-pà} \textit{duì} \textit{shēn-tǐ} \textit{hái-shì} \textit{jīng-shén} \textit{,} \textit{hái} \textit{dà-yǒu} \textit{bì-yì}.\\
simple life , even for body or mind , still great benefit .\\
\trans `A simple life, whether for the body or the mind, is still greatly beneficial.'
\ex
\glll \zh{简单} \zh{\textcolor{blue}{的}} \zh{生活} \zh{，} \zh{无论} \zh{对} \zh{身体} \zh{还是} \zh{精神} \zh{，} \zh{都} \zh{大有} \zh{裨益} \zh{。}\\
\textit{jiǎn-dān} \textcolor{blue}{\textit{de}} \textit{shēng-huó} \textit{,} \textit{wú-lùn} \textit{duì} \textit{shēn-tǐ} \textit{hái-shì} \textit{jīng-shén} \textit{,} \textit{dōu} \textit{dà-yǒu} \textit{bì-yì}.\\
simple \textcolor{blue}{\textsc{attr}} life , whether for body or mind , all great benefit.\\
\trans `A simple life, whether for the body or the mind, is greatly beneficial.'
\end{xlist}
\end{exe}

In \eqref{missing-de-as-modal}, \zh{的} is required at the end of the descriptive clause \zh{不帅，还有点丑但是很会唱歌} \textit{bù shuài, hái yǒu diǎn chǒu dànshì hěn huì chàng-gē} (`not handsome, a bit ugly, but able to sing well'). Here, \zh{的} functions as a clause-marking particle in a headless relative construction: the sequence \zh{那些 ... 的} refers to `those who ...', with the head noun, such as \zh{人} \textit{rén} (`people'), left implicit. Without \zh{的}, the descriptive clause is not properly nominalized as the subject of the sentence. Classifying this error separately from auxiliary or spelling mistakes would therefore provide more accurate feedback to learners and better represent the syntactic nature of the error.

\begin{exe}
\ex \label{missing-de-as-modal}
\begin{xlist}
\ex
\glll \zh{反而} \zh{那些} \zh{不} \zh{帅} \zh{，} \zh{还} \zh{有} \zh{点} \zh{丑} \zh{但是} \zh{很} \zh{会} \zh{唱歌} \zh{就} \zh{被} \zh{淘汰} \zh{了} \zh{。}\\
\textit{fǎn-ér} \textit{nà-xiē} \textit{bù} \textit{shuài} \textit{,} \textit{hái} \textit{yǒu} \textit{diǎn} \textit{chǒu} \textit{dàn-shì} \textit{hěn} \textit{huì} \textit{chàng-gē} \textit{jiù} \textit{bèi} \textit{táo-tài} \textit{le}.\\
instead those not handsome , also a bit ugly but very can sing then get eliminated PFV .\\
\trans `Instead, those who are not handsome, a bit ugly, but can sing well were eliminated.'
\ex
\glll \zh{反而} \zh{那些} \zh{不} \zh{帅} \zh{，} \zh{还} \zh{有} \zh{点} \zh{丑} \zh{但是} \zh{很} \zh{会} \zh{唱歌} \zh{\textcolor{blue}{的}} \zh{就} \zh{被} \zh{淘汰} \zh{了} \zh{。}\\
\textit{fǎn-ér} \textit{nà-xiē} \textit{bù} \textit{shuài} \textit{,} \textit{hái} \textit{yǒu} \textit{diǎn} \textit{chǒu} \textit{dàn-shì} \textit{hěn} \textit{huì} \textit{chàng-gē} \textcolor{blue}{\textit{de}} \textit{jiù} \textit{bèi} \textit{táo-tài} \textit{le}.\\
instead those not handsome , also a bit ugly but very can sing \textcolor{blue}{REL} then get eliminated PFV .\\
\trans `Instead, those who are not handsome, a bit ugly, but can sing well were eliminated.'
\end{xlist}
\end{exe}

\paragraph{Character order}

Another interesting type of spelling mistake in Chinese writing involves producing the correct characters for a word but placing them in the incorrect order. This is an easy mistake to make from carelessness, but it can be especially challenging for learners because some Chinese words include characters that share radicals and look alike, such as \zh{魂魄} \textit{hún-pò} (`spirit'), \zh{忐忑} \textit{tǎn-tè} (`perturbed'), and \zh{森林} \textit{sēn-lín} (`forest').  Rearranging the characters often results in something nonsensical, but sometimes it also produces a valid word with a different meaning, like \zh{牙刷} \textit{yá-shuā} (noun, `toothbrush') vs \zh{刷牙} \textit{shuā-yá}  (verb, `to brush teeth'), or \zh{著名} \textit{zhù-míng}  (adjective, `famous') vs \zh{名著} \textit{míng-zhù}  (noun, `masterpiece'). 
We create a separate error category for incorrect character order because the precise order of characters is critical to maintaining grammatical integrity, and even a slight change can significantly alter the sentence's readability and meaning.
In \eqref{character-order}, the writer reverses the characters in the common interrogative word \zh{什么} \textit{shén-me} (`what'), and the result \zh{么什} \textit{me-shén} is a nonsensical string.  

\begin{exe}
\ex \label{character-order}
\begin{xlist}
\ex
\glll \zh{要} \zh{了解} \zh{一} \zh{个} \zh{人} \zh{，} \zh{不妨} \zh{看} \zh{他} \zh{读} \zh{些} \zh{\textcolor{red}{么 什}} \zh{书} \zh{，} \zh{观察} \zh{向} \zh{他} \zh{来往} \zh{得} \zh{朋友} \zh{一样} \zh{有效} \zh{。}\\
\textit{yào} \textit{liǎo-jiě} \textit{yī} \textit{gè} \textit{rén} \textit{,} \textit{bù-fáng} \textit{kàn} \textit{tā} \textit{dú} \textit{xiē} \textcolor{red}{\textit{me shén}} \textit{shū} \textit{,} \textit{guān-chá} \textit{xiàng} \textit{tā} \textit{lái-wǎng} \textit{dé} \textit{péng-yǒu} \textit{yī-yàng} \textit{yǒu-xiào}.\\
want understand one CL person , might-as-well see he read some \textcolor{red}{what} book , observe toward him contact REL friend same effective .\\
\trans `To understand a person, it's just as effective to see what books he reads and observe the friends he interacts with.'

\ex
\glll \zh{要} \zh{了解} \zh{一} \zh{个} \zh{人} \zh{，} \zh{不妨} \zh{看} \zh{他} \zh{读} \zh{些} \zh{\textcolor{blue}{什么}} \zh{书} \zh{，} \zh{这} \zh{跟} \zh{观察} \zh{与} \zh{他} \zh{来往} \zh{的} \zh{朋友} \zh{一样} \zh{有效} \zh{。}\\
\textit{yào} \textit{liǎo-jiě} \textit{yī} \textit{gè} \textit{rén} \textit{,} \textit{bù-fáng} \textit{kàn} \textit{tā} \textit{dú} \textit{xiē} \textcolor{blue}{\textit{shén-me}} \textit{shū} \textit{,} \textit{zhè} \textit{gēn} \textit{guān-chá} \textit{yǔ} \textit{tā} \textit{lái-wǎng} \textit{de} \textit{péng-yǒu} \textit{yī-yàng} \textit{yǒu-xiào}.\\
want understand one CL person , might-as-well see he read some \textcolor{blue}{what} book , this with observe with him contact REL friend same effective .\\
\trans `To understand a person, it's just as effective to see what books he reads and observe the friends he interacts with.'
\end{xlist}
\end{exe}

\subsection{Implementation overview} \label{subsec:implementation}

\citet{gu-etal-2025-improving} implemented the refined annotation scheme in a revised Chinese \texttt{errant}-style framework. Relative to \texttt{ChERRANT}, the implementation makes two principal changes. First, it adopts annotation labels that align more closely with the English \texttt{errant} convention, using \texttt{R} (replacement), \texttt{M} (missing), and \texttt{U} (unnecessary) rather than \texttt{S} (substitution), \texttt{M} (missing), and \texttt{R} (redundant). Second, it introduces more fine-grained labels for Chinese-specific error types, including pronunciation-based confusions, shape-based confusions, mixed similarity, character-order errors, word-order errors, and \textit{de}-related errors.

The implementation is designed to preserve compatibility with \texttt{m2}-style edit annotation while making Chinese learner errors more linguistically explicit. In particular, cases that would previously receive a generic label such as \texttt{S:SPELL} in \texttt{ChERRANT} can instead be differentiated into categories such as \texttt{R:PINYIN}, \texttt{R:SHAPE}, or \texttt{R:MULTI}, depending on the nature of the confusion. Similarly, omitted particles such as \zh{的}, \zh{地}, and \zh{得} can be represented more directly through labels such as \texttt{M:DE}.

Figure~\ref{m2-differences} illustrates the difference between \texttt{ChERRANT} output and the refined implementation. The most important contrast lies not in the overall \texttt{m2} structure, which remains comparable across the two systems, but in the granularity and linguistic interpretability of the resulting labels. At the same time, the comparison also highlights a persistent issue in Chinese annotation: because edit spans are defined relative to a particular tokenization, differences in segmentation can lead to different boundaries and indices even when the underlying correction is the same.\label{r1-p24}

From a survey perspective, the importance of this implementation lies less in the engineering details than in its representational consequences. It shows how Chinese-specific orthographic and grammatical phenomena can be incorporated into automatic \texttt{m2}-style annotation without abandoning compatibility with existing GEC evaluation workflows.

\begin{figure}[ht!]
\begin{subfigure}[b]{\textwidth}    
\centering
\footnotesize{
\begin{tabular}{l}
\texttt{S \zh{我 {\color{red}一前} 没 住 过 五星级 旅馆 ， 所以 我 很 惊奇 了 。}}\\
\texttt{T0-A0 \zh{我 以前 没 住 过 五星级 旅馆 ， 所以 我 很 惊奇 。}}\\
\texttt{A 1 2|||{\color{red}S:SPELL}|||{\color{red}\zh{以前}}|||REQUIRED|||-NONE-|||0}\\
\texttt{A 12 13|||R:AUX|||-NONE-|||REQUIRED|||-NONE-|||0}\\
(`I have never stayed at a five-star hotel before, so I was very surprised.')\\
\texttt{}\\
\texttt{S \zh{她 有 两 个 姐姐 、 一个 妹妹 和 {\color{red}西} 个 哥哥 。 }}\\
\texttt{T0-A0 \zh{她 有 两 个 姐姐 、 一个 妹妹 和 四 个 哥哥 。}}\\ 
\texttt{A 9 10|||{\color{red}S:SPELL}|||{\color{red}\zh{四}}|||REQUIRED|||-NONE-|||0}\\
(`She has two older sisters, one younger sister, and four older brothers.')\\
\texttt{}\\
\texttt{S \zh{从 十六 世纪 开始 ， {\color{red}欧州人} 就 抽烟 。 }}\\
\texttt{T0-A0 \zh{从 十六 世纪 开始 ， 欧洲人 就 抽烟 。 }}\\
\texttt{A 24 25|||{\color{red}S:SPELL}|||{\color{red}\zh{欧洲人}}|||REQUIRED|||-NONE-|||0}\\
(`Since the 16th century, Europeans have been smoking.')\\
\texttt{}\\
\texttt{S \zh{反而 那些 不 帅 ， 还 有 点 丑 但是 很 会 唱歌 就 被 淘汰 了 。}}\\
\texttt{T0-A0 \zh{反而 那些 不 帅 ， 还 有 点 丑 但是 很 会 唱歌 的 就 被 淘汰 了 。}}\\
\texttt{A 13 13|||{\color{red}M:AUX}|||\zh{的}|||REQUIRED|||-NONE-|||0}\\
(`Instead, those who were not handsome, a bit ugly, but very good at singing were eliminated.')
\end{tabular}
}
\caption{m2 file generated by \texttt{ChERRANT}}
\label{chinese-word-based-m2}
\end{subfigure}

\hfill

\begin{subfigure}[b]{\textwidth}    
\centering
\footnotesize{
\begin{tabular}{l}
\texttt{S \zh{我 {\color{blue}一 前} 没住 过 五 星 级 旅 馆 ， 所以 我 很 惊讶 了 。}}\\
\texttt{A 1 3|||{\color{blue}R:PINYIN}|||{\color{blue}\zh{以前}}|||REQUIRED|||-NONE-|||0}\\
\texttt{A 15 16|||U:PART||||||REQUIRED|||-NONE-|||0}\\
\\
\texttt{S \zh{她 有 两 个 姐姐 、 一 个 妹妹 和 {\color{blue}西} 个 哥哥 。}}\\
\texttt{A 10 11|||{\color{blue}R:SHAPE}|||{\color{blue}\zh{四}}|||REQUIRED|||-NONE-|||0}\\
\\
\texttt{S \zh{从 十六 世纪 开始 ， {\color{blue}欧州} 人 就 抽烟 。}}\\
\texttt{A 5 6|||{\color{blue}R:MULTI}|||{\color{blue}\zh{欧洲}}|||REQUIRED|||-NONE-|||0}\\
\\
\texttt{S \zh{反而 那些 不帅 ， 还 有 点 丑 但是 很会 唱歌 就 被 淘汰 了 。}}\\
\texttt{A 11 11|||{\color{blue}M:DE}|||\zh{的}|||REQUIRED|||-NONE-|||0}\\
\end{tabular}
}
\caption{{m2 file generated by \citet{gu-etal-2025-improving}}}
\label{korean-m2}
\end{subfigure}

\caption{Differences in m2 files}
\label{m2-differences}

\end{figure}

\subsection{Discussion and implications of automatic annotation} \label{subsec:discussion-annotation}

Automatic annotation in Chinese grammatical error correction offers an efficient way to derive edit-based representations from parallel source and corrected sentences, but its effectiveness depends heavily on representational choices. The comparison between earlier frameworks such as \texttt{ChERRANT} and later refined schemes shows that automatic annotation is not only a matter of detecting edits, but also of deciding what counts as the relevant unit of error, how errors should be segmented, and which distinctions should be encoded in the label inventory.

One central issue is the choice between character-based and word-based annotation. Character-based annotation is attractive because it avoids committing to a particular word segmentation and maps naturally onto the graphological structure of Chinese writing. It can therefore represent local orthographic changes with high precision, especially when the error involves a single character. At the same time, character-based annotation may obscure grammatical structure, since many Chinese errors are more naturally described at the level of words or multi-character expressions rather than isolated characters. Word-based annotation, by contrast, allows labels to be linked more directly to part-of-speech categories and syntactic functions, which makes it more informative for linguistic analysis and for alignment with edit-based evaluation frameworks. Its limitation is that it inherits all the uncertainty of Chinese word segmentation, including disagreements over word boundaries and differences across tokenizers.

A related issue concerns segmentation granularity. Even within word-based annotation, different tokenizers may produce different segmentations for the same sentence, which in turn affects edit boundaries, index positions, and sometimes even the apparent type of error. A form treated as a single lexical unit under one segmentation may appear as a sequence of smaller units under another, leading to different annotations for what is intuitively the same correction. This is particularly important in Chinese, where many learner errors involve boundary ambiguities between characters, words, and partially lexicalized expressions. As a result, automatic annotation in CGEC is sensitive not only to the correctness of the final output, but also to the tokenization regime under which edits are defined.

Another important distinction concerns the granularity of the label inventory itself. Earlier frameworks such as \texttt{ChERRANT} provide a practical and compact annotation scheme built around a small number of edit operations, supplemented by limited extensions such as spelling and word-order labels. This makes automatic annotation relatively efficient and easy to use in evaluation pipelines. However, such compact schemes may collapse together error types that are linguistically distinct, for instance pronunciation-based confusions, visually motivated character confusions, mixed similarity effects, and the omission or misuse of \textit{de}-particles. Refined annotation schemes seek to make these distinctions explicit, thereby producing annotations that are more informative for linguistic analysis and learner-oriented feedback. The tradeoff is that finer-grained labels require stronger assumptions about error classification and may increase dependence on language-specific heuristics.

Automatic annotation in Chinese also raises the question of phrase-oriented versus word-oriented analysis. Some errors, especially local substitutions and insertions, are naturally represented as word-level edits. Others, however, involve broader reordering relations or corrections whose most natural description spans a larger phrase. In such cases, automatic systems may either compress the phenomenon into a dedicated order label or decompose it into a sequence of smaller edit operations. The choice is consequential, because it affects not only interpretability but also downstream evaluation. A phrase-level analysis may better reflect the linguistic nature of the error, whereas a word-level decomposition may be easier to integrate into existing \texttt{m2}-style scoring workflows.

These observations suggest that automatic annotation in CGEC should not be evaluated solely in terms of whether it produces a formally valid \texttt{m2} file. What matters equally is how well the resulting representation captures the linguistic structure of learner errors and how stable that representation remains across different tokenization and annotation choices. In this respect, earlier frameworks such as \texttt{ChERRANT} provide an important baseline for practical automatic annotation, while later refinements show that Chinese-specific phenomena require a more differentiated treatment than generic spelling or edit labels can provide.

More broadly, the comparison between existing and refined schemes highlights a methodological tension that remains unresolved. Automatic annotation benefits from compact edit inventories, stable evaluation formats, and reproducible pipelines, but Chinese learner errors often demand richer distinctions tied to orthography, morphology, lexicalization, and function words. A robust annotation framework for CGEC must therefore balance efficiency, cross-system comparability, and linguistic adequacy. Future work will need to address this balance more directly, especially in relation to segmentation-sensitive annotation, the treatment of broad reordering phenomena, and the interaction between annotation design and multi-reference evaluation.

\section{General evaluation in GEC} \label{sec:evaluation-gec}

Evaluation in grammatical error correction is not merely a matter of checking whether a system output differs from the source. A correction may improve grammaticality, preserve meaning, make only minimal formal changes, or instead introduce unnecessary rewriting. For this reason, evaluation in GEC has developed into a methodological area of its own, with metrics that differ not only in implementation but also in what they regard as evidence of a successful correction.

A common distinction is drawn between \textit{edit-based}, \textit{fluency-based}, and \textit{reference-less} evaluation. Edit-based metrics compare system output to reference corrections through explicit edit operations and typically reward systems that make accurate and minimal changes to the source. Fluency-based metrics, by contrast, evaluate the corrected sentence more holistically as an improved output, allowing broader rewrites as long as they produce a better formed sentence. Reference-less evaluation weakens or removes dependence on gold-standard corrections altogether and instead estimates output quality directly from grammaticality or post-editing behavior.

These approaches reflect different assumptions about what should count as a good correction. Edit-based evaluation offers transparency and diagnostic precision, but it is often sensitive to alignment and annotation choices. Fluency-based evaluation is more tolerant of alternative valid rewritings, but provides less explicit information about which errors were corrected. Reference-less evaluation extends this flexibility further, but raises additional questions about adequacy and meaning preservation. In this section, we review the main evaluation paradigms used in GEC and discuss their respective strengths and limitations.

\subsection{Edit-based evaluation metrics}\index{evaluation!edit-based}

\subsubsection{Helping Our Own}\index{Helping Our Own}

The Helping Our Own shared task evaluates grammatical error correction using the $F_1$ score computed over grammatical error annotation edits \citep{dale-kilgarriff-2011-helping}. It provides XML-style annotation guidelines, and each participating system is required to submit a set of edits with all relevant attributes specified. Each \texttt{<edit>} element includes a unique index, an error type, and character-level \texttt{start} and \texttt{end} offsets marking the span of the edit in the original source text. It also contains an \texttt{<original>} string and a \texttt{<correction>} list containing one or more proposed revisions. Because HOO uses character-based indexing rather than token-based spans, it supports relatively fine-grained evaluation, especially for punctuation, spacing, and formatting errors.

The evaluation aligns system edits with gold-standard edits using a strict matching criterion: both the start and end character offsets must coincide exactly. Once an edit is aligned, the system correction must match one of the reference correction strings exactly in order to count as correct. Precision, recall, and $F_1$ are then computed over the matched edits. The framework also includes a bonus mode for optional edits, so that systems are not penalized for omitting corrections marked as non-required.

Despite its historical importance, HOO presents several limitations. Because alignment is based on character spans, linguistically appropriate corrections may fail to receive credit when they differ from the reference in segmentation, for instance when a system splits or merges edits differently. In addition, the framework leaves some ambiguity regarding edit granularity, since it does not clearly constrain whether edits should remain local or may extend across broader spans. These limitations are compounded by the absence of widely used technical guidelines and standardized conversion tools. As a result, later work has generally favored evaluation frameworks with more explicit operational definitions and more reproducible implementations.

\subsubsection{\texorpdfstring{\texttt{M\textsuperscript{2}}}{M²}}

MaxMatch (\texttt{M\textsuperscript{2}}) \citep{dahlmeier-ng-2012-better} became one of the most influential edit-based evaluation metrics in GEC. It evaluates a system by comparing the edits implied by the system output with the gold edits provided in reference annotation. Rather than scoring final sentences directly, \texttt{M\textsuperscript{2}} identifies the set of edits that best transforms the source sentence into the system hypothesis and measures how well this set matches the annotated reference edits. It reports precision, recall, and a weighted $F$-score, with $F_{0.5}$ used by default to place greater emphasis on precision.\footnote{It is useful to distinguish between \texttt{M\textsuperscript{2}}, the MaxMatch evaluation metric, and M2, the file format used for grammatical error annotation.}

A key feature of \texttt{M\textsuperscript{2}} is that it formulates evaluation as a search over possible edit sequences. Given the source sentence and the system output, the scorer derives phrase-level edits and selects the edit sequence that yields the highest overlap with the gold annotation. This design made \texttt{M\textsuperscript{2}} especially attractive for shared tasks, since it offered a principled and reproducible way to evaluate correction quality while favoring conservative systems that avoid unnecessary changes.

The strength of \texttt{M\textsuperscript{2}} lies in its interpretability. Because it evaluates explicit edit overlap, it provides a direct account of which corrections were matched and which were missed. At the same time, this strength also defines its limitation. Exact matching is required not only in corrected content but also in span structure, so systems may be penalized even when they produce a valid final sentence if their edits are segmented differently from the reference. In this respect, \texttt{M\textsuperscript{2}} remains sensitive to superficial mismatches in edit representation and inherits the broader limitations of reference-dependent edit-based evaluation.

When multiple references are available, \texttt{M\textsuperscript{2}} scores the system output against each reference separately and retains the highest-scoring result. This best-match strategy avoids penalizing systems for producing a correction that aligns with one valid reference but not another. However, it also means that the full structure of the multi-reference set is not taken into account during evaluation.

\subsubsection{I-measure}\index{I-measure}

I-measure (improvement measure) evaluates grammatical error correction at the token level by comparing the source sentence, the system output, and a reference correction through a unified alignment procedure \citep{felice-briscoe-2015-towards}. Unlike \texttt{M\textsuperscript{2}}, which scores overlap between phrase-level edits, I-measure is based on a globally optimal three-way token alignment computed with dynamic programming. This design allows the metric to assess correction behavior directly at the token level rather than through extracted edit spans.

On the basis of this alignment, tokens are categorized for both detection and correction evaluation using standard measures such as precision, recall, and $F_\beta$. In addition, I-measure introduces a special category, FPN (false positive and false negative), for cases in which a token simultaneously reflects an unnecessary change and a missed correction. This category is included in correction-oriented scoring but excluded from detection scoring, allowing the framework to distinguish more carefully between identifying an error and repairing it appropriately.

A central feature of I-measure is its use of weighted accuracy ($\text{WAcc}$), which assigns greater importance to successful corrections than to simply leaving correct tokens unchanged. This reflects the intuition that GEC systems should be rewarded not only for preserving correct material but also for making beneficial edits where needed. Building on weighted accuracy, I-measure defines an improvement score ($I$) that compares system performance against a no-change baseline, that is, a baseline that outputs the source sentence unchanged. A positive $I$ score indicates that the system improves on this baseline, whereas a negative $I$ score indicates that the system degrades the input overall.

The improvement score is defined in terms of the system weighted accuracy ($\text{WAcc}_{\text{sys}}$) and the baseline weighted accuracy ($\text{WAcc}_{\text{base}}$) as follows:

\begin{equation}
I =
\begin{cases}
0 & \text{if } \text{WAcc}_{\text{sys}} = \text{WAcc}_{\text{base}} \\
\displaystyle \frac{\text{WAcc}_{\text{sys}} - \text{WAcc}_{\text{base}}}{1 - \text{WAcc}_{\text{base}}} & \text{if } \text{WAcc}_{\text{sys}} > \text{WAcc}_{\text{base}} \\
\displaystyle \frac{\text{WAcc}_{\text{sys}}}{\text{WAcc}_{\text{base}}} - 1 & \text{if } \text{WAcc}_{\text{sys}} < \text{WAcc}_{\text{base}}
\end{cases}
\end{equation}

Weighted accuracy is computed as follows:

\begin{equation}
\text{WAcc} =
\frac{w \cdot TP + TN}
{w \cdot (TP + FP) + TN + FN - \frac{(w + 1) \cdot \text{FPN}}{2}}
\end{equation}

\noindent where $w$ controls the relative importance assigned to successful corrections as opposed to the preservation of already correct tokens.

When multiple references are available, I-measure evaluates the system output against each reference independently and retains the highest-scoring result for each sentence. This best-match strategy acknowledges that multiple corrections may be linguistically valid and reduces the risk of penalizing systems for producing acceptable outputs that differ from a single fixed reference.

Compared with span-based edit metrics, I-measure offers a more explicitly token-centered view of correction quality and provides a principled way to quantify whether a system truly improves the source sentence. At the same time, like other reference-based metrics, it remains dependent on alignment decisions and on the availability of suitable human corrections.

\subsubsection{\texttt{errant\_compare}}\index{errant\_compare}

\texttt{errant\_compare} \citep{bryant-etal-2017-automatic} is the main evaluation component of the \texttt{errant} framework and has become one of the most widely used tools for grammatical error correction evaluation. It compares a system-produced M2 annotation file against a gold-standard M2 reference and computes edit-based precision, recall, and $F$-score, with $F_{0.5}$ commonly used as the default in order to place greater weight on precision. In addition to aggregate scores, it reports the underlying counts of true positives, false positives, and false negatives, which makes the evaluation readily interpretable and useful for detailed system analysis.

A central feature of \texttt{errant} is that it does not treat edits as undifferentiated span changes. Instead, edits are assigned structured error labels that combine a core edit operation with a more specific linguistic category. At the most general level, \texttt{errant} distinguishes three edit operations: M (Missing), for material that should be inserted; R (Replacement), for material that should be substituted; and U (Unnecessary), for material that should be deleted. These core operations can then be refined into more specific categories such as \texttt{M:PUNCT} or \texttt{R:VERB:TENSE}, enabling evaluation not only of overall correction quality but also of performance on particular classes of errors.

This classification scheme makes \texttt{errant\_compare} especially useful for diagnostic evaluation. Because edits are typed rather than merely matched as spans, the tool can support fine-grained analyses of where a system performs well and where it fails, for example across punctuation, morphology, verbal inflection, or lexical choice. In practice, this richer analysis depends on language-specific annotation resources and preprocessing pipelines. For English, the original framework provides a standard configuration, whereas adaptation to other languages requires corresponding decisions about tokenization, part-of-speech mapping, and error taxonomy design.

The M2 files used by \texttt{errant\_compare} are typically produced with \texttt{errant\_parallel}, which takes a source sentence and a corrected version as input and derives the corresponding edit sequence automatically. The evaluation performed by \texttt{errant\_compare} is therefore grounded in explicit alignment between source, hypothesis, and reference corrections, rather than only in surface similarity between final sentences.

When multiple human references are available for a given sentence, \texttt{errant\_compare} evaluates the system output against each reference separately and retains the reference that yields the highest score for that sentence. This best-match strategy acknowledges that more than one correction may be acceptable and reduces the risk of penalizing systems for producing a valid correction that differs from a single annotated reference. However, like \texttt{M\textsuperscript{2}}, the implementation places a practical limit on the number of annotators considered in this search.

Because it combines explicit edit alignment, interpretable scoring, and fine-grained error typing, \texttt{errant\_compare} has become a \textit{de facto} standard\index{de facto standard} in GEC evaluation. Its main strength lies in the transparency of its scoring procedure and its usefulness for diagnostic analysis, although, like other edit-based metrics, it remains sensitive to how edits are segmented and annotated.

\subsection{Fluency-based and reference-less metrics}\index{evaluation!fluency-based}\index{evaluation!reference-less}

\subsubsection{GLEU}\index{GLEU}

GLEU \citep{napoles-etal-2015-ground} is a fluency-based evaluation metric designed for grammatical error correction and inspired by BLEU \citep{papineni-etal-2002-bleu}. Unlike BLEU, which measures overlap between system output and reference translations alone, GLEU also incorporates the source sentence into the evaluation. This allows it to reward corrections that bring the hypothesis closer to the reference while penalizing cases in which erroneous material from the source is unnecessarily preserved.

The metric was introduced in response to a limitation of edit-based evaluation: a system may produce a fluent and acceptable correction without matching the reference in terms of explicit edit boundaries. Rather than comparing annotated edits directly, GLEU evaluates the corrected sentence as an output string and measures how well its $n$-grams align with those of the reference, while taking into account the relationship between the source and the hypothesis. In this respect, it is better suited than BLEU to the monolingual rewriting setting of GEC, where systems are expected to improve grammaticality and fluency without making gratuitous changes to already correct material.

GLEU rewards $n$-grams that appear in the reference and in the system output, especially when they correspond to improvements over the source, and penalizes $n$-grams that are retained from the source even though they should have been corrected. This asymmetric treatment of source and reference reflects the central objective of GEC: to make necessary revisions while preserving correct content. As a result, GLEU is generally more sensitive than BLEU to the difference between valid correction and mere copying of the original sentence.

A later variant, GLEU$^{+}$ \citep{napoles-etal-2016-gleu} simplifies the original formulation and removes the need for tuning with respect to the number of references. It preserves the main intuition of the original metric while improving robustness across evaluation settings with different numbers of human corrections. In GLEU$^{+}$, the modified $n$-gram precision is defined as follows:

\begin{equation}
p_n =
\frac{\displaystyle \sum_{i=1}^{|D|} \max\left(0, \sum_{g_n \in h_i} \min(\rho_{i,g_n}, \eta_{i,g_n}) - \min(\sigma_{i,g_n}^{\text{diff}}, \eta_{i,g_n})\right)}
{\displaystyle \sum_{i=1}^{|D|} \sum_{g_n \in h_i} \eta_{i,g_n}}
\end{equation}

\noindent where evaluation is carried out over sentence triples $(s_i, h_i, r_i)$ consisting of a source sentence, a system hypothesis, and a reference correction. The formulation rewards overlap between hypothesis and reference while subtracting unsupported $n$-grams carried over from the source. In this way, GLEU$^{+}$ captures both fluency improvement and failure to correct.

Compared with edit-based metrics, GLEU and GLEU$^{+}$ are less dependent on explicit edit segmentation and are therefore more tolerant of alternative but acceptable rewritings. Their main strength lies in evaluating the corrected sentence as a fluent output rather than as a sequence of aligned edits. At the same time, they provide less diagnostic detail than edit-based metrics, since they do not identify which specific errors were corrected or missed.

\subsubsection{Reference-less}\index{evaluation!reference-less}

Reference-less evaluation in grammatical error correction aims to assess system output without comparing it directly against gold-standard reference corrections. This line of work arises from a well-known limitation of reference-based metrics: a system may produce a valid and fluent correction yet receive little credit if that correction is not represented in the available references. To address this problem, \citet{napoles-etal-2016-theres} proposed reference-less metrics that estimate grammatical quality directly from the corrected output.

Their approach focuses on grammaticality-based evaluation. Three metrics are introduced: ER, which uses e-rater\textsuperscript{\textregistered}'s grammatical error detection modules; LT, which uses LanguageTool; and LFM, a regression-based model built from linguistic features such as misspellings, language model scores, out-of-vocabulary rates, and syntactic well-formedness indicators. ER and LT estimate sentence quality through normalized error counts, while LFM predicts a continuous grammaticality score from learned features. These methods shift the focus away from edit overlap and toward the intrinsic well-formedness of the system output.

A natural limitation of grammaticality-based evaluation is that it does not directly measure adequacy, namely whether the corrected sentence preserves the meaning of the source. To address this, \citet{napoles-etal-2016-theres} also propose interpolated metrics that combine a grammaticality score ($S_G$) with a reference-based score ($S_R$), such as GLEU or \texttt{M\textsuperscript{2}}:

\begin{equation}
S_I = (1 - \lambda) S_G + \lambda S_R
\end{equation}

\noindent where $\lambda \in [0, 1]$ controls the balance between grammaticality and reference-based adequacy. This interpolation reflects the broader point that evaluation in GEC often requires more than one criterion, since grammatical well-formedness alone is not sufficient if the output distorts the intended meaning.

More recent work extends reference-less evaluation by using large language models as post-editors rather than fixed grammar-checking tools. In this framework, a large language model rewrites the system output into a more fluent and grammatical form, and the similarity between the original output and the post-edited version is used as the evaluation signal \citep{ostling-etal-2025-llm}. This approach avoids dependence on fixed human references while still grounding evaluation in an explicit notion of post-editing quality. It has also been shown to correlate well with human judgments across different evaluation setups.

Compared with reference-based metrics, reference-less methods are more tolerant of valid alternative corrections and are especially attractive in multilingual or low-resource settings where high-quality references may be sparse or unavailable. Their weakness is that they may capture grammaticality more readily than faithfulness to the source, unless adequacy is incorporated explicitly through interpolation or post-editing-based designs. Even so, they represent an important expansion of the GEC evaluation landscape by moving beyond exact agreement with a fixed set of annotated corrections.

\subsection{Discussion}

Evaluation in grammatical error correction is shaped by a persistent tension between interpretability and flexibility. Edit-based metrics reward systems for producing corrections that can be aligned closely with annotated reference edits, whereas fluency-based metrics assess the corrected sentence more holistically as an improved output. Reference-less approaches extend this shift further by attempting to evaluate grammaticality and acceptability without relying directly on gold-standard corrections. These three strands of work reflect different assumptions about what counts as a successful correction and about how much variation an evaluation metric should tolerate.

Edit-based metrics such as \textsc{HOO}, \texttt{M\textsuperscript{2}}, I-measure, and \texttt{errant\_compare} remain attractive because they provide transparent and diagnostically useful results. They make it possible to determine not only whether a system improved a sentence, but also which edits were proposed, which were correct, and which types of errors were handled well or poorly. This degree of interpretability is particularly valuable in learner-oriented applications and in comparative system analysis. Their main limitation, however, lies in their dependence on alignment decisions and on the particular segmentation of edits in the reference. As a result, valid corrections may receive reduced credit when they differ structurally from the annotated correction, even if the final output is grammatical and meaning preserving.

Fluency-based metrics such as GLEU and GLEU$^{+}$ address part of this limitation by evaluating the corrected sentence as a rewritten output rather than as a sequence of explicit edits. Because they compare the hypothesis against the reference while also taking the source into account, they are better able to reward successful revision and penalize unnecessary preservation of erroneous source material. They are therefore more tolerant of alternative phrasings and broader rewrites than strictly edit-based metrics. This greater flexibility, however, comes at the cost of reduced diagnostic granularity, since such metrics do not directly identify which error types were corrected or missed.

Reference-less evaluation extends this flexibility further by weakening or removing dependence on human references. Grammaticality-based metrics and approaches based on large language model post-editing are particularly appealing when high-quality references are sparse, costly, or incomplete. These methods broaden the scope of GEC evaluation, especially in multilingual and low-resource settings, but they also highlight a persistent challenge: grammatical well-formedness alone is not sufficient if semantic adequacy is not preserved. For this reason, reference-less evaluation is often most informative when used alongside reference-based metrics rather than as a full replacement for them.

No single metric is sufficient for all GEC evaluation scenarios. Edit-based metrics remain essential for interpretability and error analysis, while fluency-based and reference-less metrics are better suited to more flexible rewriting. In practice, evaluation protocols often benefit from combining several metrics, depending on whether the goal is benchmark comparison, learner feedback, or the assessment of more generative correction systems.

\section{Chinese Grammatical Error Detection} \label{sec:cged-appendix}

Chinese Grammatical Error Detection (CGED) is the task of identifying 
the presence, location, and type of grammatical errors in Chinese text, 
without generating corrections. Although detection and correction are 
closely related, they impose different modeling requirements: detection 
can be cast as sequence labeling or classification, whereas correction 
additionally requires generation or rewriting. As noted in 
Section~\ref{sec:methods}, early Chinese GEC research was predominantly 
detection-oriented; the present appendix reviews that body of work. CGED 
has been studied primarily through shared task campaigns organized 
alongside the CGED datasets discussed in Section~\ref{sec:datasets}. 
Table~\ref{summary-cged-systems} provides a summary of the systems 
reviewed in this section.

\paragraph{Rule-based and statistical approaches}

Early approaches to CGED relied on manually crafted rules or feature-engineered classifiers. \citet{jiang-etal-2012-rule-based} proposed a rule-based model for grammar error detection using XML-based rules derived from examination data and grammar reference books, covering 803 manually crafted rules targeting quantifier misuse, particle misuse, and structural mismatches. Building on this framework, 
\citet{wu-etal-2015-research} extended rule-based methods within the LanguageTool framework to address semantic-level errors such as incorrect geopolitical references and inappropriate adjective-noun collocations, representing an early attempt to move from detection toward limited correction for specific error categories.

Statistical classifiers broadened coverage beyond manually defined rules by learning error patterns from data. \citet{yu-chen-2012-detecting} adopted a hybrid approach combining word n-gram statistics, POS tagging from large-scale web corpora, and SVM classifiers for word-ordering error detection. By combining explicit linguistic knowledge with data-driven classification, this approach improved detection of misplaced adverbs, verbs, and objects, which are common error types for Chinese learners due to syntactic divergences from Indo-European languages. \citet{lee-etal-2014-sentence} integrated 142 linguistically defined rules alongside n-gram probability analysis for detection of POS misuse, word-order errors, and idiomatic expression violations. Such hybrid systems illustrate the benefit of augmenting explicit linguistic knowledge with statistical probability models: the rules provide targeted precision while the n-gram models improve generalization.

These systems demonstrate a recurrent trade-off: high precision on targeted error types but limited recall on errors not anticipated by the rule inventory or feature set. In Chinese, this limitation is compounded by the need to handle language-specific phenomena such as particle misuse (\zh{的}/\zh{地}/\zh{得}), measure word agreement, and context-dependent function words, which require large and detailed rule inventories. Because Chinese text lacks explicit word delimiters, these systems also depend on a word segmentation step before grammatical analysis can be applied, making segmentation quality a precondition for downstream rule matching. Developing, updating, and maintaining such rules requires considerable manual effort, and the resulting systems are inherently difficult to scale to the full range of grammatical phenomena in Chinese.

\paragraph{CRF-based approaches}

CRF-based detection systems addressed some of these scalability limitations by learning sequential patterns from annotated data rather than relying on manually specified rules. \citet{cheng-etal-2014-chinese} applied CRFs trained on the Google Chinese Web 5-gram corpus for word-ordering error detection, integrating pointwise mutual information (PMI) and language model probabilities. Their system combined CRF-based detection with a Ranking SVM-based correction module that generated and ranked candidate reorderings, achieving an $F_1$ of 0.800 for detection and a recall rate of 0.858 for correction on the HSK dataset. \citet{chen-etal-2015-chinese} further developed CRF-based methods combined with n-gram features to detect multiple error types, including missing words, redundant words, word misordering, and incorrect word 
selection. \citet{liu-etal-2016-automatic} applied CRFs with n-gram features for grammatical error detection on the CGED shared task, achieving strong results at the detection and identification levels. \citet{wang-shih-2018-hybrid} combined CRFs with the Stanford Word Segmenter and post-processing using the Google n-gram corpus for structured error detection.

CRFs captured sequential dependencies more effectively than independent classifiers, making them well suited for localizing error spans in Chinese text. However, their performance remained sensitive to the quality of input features and the accuracy of word segmentation. Although CRF-based systems improved coverage relative to purely rule-based methods, they shared the same fundamental constraint: their effectiveness was bounded by the hand-engineered features used as input, and they did not by themselves generate corrections.

\paragraph{Neural approaches}

Neural architectures improved detection performance by replacing hand-engineered features with learned representations. \citet{huang-wang-2016-bi} applied a BiLSTM network for error detection using the CKIP Chinese Segmentation System for preprocessing, achieving notable results at the detection level on the NLP-TEA shared tasks. \citet{yang-etal-2016-chinese} compared CNN, RNN, and LSTM architectures with SVM classifiers for position-level detection on the TOCFL and HSK datasets, providing an early systematic comparison of neural architectures for CGED. \citet{zheng-etal-2016-chinese} combined CRF and LSTM models in a stacking architecture, where CRF outputs served as additional features for the LSTM, achieving strong results at both detection and identification levels on NLP-TEA~3.

In the transformer era, \citet{cheng-duan-2020-chinese} fine-tuned three pretrained models, BERT-Base \citep{devlin-etal-2019-bert}, RoBERTa \citep{liu-etal-2019-roberta}, and RoBERTa-wwm \citep{cui-etal-2021-pre}, for error detection on the CGED shared task, demonstrating that pretrained language models could improve detection accuracy, particularly with data augmentation strategies.

The shift from feature-engineered to neural models brought clear gains in detection coverage and reduced the need for task-specific feature design. At the same time, neural models introduced their own limitations: they require larger annotated training sets, their predictions are less interpretable than rule-based or CRF-based outputs, and their sensitivity to segmentation, while reduced, has not been eliminated entirely.

\paragraph{Summary}

The detection systems reviewed above follow the same methodological progression seen in correction research: from manually crafted rules and statistical classifiers, through structured prediction with CRFs, to neural and pretrained transformer models. Across this progression, two consistent patterns emerge. First, each paradigm shift improved coverage by reducing dependence on manually specified knowledge, but no approach fully resolved the challenge of Chinese-specific phenomena such as segmentation instability and context-dependent function words. Second, the boundary between detection and correction has become increasingly blurred: several detection models have served as components within larger correction pipelines (Section~\ref{sec:methods}), and some systems, such as \citet{cheng-etal-2014-chinese} and \citet{wu-etal-2015-research}, incorporated limited correction capabilities alongside their detection functions. This convergence suggests that, going forward, detection and correction may be most productively studied as complementary stages within integrated systems rather than as independent tasks.

\begin{landscape}
\scriptsize
\begin{longtable}{@{}p{0.10\linewidth} p{0.28\linewidth} 
p{0.21\linewidth} 
p{0.10\linewidth} p{0.21\linewidth}@{}}

\caption{Summary of Chinese grammatical error detection systems.}\label{summary-cged-systems} \\

\toprule
\textbf{Study} & \textbf{Method} & \textbf{Dataset} & \textbf{Metric} & \textbf{Result} \\
\midrule
\endfirsthead

\multicolumn{5}{@{}l}{\itshape Table~\ref{summary-cged-systems} continued} \\
\toprule
\textbf{Study} & \textbf{Method} & \textbf{Dataset} & \textbf{Metric} & \textbf{Result} \\
\midrule
\endhead

\midrule
\multicolumn{5}{r@{}}{\itshape Continued on next page} \\
\endfoot

\bottomrule
\endlastfoot

\multicolumn{5}{@{}l}{\textit{Rule-based and feature-based statistical approaches}} \\
\midrule

\citet{jiang-etal-2012-rule-based}
& Rule-based Chinese spelling and grammar detection utility using ICTCLAS4J segmentation/POS tagging and an XML-style rule base; rules target spelling, quantifier misuse, particle misuse, mismatch errors, and related local grammatical patterns.
& Development resources include ten years of China National Matriculation Examination data and common errors collected from Chinese Wikipedia; testing uses a large Chinese Wikipedia corpus.
& Correct rate (precision)
& Correct rate for quantifier / particle / mismatch errors = 100\% / 97.05\% / 94.85\%. \\
\midrule

\citet{wu-etal-2015-research}
& Rule-based semantic correction using ICTCLAS4J segmentation and XML rules built in the LanguageTool framework; targets general semantic collocation errors and politically sensitive semantic usage errors.
& Rule development from manually collected semantic-error categories; testing on a news corpus (Corpus 1) and Sina Weibo microblog data (Corpus 2).
& Correct rate (precision)
& Correct rate before / after rule revision: general semantic rules = 59.7\% / 78.3\% on Corpus 1 and 82.3\% / 87.9\% on Corpus 2; political semantic rules = 62.3\% / 78.4\% on Corpus 1 and 86.4\% / 89.1\% on Corpus 2. \\
\midrule

\citet{yu-chen-2012-detecting}
& SVM-based word-order error detection using syntactic features, web-corpus features from Google Chinese Web 5-gram and Chinese POS-tagged web data, and perturbation features.
& HSK-HSK datasets: word-order error sentences from HSK paired with HSK correct sentences; NAT-HSK datasets: same error sentences paired with native-speaker web sentences. Each balanced dataset contains 2,200 sentences, evaluated by five-fold cross-validation.
& Accuracy
& All features: HSK-HSK / NAT-HSK accuracy = 64.81\% / 71.64\%. Syntactic features only: 63.89\% / 69.61\%. \\
\midrule

\citet{lee-etal-2014-sentence}
& Sentence judgment system combining expert-written linguistic rules and n-gram statistical models; rule-based method uses 142 rules, while the statistical method compares correct and incorrect sentence models built from n-gram scores.
& Development: rules based on common erroneous Chinese sentence patterns; statistical model uses 19,080 HSK error sentences and corrected counterparts. Testing: 880 CSL learner error sentences from NCKU Chinese Language Center and 880 corresponding corrections.
& $F_{1}$ / false positive rate (FPR)
& 3-gram model: $F_{1}$ / FPR = 0.689 / 0.595.\\

\midrule
\multicolumn{5}{@{}l}{\textit{CRF-based and statistical sequence-labeling approaches}} \\
\midrule

\citet{cheng-etal-2014-chinese}
& Word-order error detection and correction: CRF-based segment labeling detects sentence segments containing word-order errors; candidate corrections are generated by single-word, bi-word, and tri-word reordering, then ranked with Ranking SVM using POS n-gram features.
& HSK Dynamic Composition Corpus; 1,150 word-order error sentences corrected by two native-speaker annotators; Google Chinese Web 5-gram corpus used for feature extraction.
& Accuracy / $F_{1}$
& Segment-level detection: accuracy / $F_{1}$ = 0.834 / 0.800; sentence-level: accuracy = 0.788. \\
\midrule

\citet{chen-etal-2015-chinese}
& CRF-based CGED shared-task system using CRF++ with CKIP word segmentation/POS tags and feature templates over terms, POS tags, and neighboring context; predicts O/R/M/S/D labels for no error, redundant, missing, selection, and disorder errors.
& Training: NLP-TEA-1 training/test data and NLP-TEA-2 training data. Testing: NLP-TEA-2 CGED official test set.
& FPR / $F_{1}$
& Across submitted runs, lowest FPR = 0.082 (Run 2); best detection / identification / position $F_{1}$ = 0.4079 (Run 3) / 0.2265 (Run 1) / 0.1742 (Run 3). \\
\midrule

\citet{liu-etal-2016-automatic}
& CRF-based CGED system using n-gram feature templates for sequence labeling of four CGED error types: missing, redundant, selection, and disorder errors.
& Training: CGED 2016 HSK data, with 2015 TOCFL data used in some settings. Testing: NLP-TEA 2016 CGED evaluation data.
& FPR / $F_{1}$
& Across submitted runs, lowest FPR = 0.0481 (Run 2); best detection / identification / position $F_{1}$ = 0.5232 / 0.4669 / 0.3627 (Run 3). \\
\midrule

\citet{wang-shih-2018-hybrid}
& CRF sequence-labeling model with Stanford Word Segmenter/POS preprocessing and a Google Chinese Web 5-gram post-processing layer designed mainly for word-selection and word-order errors.
& Training: NLPTEA-2018 data plus NLPTEA-2016 data; testing: NLPTEA-2018 CGED shared task.
& $F_{1}$
& Hybrid model: detection / identification / position $F_{1}$ = 0.4324 / 0.2673 / 0.0875. \\

\midrule
\multicolumn{5}{@{}l}{\textit{Neural and pretrained-model approaches}} \\
\midrule

\citet{huang-wang-2016-bi}
& BiLSTM sequence-labeling model for CGED; uses CKIP word segmentation and POS tagging, trains separate BiLSTM models for missing, redundant, and selection/word-order errors, and separates selection from word-order errors by error-span length.
& Training: NLP-TEA 1 and NLP-TEA 2 data plus NLP-TEA 3 TOCFL/HSK data. Testing: NLP-TEA 3 CGED; reported results focus on the TOCFL track.
& $F_{1}$
& Across submitted runs, best detection / identification / position $F_{1}$ = 0.6224 / 0.3187 / 0.0745. \\
\midrule

\citet{yang-etal-2016-chinese}
& Neural CGED system using single-word embeddings with CNN and LSTM models; multi-class SVM is used for position-level classification.
& NLP-TEA 3 CGED TOCFL and HSK tracks.
& FPR / $F_{1}$
& Across submitted runs, lowest TOCFL FPR = 0.3382, HSK = 0.271 (Run 3); best TOCFL detection / identification / position $F_{1}$ = 0.6130 / 0.2824 / 0.0002 (Run 1), HSK = 0.5793 / 0.3378 / 0.0034 (Run 2). \\
\midrule

\citet{zheng-etal-2016-chinese}
& Sequence-labeling CGED system comparing CRF, LSTM with character unigram/bigram embeddings, stacking using CRF outputs as LSTM features, and LSTM post-processing.
& CGED-HSK data from NLP-TEA 3; training and validation are merged for final evaluation.
& $F_{1}$
& Across submitted runs, best detection / identification / position $F_{1}$ = 0.6628 (Run 3) / 0.5215 (Run 3) / 0.3855 (Run 1). \\
\midrule

\citet{cheng-duan-2020-chinese}
& Sentence-level grammatical-error detection with pretrained BERT-family classifiers; compares BERT-Base, RoBERTa, and RoBERTa-wwm and studies data augmentation with HSK, School, and Lang-8 data.
& Training: CGED 2014--2018 data; validation: CGED 2020 training set; testing: CGED 2020 test set. Augmentation experiments add HSK, School, or Lang-8 data.
& FPR / $F_{1}$
& RoBERTa-wwm on CGED test: FPR / $F_{1}$ = 0.062 / 0.810. Augmented test $F_{1}$: CGED+HSK / CGED+School / CGED+Lang-8 = 0.766 / 0.879 / 0.413. \\

\midrule
\multicolumn{5}{@{}l}{\textit{Hybrid detection--correction systems evaluated in CGED settings}} \\
\midrule

\citet{li-etal-2018-hybrid}
& Hybrid NLPTEA-2018 CGED system with two stages: BiLSTM-CRF detection with handcrafted features, followed by rule-based, NMT, and SMT correction models. The correction models are also used to adjust detection outputs and to generate candidate corrections for missing-word and word-selection errors.
& Training: CGED 2015--2018 for detection; Lang-8 parallel data for NMT; Gigaword, Chinese Wikipedia, CGED, and NLPCC data for SMT/language-model components. Reported detection experiments use the 17-dev set; correction ablation uses CGED-2018 ground truth.
& $F_{1}$
& Across NMT and SMT types, on detection ablations: best merge aggregation detection / identification / position $F_{1}$ = 0.6747 / 0.3566 / 0.1856, best voting eleven models = 0.6494 / 0.4669 / 0.2648; best correction ablation: $F_{1}$ = 0.0519. \\
\midrule

\citet{zan-etal-2020-chinese}
& Hybrid NLPTEA-2020 CGED system with BERT-BiLSTM-Attention-CRF for detection and an n-gram/Seq2Seq correction stage.
& Training: CGED 2016--2018 and CGED 2020 data for detection; NLPCC-2018 Task 2 sentence pairs for correction. Testing: CGED 2020 test set.
& FPR / $F_{1}$
& BERT-BiLSTM-Attention-CRF+Correction: FPR = 0.6645; detection / identification / position / correction $F_{1}$ = 0.8348 / 0.5035 / 0.1854 / 0.0017. \\
\midrule

\citet{liang-etal-2020-bert}
& Hybrid NLPTEA-2020 CGED system combining BERT-based position tagging, correction tagging, and BERT-fused NMT, with synthetic-data pretraining and weighted edit voting.
& Training: CGED 2014--2018, HSK, NLPCC-2018, and synthetic corpora. Testing: CGED 2020 test set.
& $F_{1}$
& Across submitted runs, best detection / identification $F_{1}$ = 0.8966 / 0.6463 (Run 3); best position / correction-top1 / correction-top3 $F_{1}$ = 0.3140 / 0.1891 / 0.1876 (Run 1). \\

\end{longtable}
\end{landscape}

\end{document}